\documentclass[10pt,twocolumn,letterpaper]{article}

\usepackage[pagenumbers]{cvpr} 

\usepackage[export]{adjustbox} 
\usepackage[table,xcdraw]{xcolor}
\usepackage{makecell}
\usepackage{booktabs}
\usepackage{multirow}
\usepackage{graphicx} 
\usepackage{subcaption}
\usepackage{amsthm} 

\usepackage{algorithm}
\usepackage{algpseudocode}
\definecolor{cvprblue}{rgb}{0.21,0.49,0.74}
\usepackage[pagebackref,breaklinks,colorlinks,allcolors=cvprblue]{hyperref}

\def\paperID{19117} 
\def\confName{CVPR}
\def\confYear{2026}

\title{UltraImage: Rethinking Resolution Extrapolation in \\ Image Diffusion Transformers}

\author{Min Zhao$^{1,2}$, Bokai Yan$^{3}$, Xue Yang$^{1,2}$, Hongzhou Zhu$^{1,2}$, Jintao Zhang$^{1,2}$, \\
Shilong Liu$^{4}$, Chongxuan Li$^{3}$, Jun Zhu$^{1,2}$ \\
$^{1}$Dept. of Comp. Sci. \& Tech., BNRist Center, Tsinghua University. $^{2}$ShengShu.\\
$^{3}$Gaoling School of Artificial Intelligence, Renmin University of China. 
$^{4}$ Princeton University.\\
\texttt{gracezhao1997@gmail.com, dcszj@tsinghua.edu.cn}
}

\usepackage{bm}

\def\eqref#1{Eqn.~(\ref{#1})}

\newcommand{\vect}[1]{\bm{#1}}

\newcommand{\xv}{\vect x}

\def\mK{{\bm{K}}}

\def\mO{{\bm{O}}}
\def\mP{{\bm{P}}}
\def\mQ{{\bm{Q}}}

\def\mS{{\bm{S}}}

\def\mV{{\bm{V}}}

\begin{document}
\maketitle

 \begin{figure*}
    \centering
    \includegraphics[width=2.0\columnwidth]{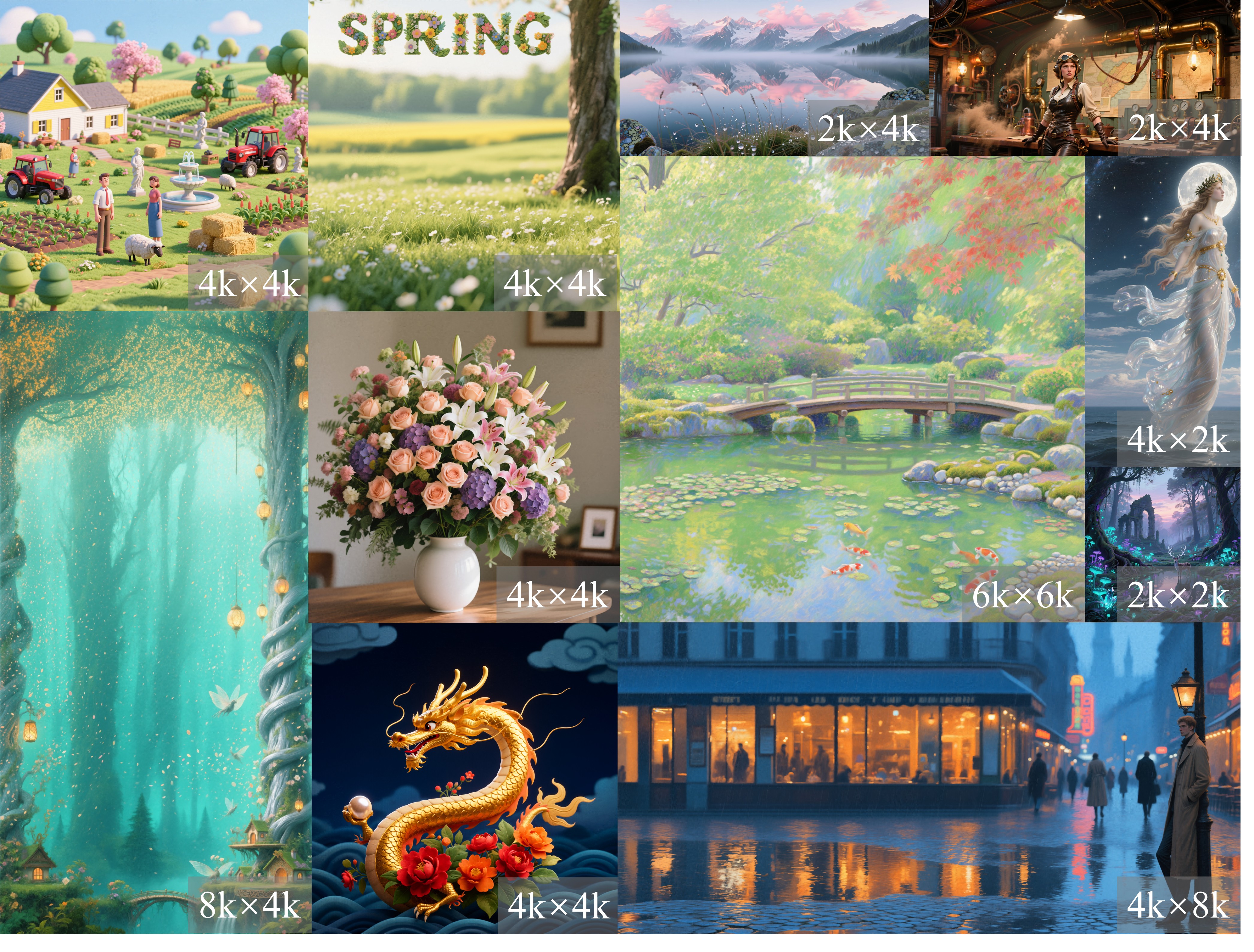}
  \caption{\textbf{Generated results of UltraImage.} Starting from the base Qwen-Image model trained at $1328\text{p}$ resolution, UltraImage can generate high-quality images up to $6\text{K} \times 6\text{K}$ without any low-resolution guidance, demonstrating its extreme extrapolation capability. All prompts used in this paper are provided in the Appendix.}
    \label{fig: demo}
\end{figure*}

\begin{abstract}
Recent image diffusion transformers achieve high-fidelity generation, but struggle to generate images beyond these scales, suffering from content repetition and quality degradation. In this work, we present UltraImage, a principled framework that addresses both issues. Through frequency-wise analysis of positional embeddings, we identify that repetition arises from the periodicity of the \emph{dominant frequency}, whose period aligns with the training resolution. We introduce a recursive dominant frequency correction to constrain it within a single period after extrapolation. Furthermore, we find that quality degradation stems from diluted attention and thus propose entropy-guided adaptive attention concentration, which assigns higher focus factors to sharpen local attention for fine detail and lower ones to global attention patterns to preserve structural consistency. Experiments show that UltraImage consistently outperforms prior methods on Qwen-Image and Flux (around 4K) across three generation scenarios, reducing repetition and improving visual fidelity. Moreover, UltraImage can generate images up to $6\text{K} \times 6\text{K}$ without low-resolution guidance from a training resolution of 1328p, demonstrating its extreme extrapolation capability. Project page is available at \href{https://thu-ml.github.io/ultraimage.github.io/}{https://thu-ml.github.io/ultraimage.github.io/}.
\end{abstract}    
\section{Introduction}
\label{sec:intro}

Building upon the expressive power of image diffusion transformers~\cite{peebles2023scalable,bao2023all}, recent advances in text-to-image generation~\cite{podell2023sdxl,yang2023diffusion,wallace2024diffusion,chen2023pixart,yin2024one} have enabled high-fidelity image synthesis. Despite these advances, these models still struggle to generate ultra-resolution images beyond their trained spatial scale~\cite{flux2024,wu2025qwen,zhuo2024lumina}, a task we refer to as resolution extrapolation. This capability is critical for practical applications such as large-format printing, high-resolution content creation, and detailed visual simulations, where generating images at scales larger than the training resolution is required.

To investigate the challenges of resolution extrapolation in image diffusion transformers, we first conduct experiments on two representative models, Flux~\cite{flux2024} and Qwen-Image~\cite{wu2025qwen}. We identify two typical failure modes: \emph{content repetition}, where visual elements repeat periodically across the image, and \emph{quality degradation}, where fine details are blurred. Existing positional extrapolation methods either fail to mitigate repetition or reduce repetition at the cost of introducing over-smoothed textures. We argue that these failures stem from an incomplete understanding of frequency components in position encoding, leading to flawed modifications or targeting incorrect frequencies.

In this paper, we establish a principled guideline for \emph{which} frequency components should be modified and \emph{how} to modify them. Through systematic analysis, we find that the high-frequency component, with short period, primarily governs local textures, and interpolating it leads to blurring. The low-frequency component has minimal impact. In contrast, the mid-band frequency—whose period is comparable to the training resolution—controls global structure; we refer to it as \emph{dominant frequency}. By ensuring that the dominant frequency remain within a single period after extrapolation, repetition can be effectively mitigated. Furthermore, due to dynamic-resolution training, some models may contain multiple dominant frequencies near the training resolution. To address this, we apply a \emph{recursive dominant frequency correction} procedure to the dominant frequency until repetition is resolved.

To address the second challenge, quality degradation, our analysis finds that the observed blurring stems from flattened attention distributions that dilute focus. 
Applying a single global focus factor sharpens local details but disrupts long-range dependencies, creating a trade-off between fine textures and structural consistency. This trade-off arises from the functional specialization of attention patterns: global patterns (high-entropy) require lower focus factor to preserve structural coherence, while local patterns (low-entropy) benefit from higher focus factor to enhance fine details. To resolve this, we propose an \emph{entropy-guided adaptive concentration} strategy, which assigns a distinct focus factor to each attention pattern based on its entropy. Furthermore, to handle high-resolution images, we implement a custom Triton kernel that computes the softmax in a block-wise, online fashion, avoiding out-of-memory issues while dynamically applying the pattern-specific focus. 

Extensive experiments on Flux~\cite{flux2024} and Qwen-Image~\cite{wu2025qwen} demonstrate the effectiveness of UltraImage. Our method consistently outperforms state-of-the-art baselines~\cite{bu2025hiflow,peng2023yarn,chen2023extending,bloc97,du2024max} across almost all metrics including FID, KID, and CLIP score around $4\text{K}$, in three generation scenarios—direct resolution extrapolation, guided resolution extrapolation, and guided view extrapolation. Qualitative results further confirm that UltraImage reduces repetition and improves visual fidelity, validating its practicality for generating ultra-resolution outputs beyond the training scale. Moreover, UltraImage can generate images up to $6\text{K} \times 6\text{K}$ from a training resolution of 1328p, demonstrating its extreme extrapolation capability.

\section{Background}
\label{sec: background}

\textbf{Rotary position embedding (RoPE) in image diffusion transformers.} RoPE~\cite{su2021roformer} is a widely adopted method in transformers for encoding relative positional information. RoPE works by rotating pairs of feature dimensions according to their positions, effectively introducing structured, position-dependent modulation. 
For a one-dimensional sequence, given an input vector $\xv \in \mathbb{R}^d$ at position $p$, the embedding is computed as
\begin{equation}
\label{eqn:RoPE-Complex-rewrite}
\boldsymbol f^{\mathrm{RoPE}}(\xv, p, \bm{\theta})_j =
\begin{bmatrix}
\cos(p \theta_j) & -\sin(p \theta_j) \\
\sin(p \theta_j) & \cos(p \theta_j)
\end{bmatrix}
\begin{bmatrix}
x_{2j} \\ x_{2j+1}
\end{bmatrix}, 
\end{equation}
where $j = 1, \dots, d^\prime/2$, $\bm{\theta} \in \mathbb{R}^{d^\prime/2}$ defines the frequency for each dimension, and $b$ controls the base of the exponential schedule $\theta_j = b^{-2(j-1)/d^\prime}$.  For images with height and width, existing works use two independent RoPE embeddings: one along the height axis with frequencies $\theta_i^h$, and one along the width axis with frequencies $\theta_i^w$~\cite{wu2025qwen,flux2024}. 
Each axis applies the standard single-axis RoPE, and the resulting embeddings are concatenated to obtain the final positional encoding.

\textbf{Attention mechanism.} 
Given an input image with height $H$ and width $W$, let $\mQ, \mK \in \mathbb{R}^{HW \times D}$ and $\mV \in \mathbb{R}^{HW \times D'}$ denote the queries, keys, and values after RoPE. 
The attention logits, scores, and output are then computed as
\begin{equation}
\label{eq:attention}
\mS = \mQ \mK^\top, \quad
\mP = \text{softmax}\Big(\frac{\mS}{\sqrt{D}}\Big), \quad
\mO = \mP \mV  ,
\end{equation}
where $\mS \in \mathbb{R}^{HW \times HW}$ is the attention logits, $\mP$ stores the pairwise similarities between queries and keys, and $\mO \in \mathbb{R}^{HW \times D'}$ is the final attended output.

\textbf{Length extrapolation in image generation.} Extending sequence lengths in transformers is challenging due to positional encoding limitations. Position Interpolation (PI)~\cite{chen2023extending} rescales RoPE frequencies to match the target sequence length, $\boldsymbol\theta'= \boldsymbol\theta / s$, where $s = L'/L$ and $L$, $L^\prime$ is the sequence length for training and inference, respectively. NTK-Aware Scaled RoPE (NTK)~\cite{bloc97, zhuo2024lumina} adjusts the base frequency $b$ for all dimensions to extrapolate high-frequency components while interpolate low-frequency ones:
\begin{equation}
    \theta_j^{\mathrm{NTK}} = (\lambda b)^{-2(j-1)/d^\prime}, \lambda = s^{d^\prime / (d^\prime - 2)}, j = 1,\ldots,d^\prime/2.
\end{equation}
YaRN~\cite{peng2023yarn} further refines this via frequency-wise grouping and gradual interpolation–extrapolation. For images, these adjustments can be applied independently along height and width, termed Vision NTK and Vision YaRN~\cite{lu2024fit}. 

In early U-Net-based image generation, extending the spatial resolution often leads to object repetition due to the limited convolutional receptive field~\cite{he2023scalecrafter}. 
Methods such as ScaleCrafter~\cite{he2023scalecrafter} or DemoFusion~\cite{du2024demofusion} incorporating global perceptive layers have been proposed to enlarge the effective receptive field and mitigate repetition. 
Further analysis in FouriScale attribute repetitive artifacts to frequency misalignment across scales~\cite{huang2024fouriscale}, highlighting challenges in naive length extrapolation. Recent transformer-based approaches leverage low-resolution cues for high-resolution synthesis, e.g., I-Max~\cite{du2024max} and HiFlow~\cite{bu2025hiflow} project or guide high-resolution flows using low-resolution information, enabling training-free or guided super-resolution. These methods focus on external guidance, whereas the intrinsic extrapolation ability of image diffusion transformers remains underexplored. Recent work on video length extrapolation~\cite{zhao2025riflex} closely aligns with our finding in image generation. See more work in the related work section in supplementary materials.

\section{Problem Setting and Challenges}
\label{sec: challenge}
\renewcommand\arraystretch{1.1}
\begin{figure}[htbp]
    \centering
    \begin{tabular}{
        >{\centering\arraybackslash}m{0.13\textwidth} |
        >{\centering\arraybackslash}m{0.13\textwidth} |
        >{\centering\arraybackslash}m{0.13\textwidth}
    }
    \toprule
    \textbf{$1024^2$} & \textbf{$3200^2$} & \textbf{$4096^2$} \\ 
    \midrule
    \includegraphics[width=\linewidth,valign=m]{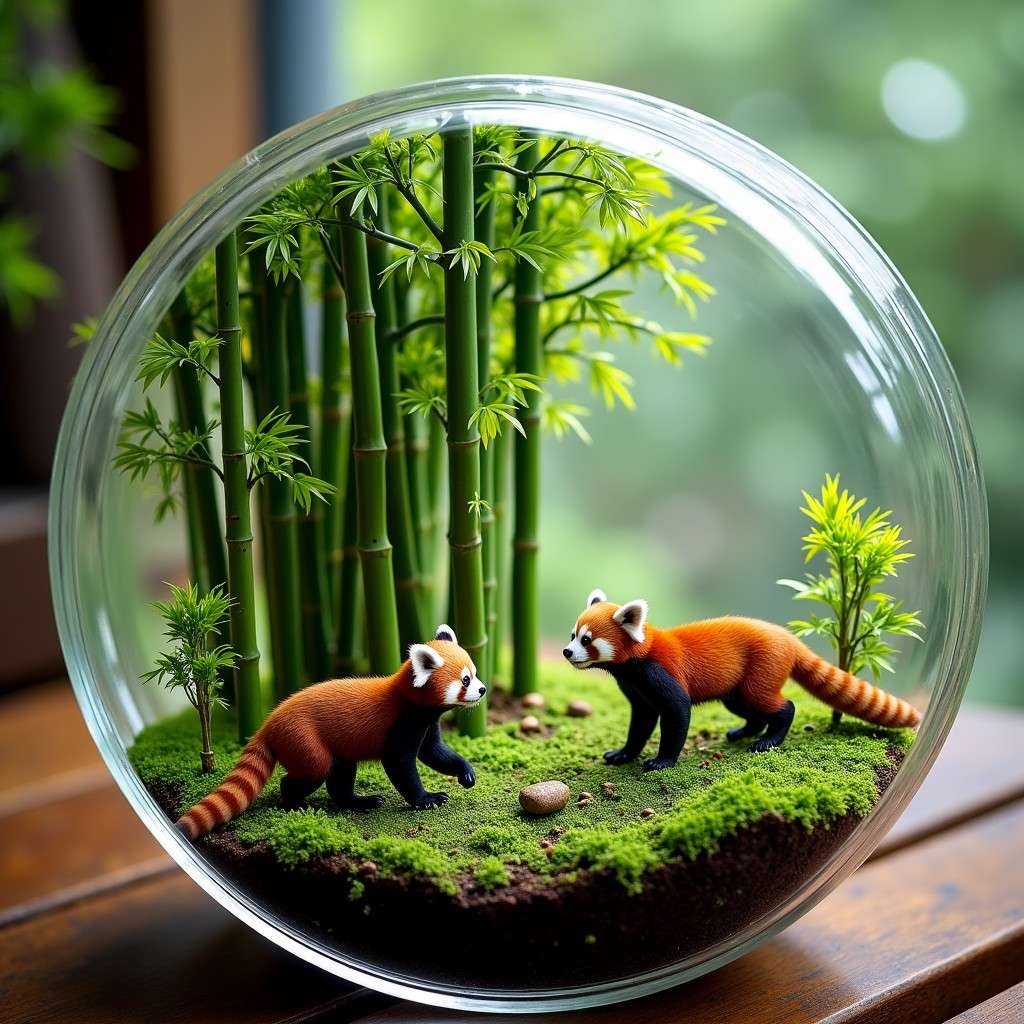} &
    \includegraphics[width=\linewidth,valign=m]{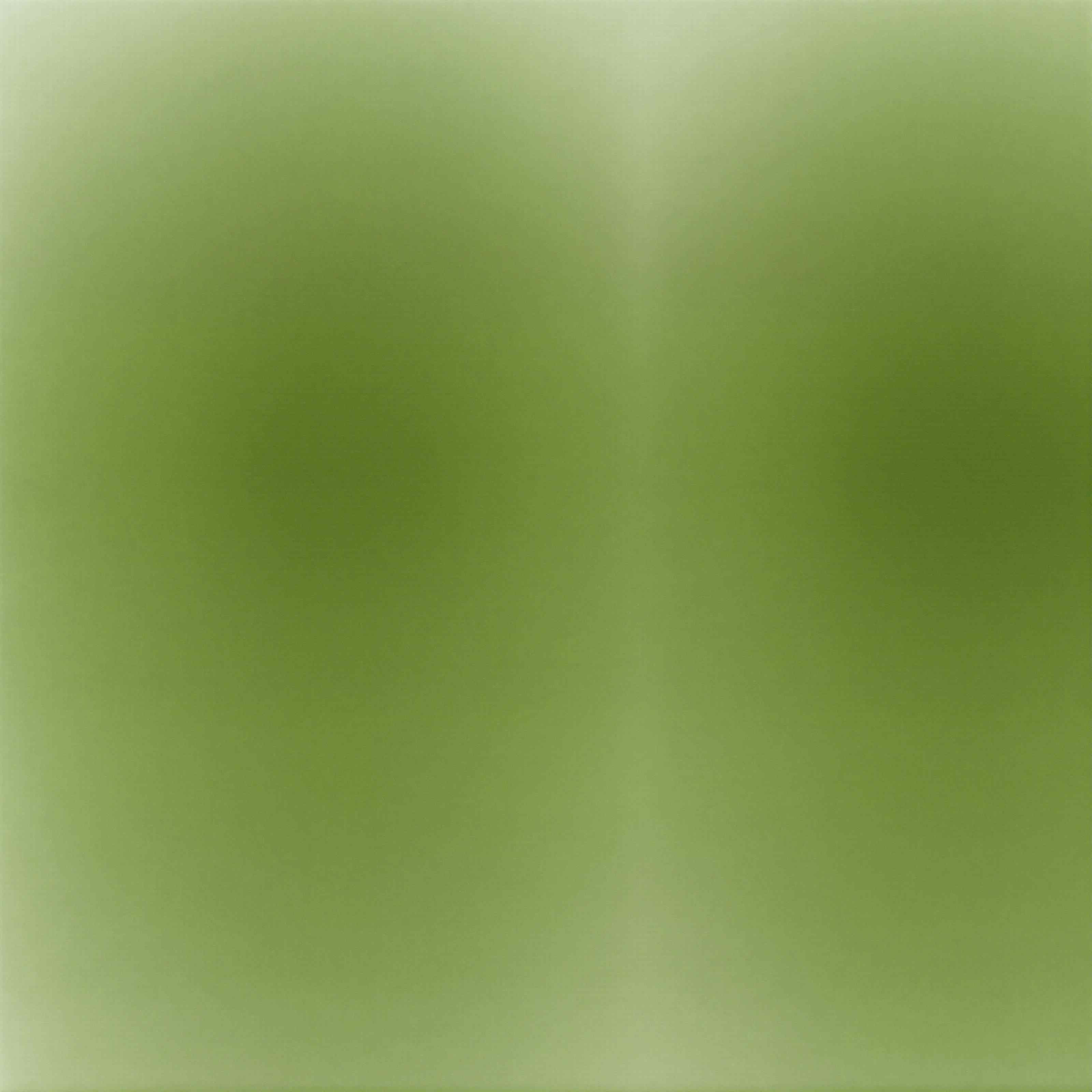} &
    \includegraphics[width=\linewidth,valign=m]{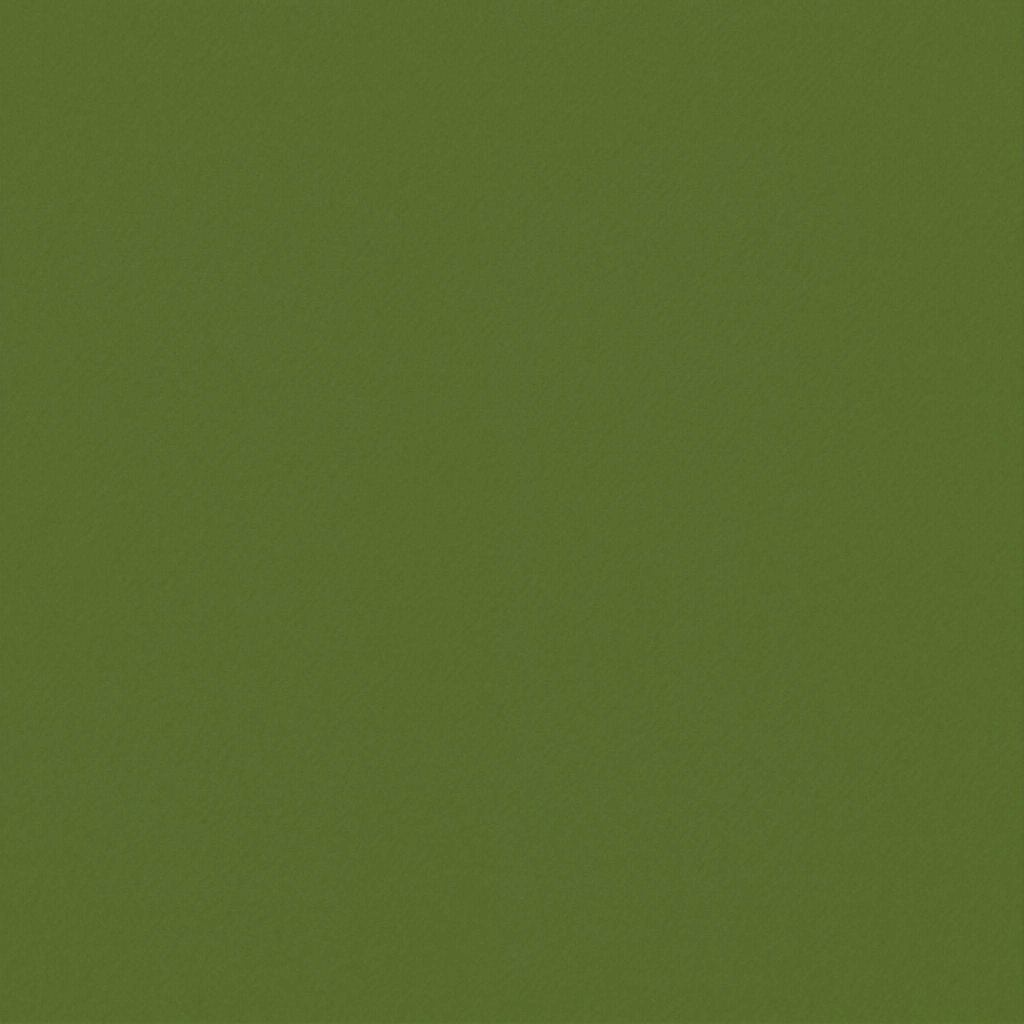} \\[1ex]
    \midrule
    \includegraphics[width=\linewidth,valign=m]{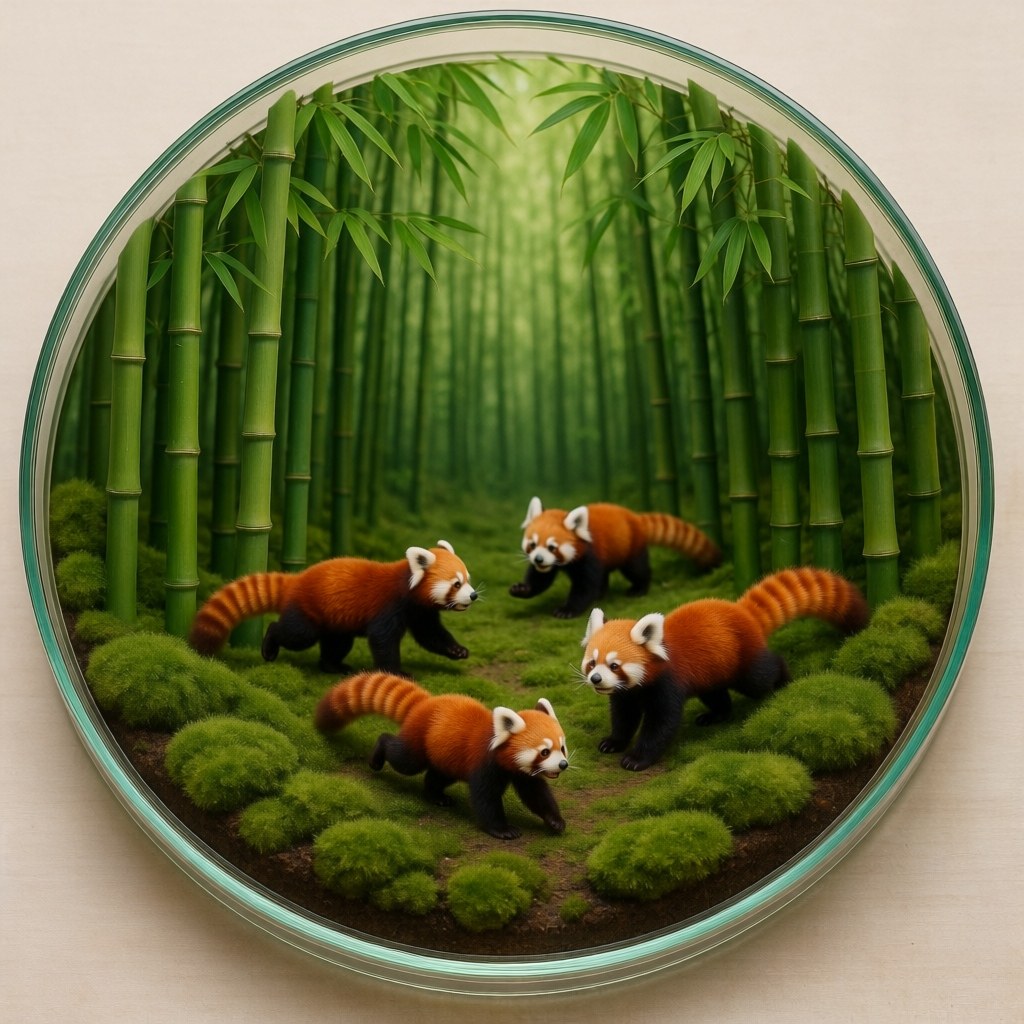} &
    \includegraphics[width=\linewidth,valign=m]{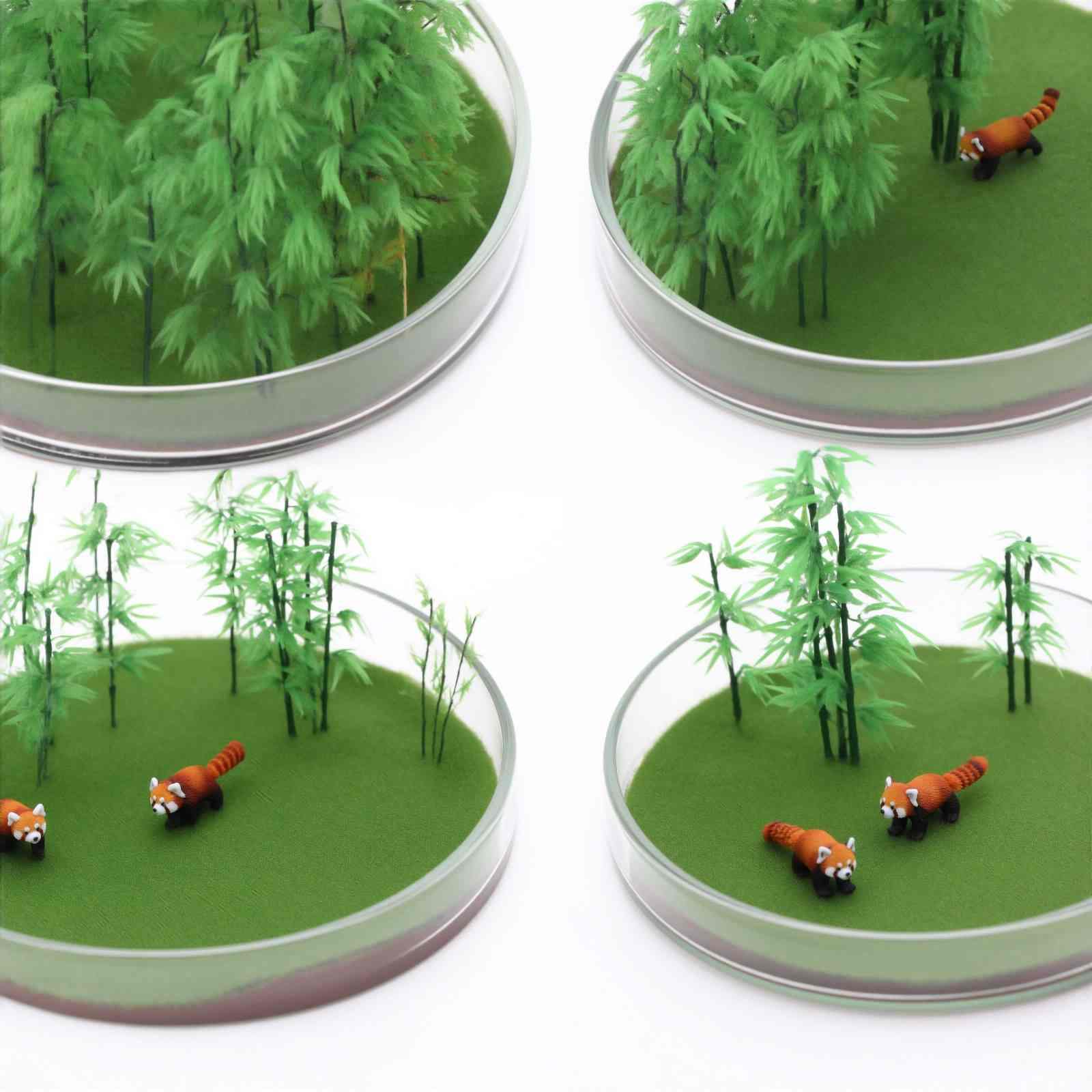} &
    \includegraphics[width=\linewidth,valign=m]{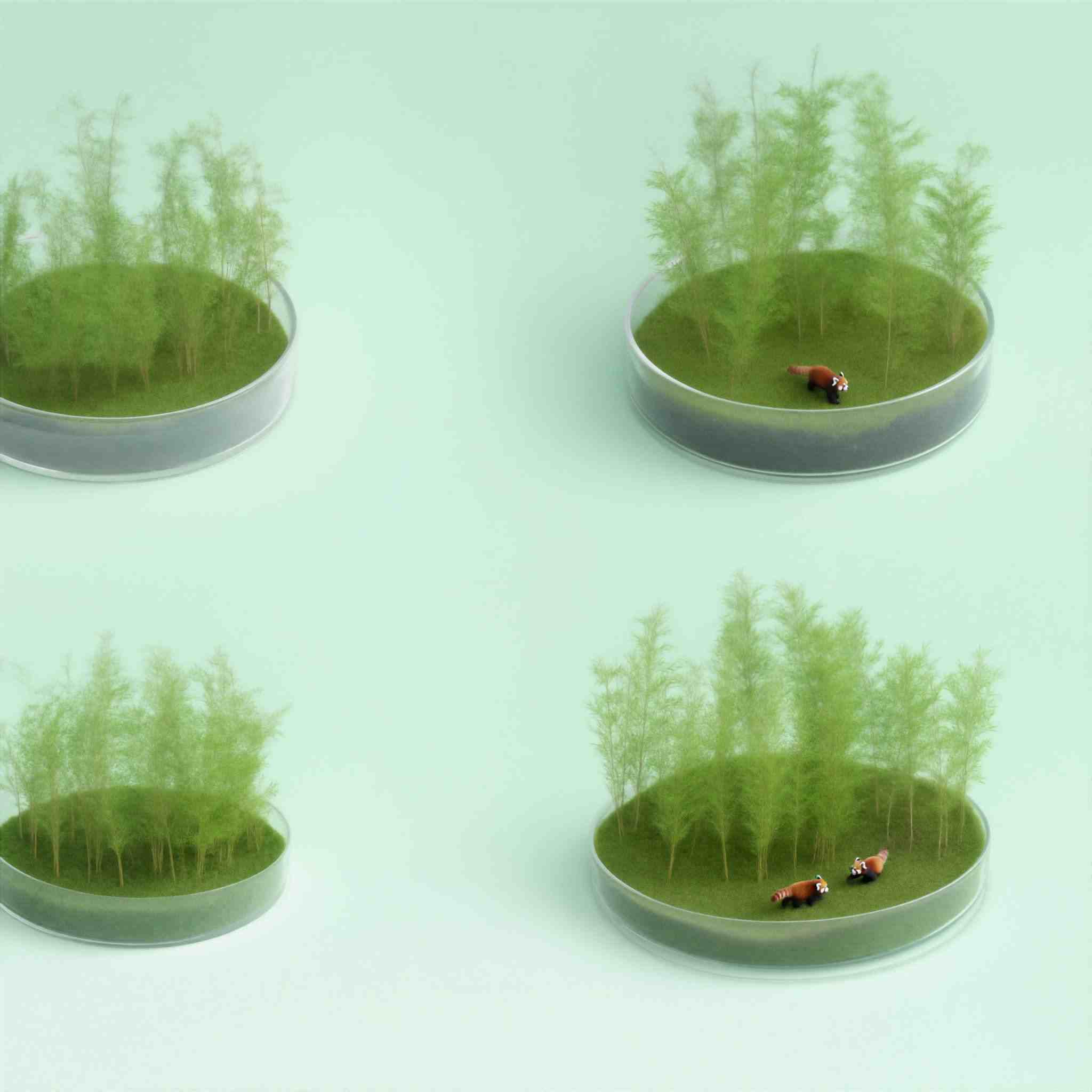} \\[1ex]
    \bottomrule
    \end{tabular}
    \caption{\textbf{Challenges of resolution extrapolation in image diffusion transformers.} Top row: Flux (training resolution with $2048^2$). Bottom row: Qwen-Image (training resolution $1328\text{p}$). Both models exhibit typical failure modes: \emph{content repetition} and \emph{quality degradation} at higher resolutions.}
    \label{fig:challenge}
    
\end{figure}

In this paper, we address the task of \emph{resolution extrapolation} for image diffusion transformers.
Given a model trained on a resolution $h \times w$, the goal is to enable it to generate high-quality and spatially coherent images beyond its trained spatial scale, at larger resolutions $H \times W$, without any re-training.
We define the extrapolation ratios as
$s_h = H / h$ and $s_w = W / w$, where $s_h \ge 1, s_w \ge 1$.

\renewcommand\arraystretch{1.1}
\begin{figure*}[t]
\centering
 \begin{tabular}{
 >{\centering\arraybackslash}m{0.12\textwidth} | 
 >{\centering\arraybackslash}m{0.12\textwidth} | 
 >{\centering\arraybackslash}m{0.12\textwidth} | 
 >{\centering\arraybackslash}m{0.12\textwidth} || 
 >{\centering\arraybackslash}m{0.12\textwidth} | 
 >{\centering\arraybackslash}m{0.12\textwidth} | 
 >{\centering\arraybackslash}m{0.12\textwidth} 
}
 \toprule
    
    \multicolumn{4}{c||}{\textbf{Role of each frequency in RoPE}} & 
    \multicolumn{3}{c}{\textbf{Validation of the repetition cause}} \\ 
    \cmidrule(lr){1-4} \cmidrule(lr){5-7} 
    
 \textbf{reference} & \textbf{high frequency} ($T_0^h \ll h$) & \textbf{mid-band frequency} ($T_8^h \approx h$) & \textbf{low frequency} ($T_{20}^h \gg h $) &  $H>T_8^h$ & $H=T_8^h$ & $H < T_8^h $ \\
 \cmidrule(lr){1-4} \cmidrule(lr){5-7} 
    
\includegraphics[width=\linewidth,valign=m]{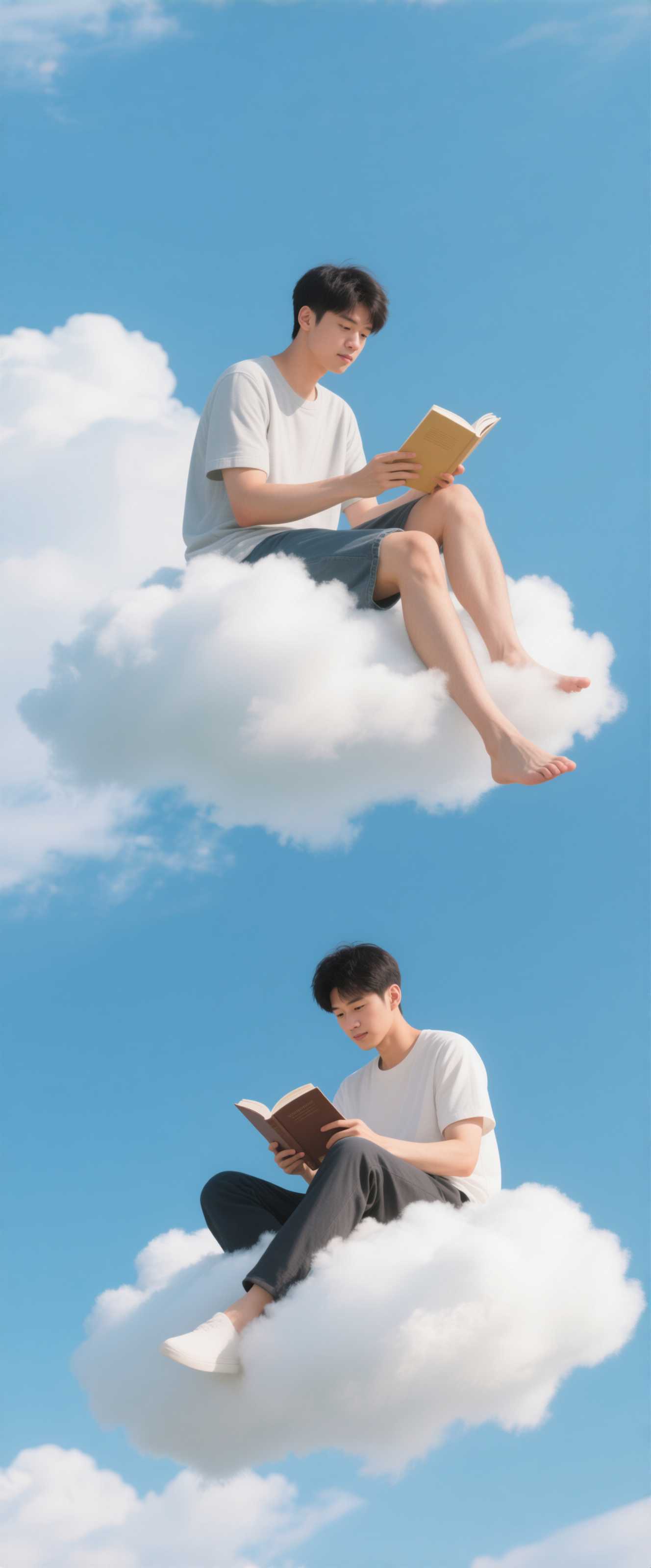} &
 \includegraphics[width=\linewidth,valign=m]{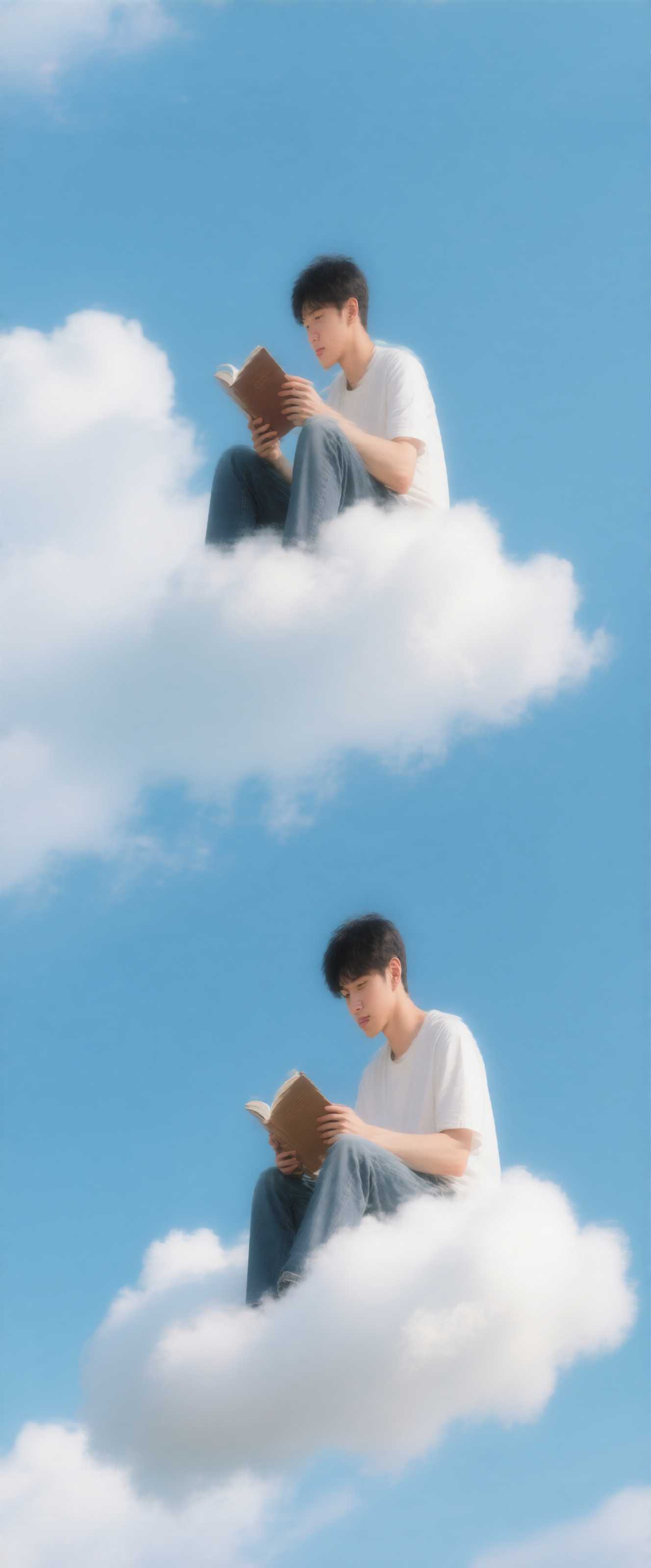} &
 \includegraphics[width=\linewidth,valign=m]{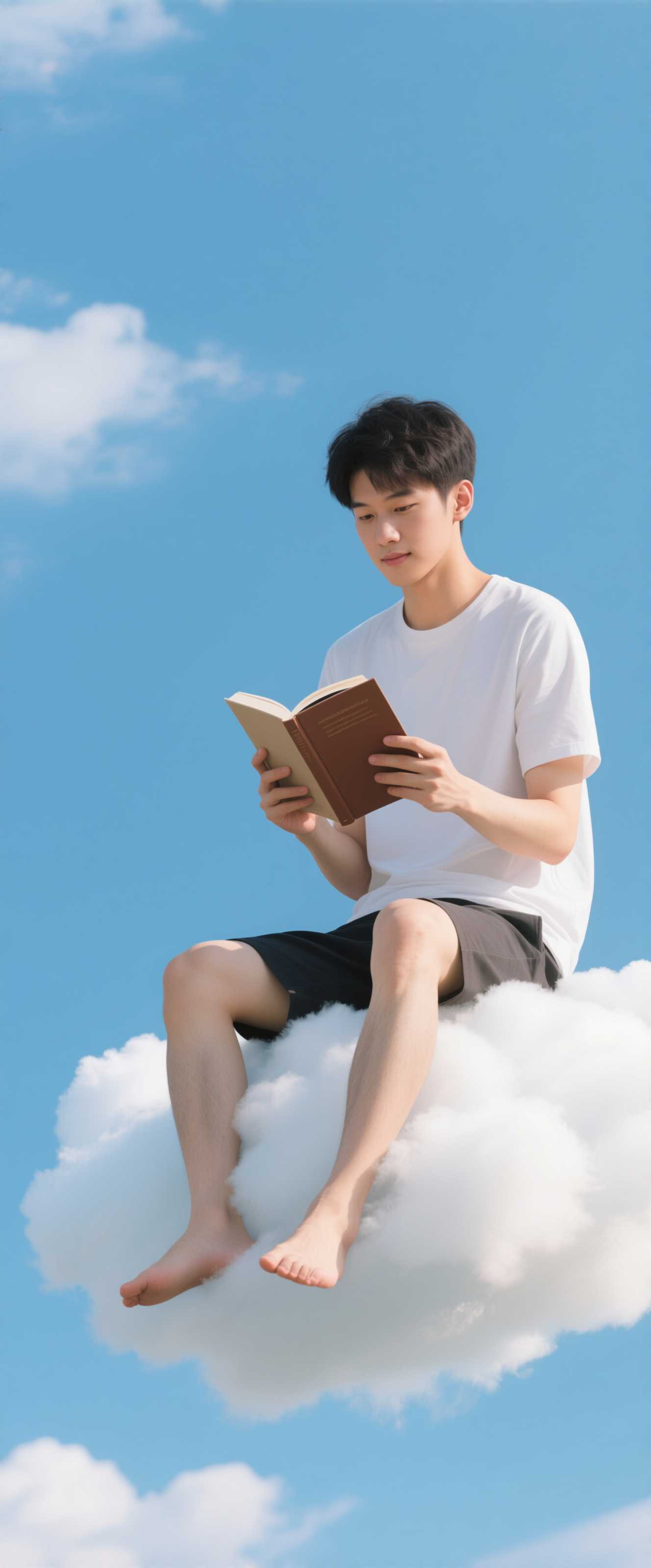} &
 \includegraphics[width=\linewidth,valign=m]
 {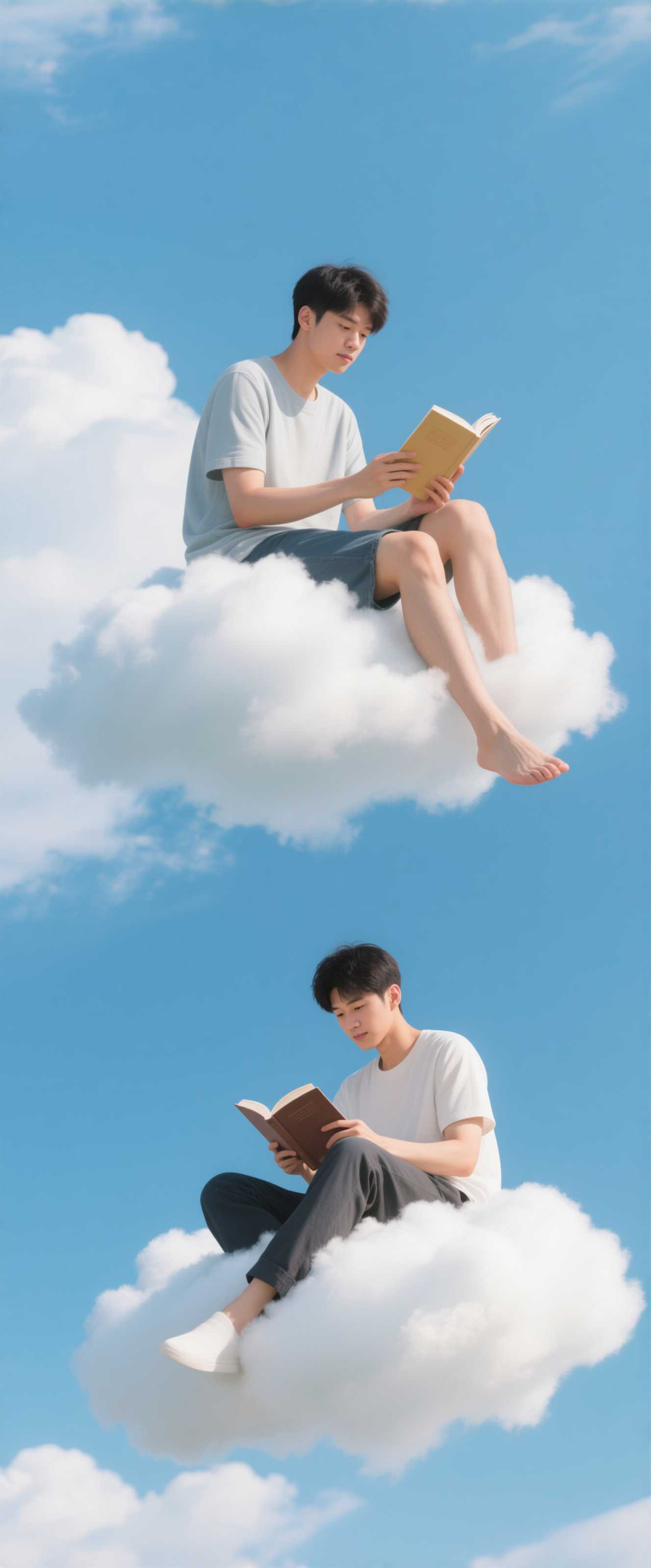} &
 \includegraphics[width=\linewidth,valign=m]{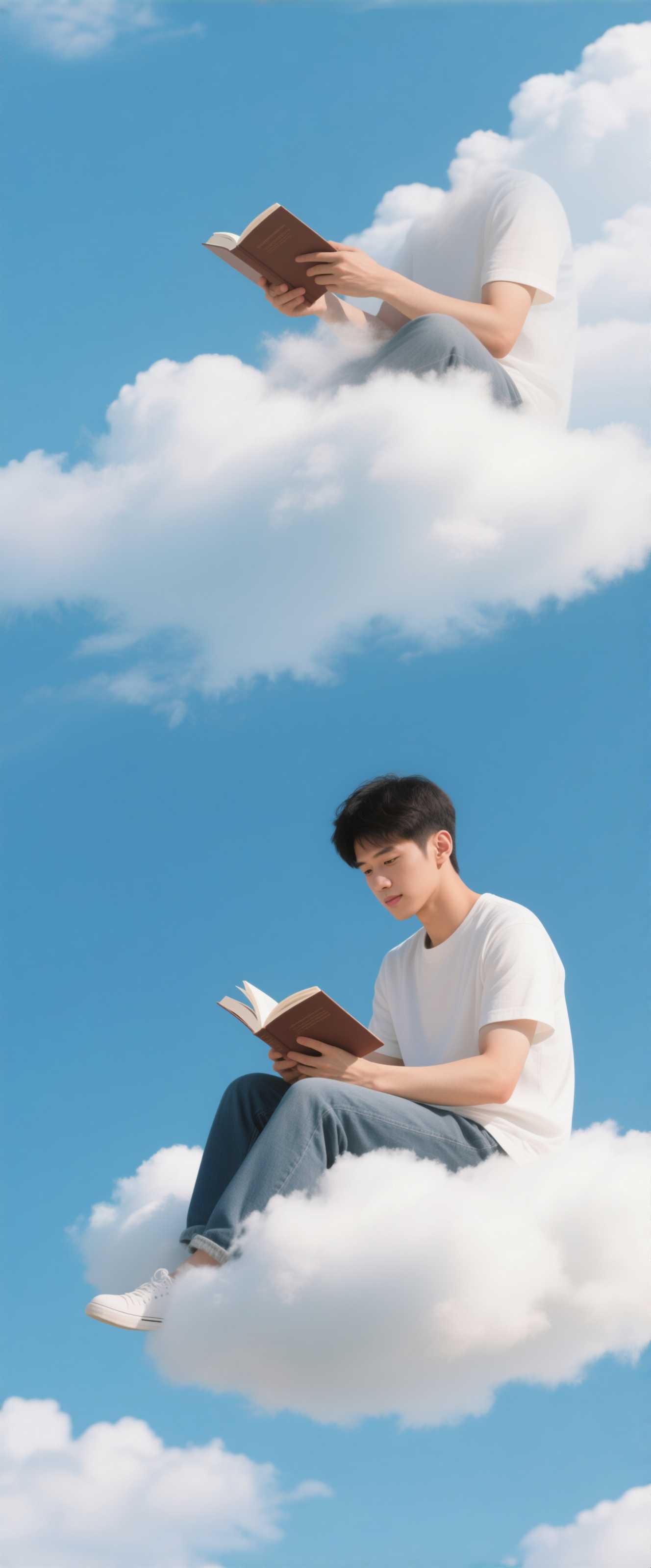} &
 \includegraphics[width=\linewidth,valign=m]{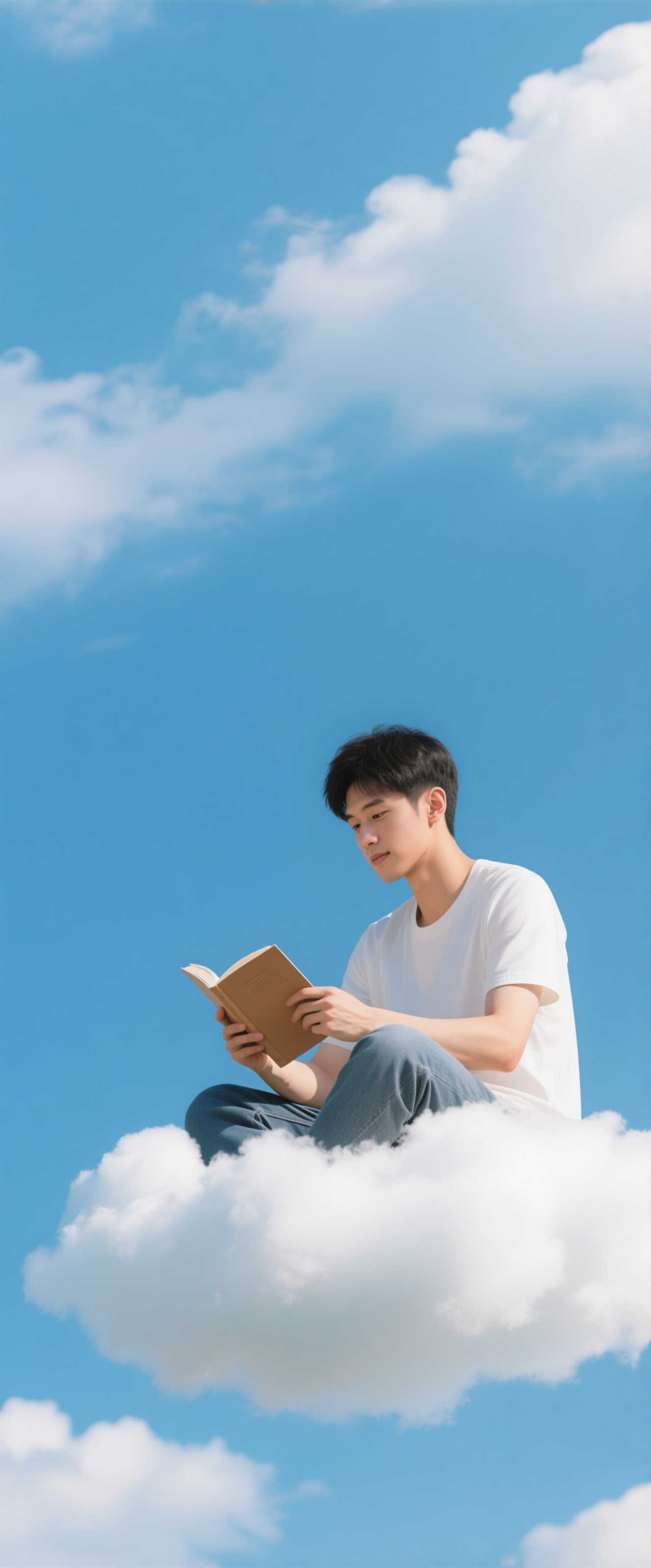} &
 \includegraphics[width=\linewidth,valign=m]{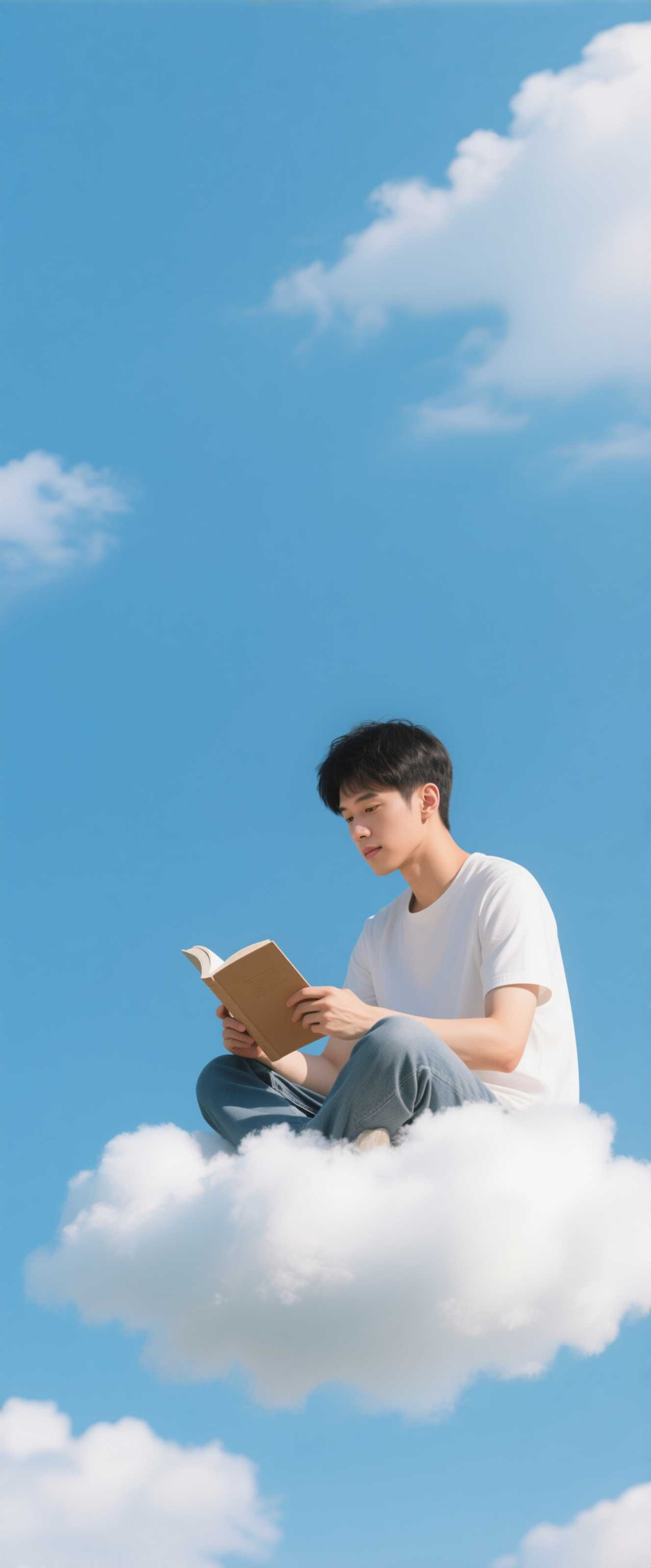}
 \\[1ex]
 \small{(a)} & \small{(b)} & \small{(c)} & \small{(d)} & \small{(e)} & \small{(f)} & \small{(g)} \\
 \bottomrule
 \end{tabular}

 \caption{\textbf{Cause of content repetition.}
\textbf{Left:} (a) Height extrapolation baselines. (b) High-frequency interpolation blurs local textures. (c) The mid-band \emph{dominant frequency}, whose period aligns with the training height~$h$, governs global structure and introduces repetition. (d) Low-frequency components have minimal effect.
\textbf{Right:} Validation: repetition appears when the extrapolated height $H$ (e) exceeds the dominant period $T_k^h$, and disappears when $H \le T_k^h$ (f,g).}

 \label{fig:each frequency}
\end{figure*}

When directly applying diffusion transformers such as Flux~\cite{flux2024} and Qwen-Image~\cite{wu2025qwen} beyond their training resolution, we identify two typical failure modes (see Fig.~\ref{fig:challenge}).
The first is \emph{content repetition}, where visual content repeats periodically across regions.
The second is \emph{quality degradation}, where texture fidelity declines and images become overly blurred, The second is quality degradation, where texture fidelity declines and images become overly blurred, especially under large extrapolation. In the following sections, we analyze and address them separately.

\section{Understanding and Solving Repetition from Positional Encoding}

In this section, we first focus on the problem of spatial repetition, analyzing its origin from positional encoding in Sec.~\ref{sec: repetition analysis} and then proposing a frequency-based solution in Sec.~\ref{sec: repetition method} to solve it.

\subsection{Frequency Analysis of RoPE in Image Diffusion Transformers}

\label{sec: repetition analysis}

\renewcommand\arraystretch{1.1}
\begin{figure}[htbp]
    \centering
    \begin{tabular}{
        >{\centering\arraybackslash}m{0.13\textwidth} |
        >{\centering\arraybackslash}m{0.13\textwidth} |
        >{\centering\arraybackslash}m{0.13\textwidth}
    }
    \toprule
    \textbf{NTK} & \textbf{PI} & \textbf{Ours}\\ 
    \midrule
    \includegraphics[width=\linewidth,valign=m]{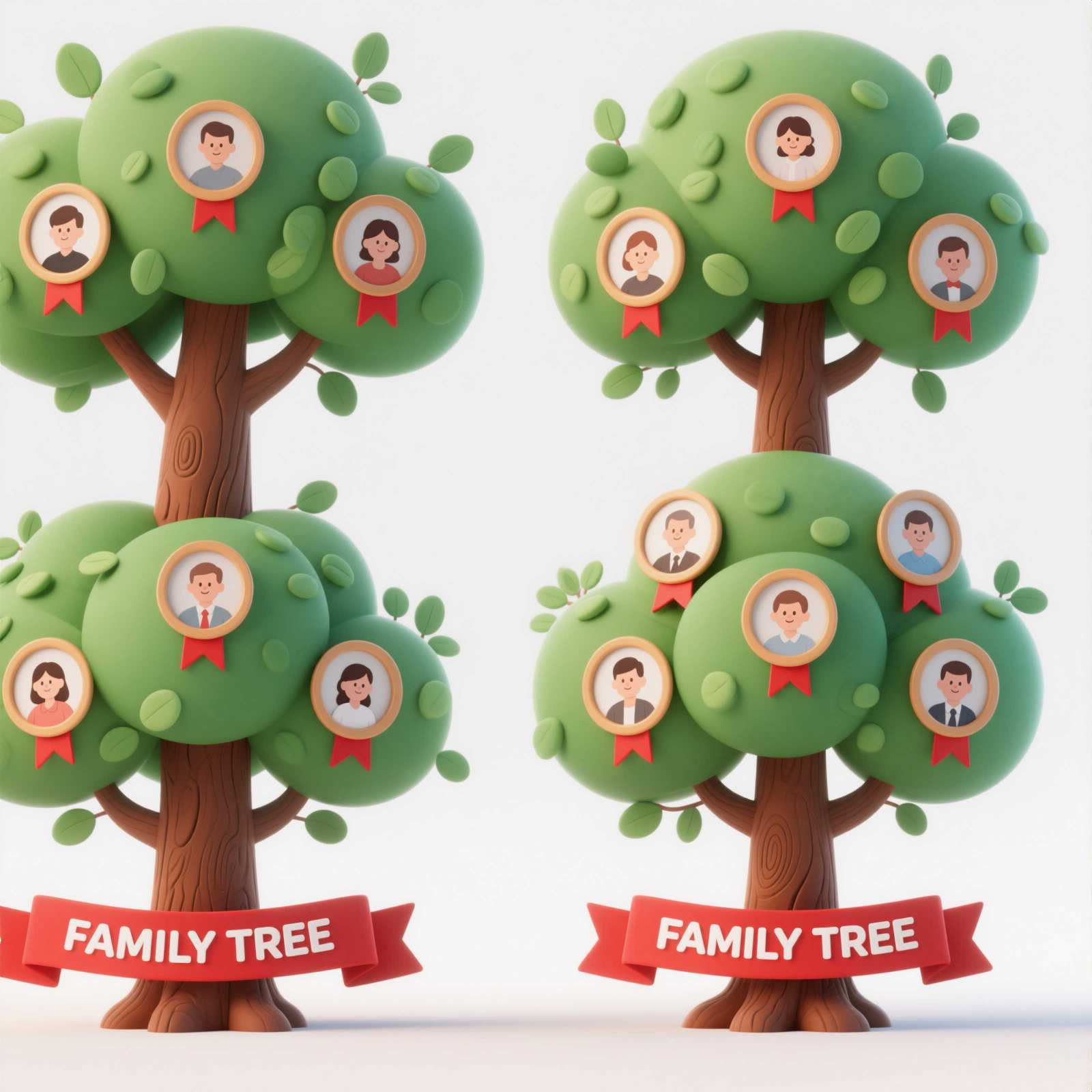} &
    \includegraphics[width=\linewidth,valign=m]{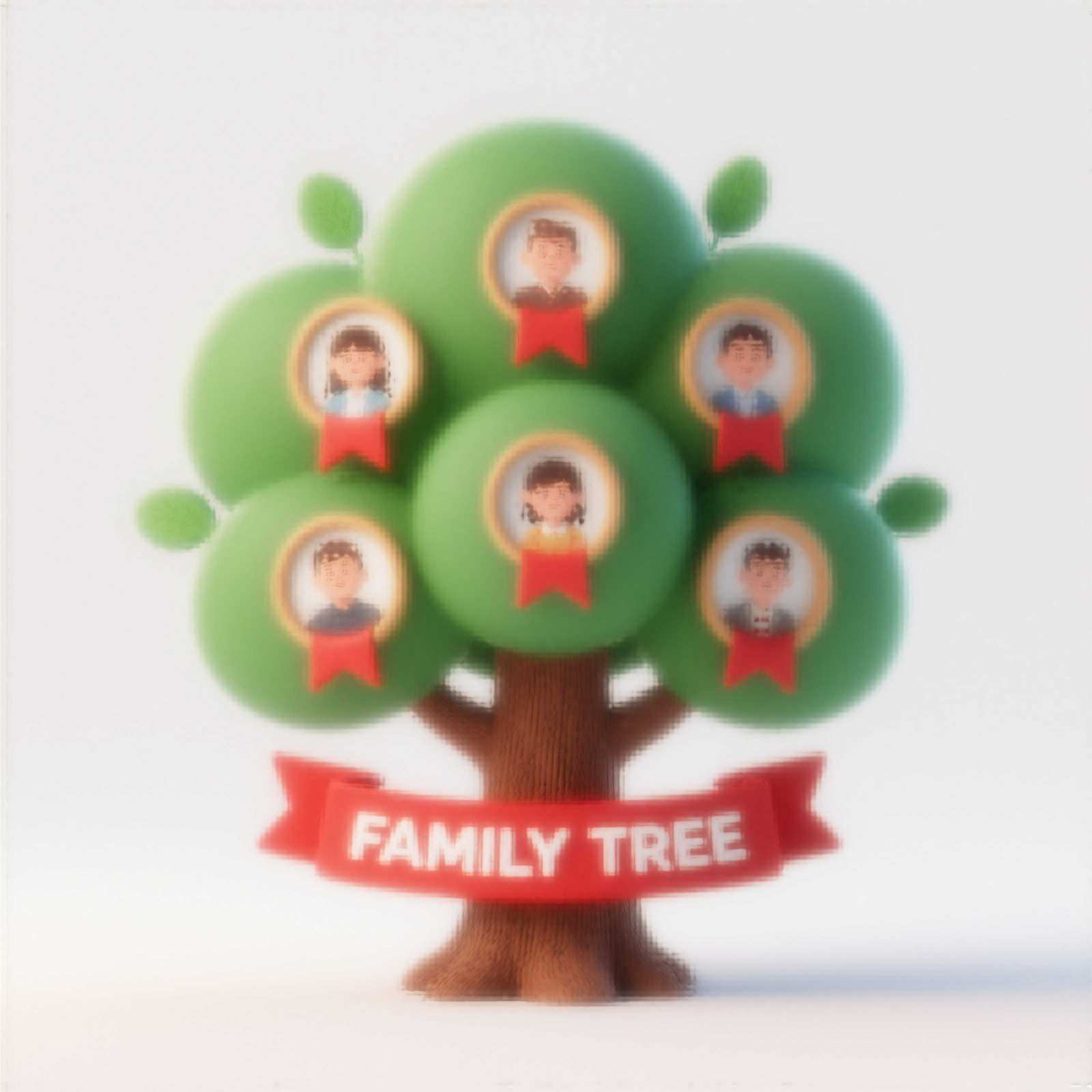} &
    \includegraphics[width=\linewidth,valign=m]{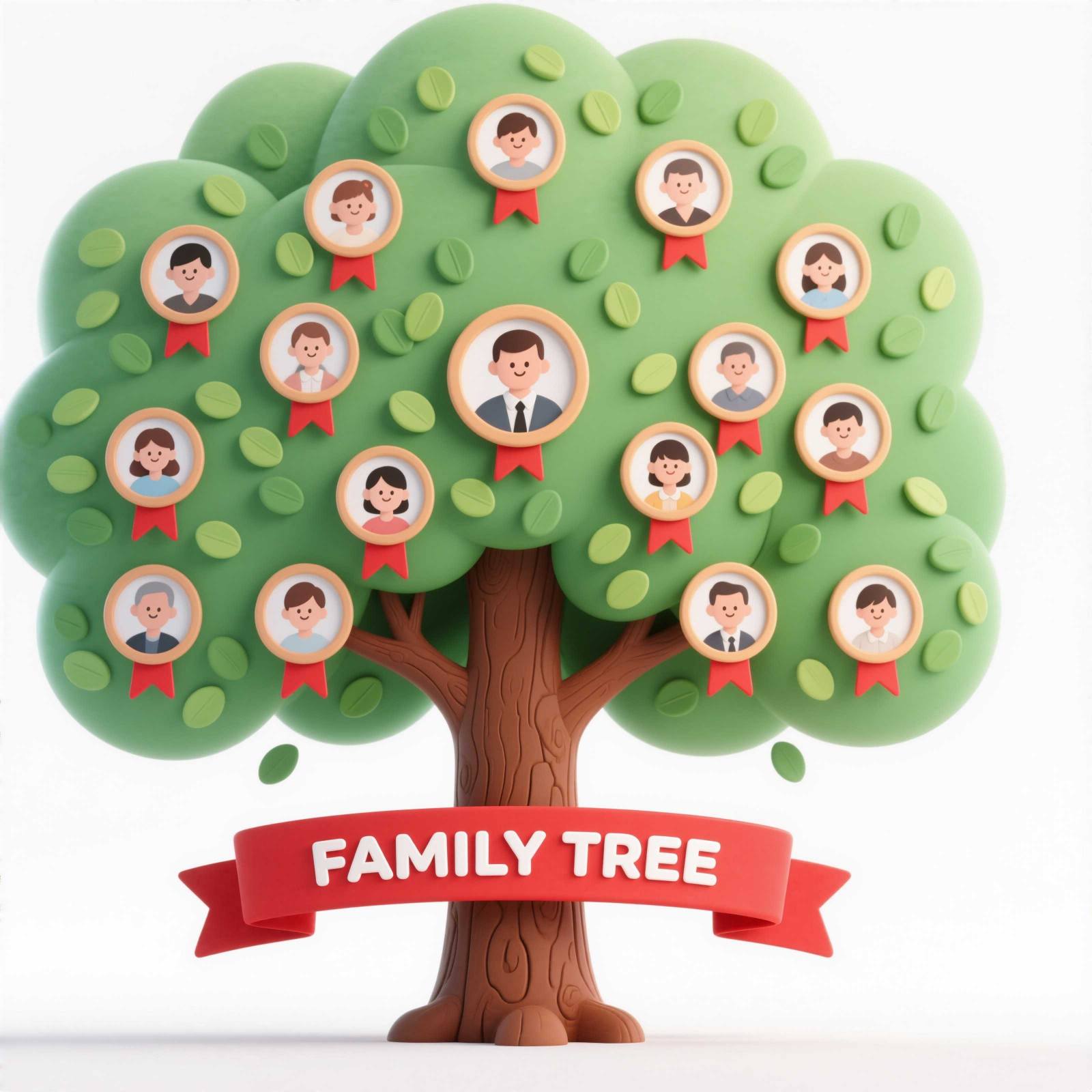} 
    \\
    \bottomrule
    \end{tabular}
    \caption{\textbf{Failure modes of existing positional extrapolation.} NTK suffers from repetition during extrapolation, while PI reduces it but causes over-smoothed textures. In contrast, our method mitigates repetition without sacrificing image fidelity by correctly identifying and adjusting the relevant frequency components.}
    \label{fig: ntk and pi}
\end{figure}

Inspired by advances in language models, prior works have attempted to mitigate spatial repetition by modifying positional encodings~\cite{zhuo2024lumina,lu2024fit,du2024max}. For example, NTK alleviates repetition via frequency scaling, while PI interpolates all RoPE components for structural preservation. However, as shown in Fig.~\ref{fig: ntk and pi}, NTK still shows noticeable repetition, and PI introduces over-smoothed textures similar to those in Sec.~\ref{sec: challenge}. We argue that these issues stem from targeting incorrect components or modifying them improperly, which leads to suboptimal results. In this paper, we aim to establish a principled guideline by answering two key questions:
\begin{enumerate}
    \item \emph{Which frequencies} in RoPE are primarily responsible for governing image structure and causing repetition?
    \item \emph{How} should these specific frequencies be adjusted to mitigate repetition while preserving image fidelity?
\end{enumerate}

\paragraph{The role of each frequency component.} 
To answer the above questions, we analyze different RoPE frequency components by interpolating one component at a time along a spatial dimension (e.g., height), where $\theta_i^h$ is scaled by $1/s_h$. As shown in Fig.~\ref{fig:each frequency}, \textbf{different frequency components control visual features at different scales.} The High-frequency component primarily affects \emph{local textures}—interpolating it causes blurring but preserves overall layout (Fig.~\ref{fig:each frequency}b). The low-frequency components have negligible impact when modified (Fig.~\ref{fig:each frequency}d). In contrast, the mid-band frequency governs \emph{global structure}, and adjusting it effectively eliminates content repetition (Fig.~\ref{fig:each frequency}c). These results indicate that spatial repetition mainly arises from the mid-band frequency responsible for global structure.

\renewcommand\arraystretch{1.1}

\begin{figure*}[t]
    \centering
    \begin{tabular}{
        >{\centering\arraybackslash}m{0.35\textwidth} |
        >{\centering\arraybackslash}m{0.18\textwidth} |
        >{\centering\arraybackslash}m{0.18\textwidth} |
        >{\centering\arraybackslash}m{0.18\textwidth} 
    }
    \toprule
    \textbf{Static attention scores} & \textbf{Reference}& \textbf{HF interpolation}& \textbf{Refocused HF}\\ 
    \midrule
    \includegraphics[width=\linewidth,valign=m]{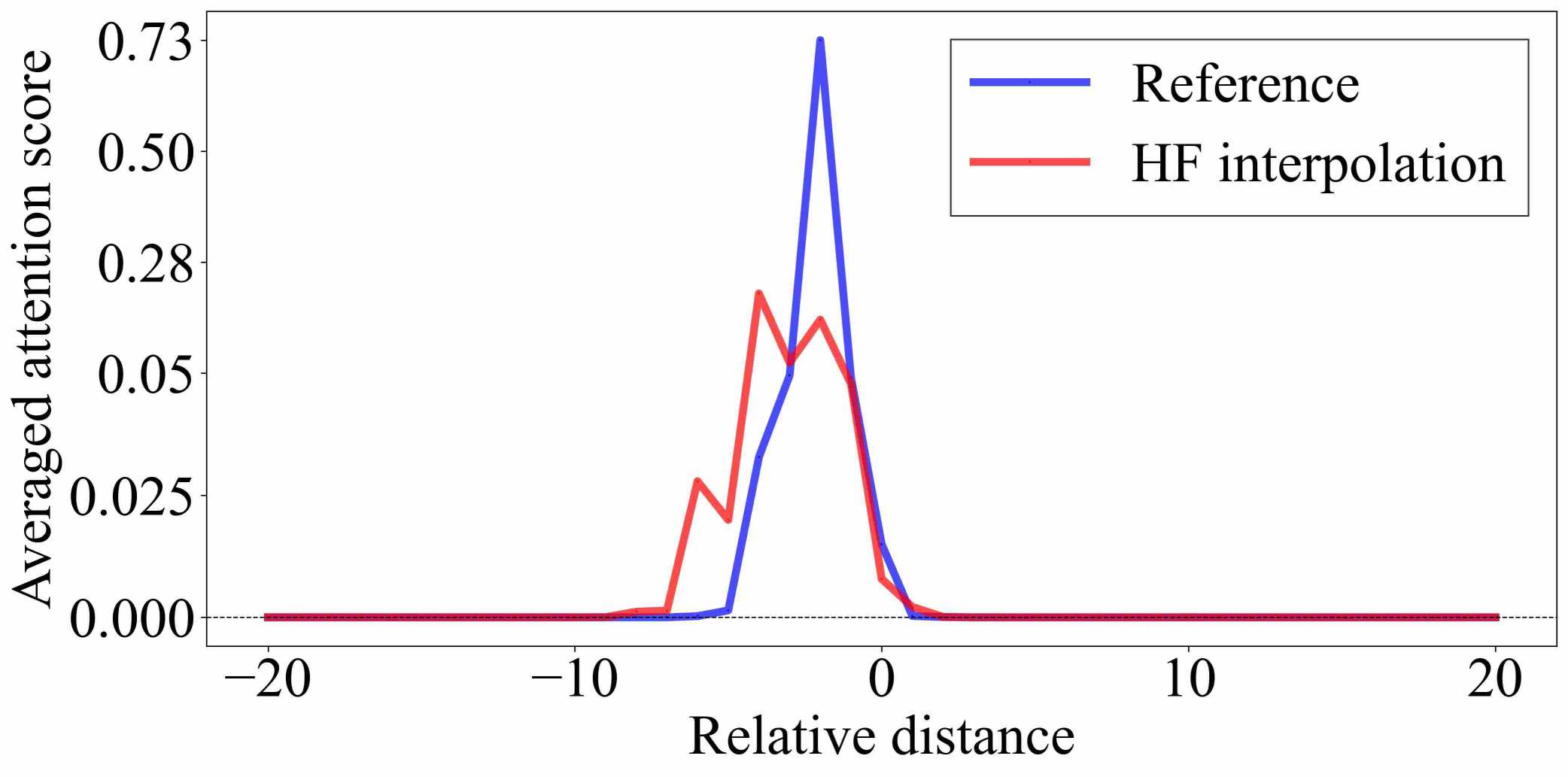} &
    \includegraphics[width=\linewidth,valign=m]{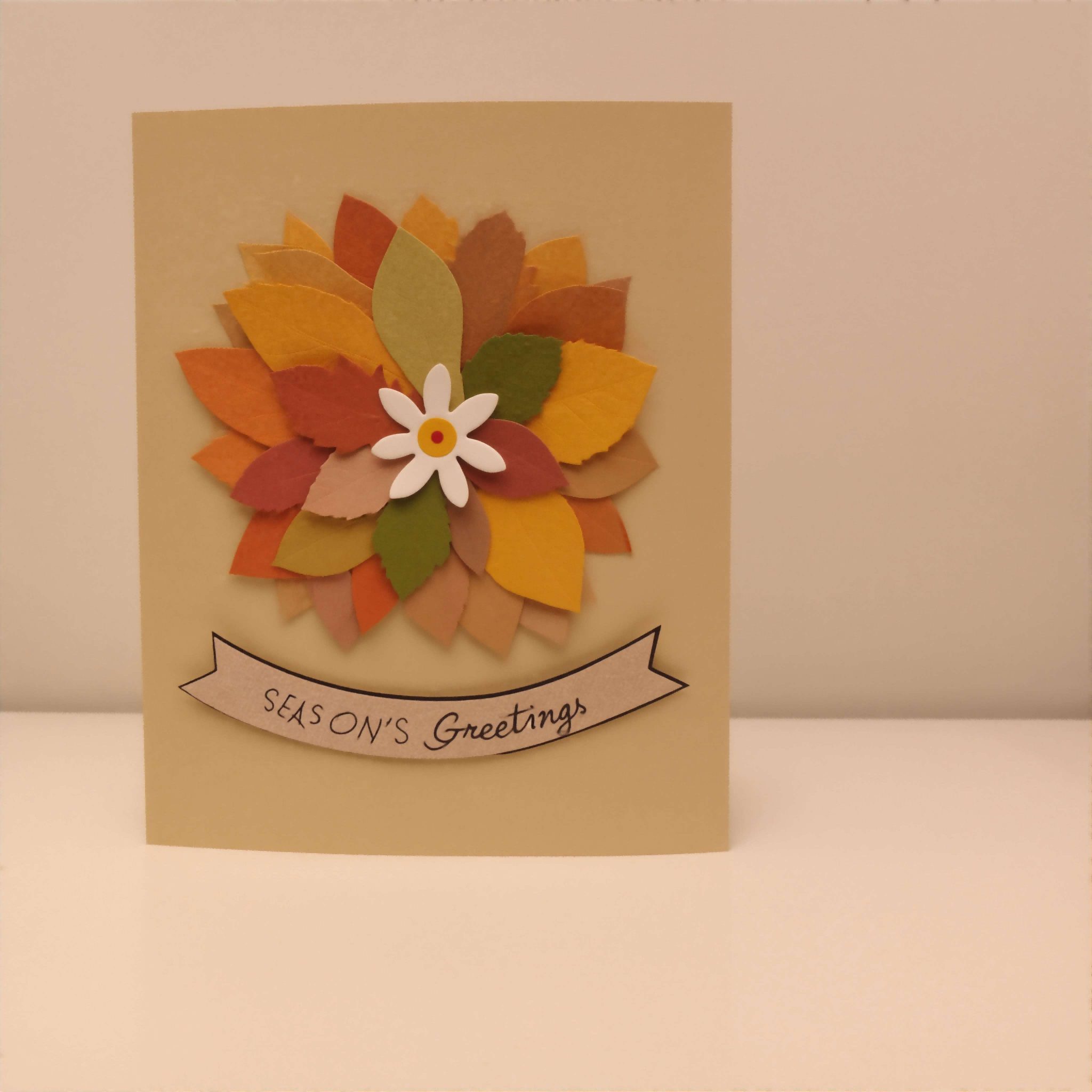} &
    \includegraphics[width=\linewidth,valign=m]{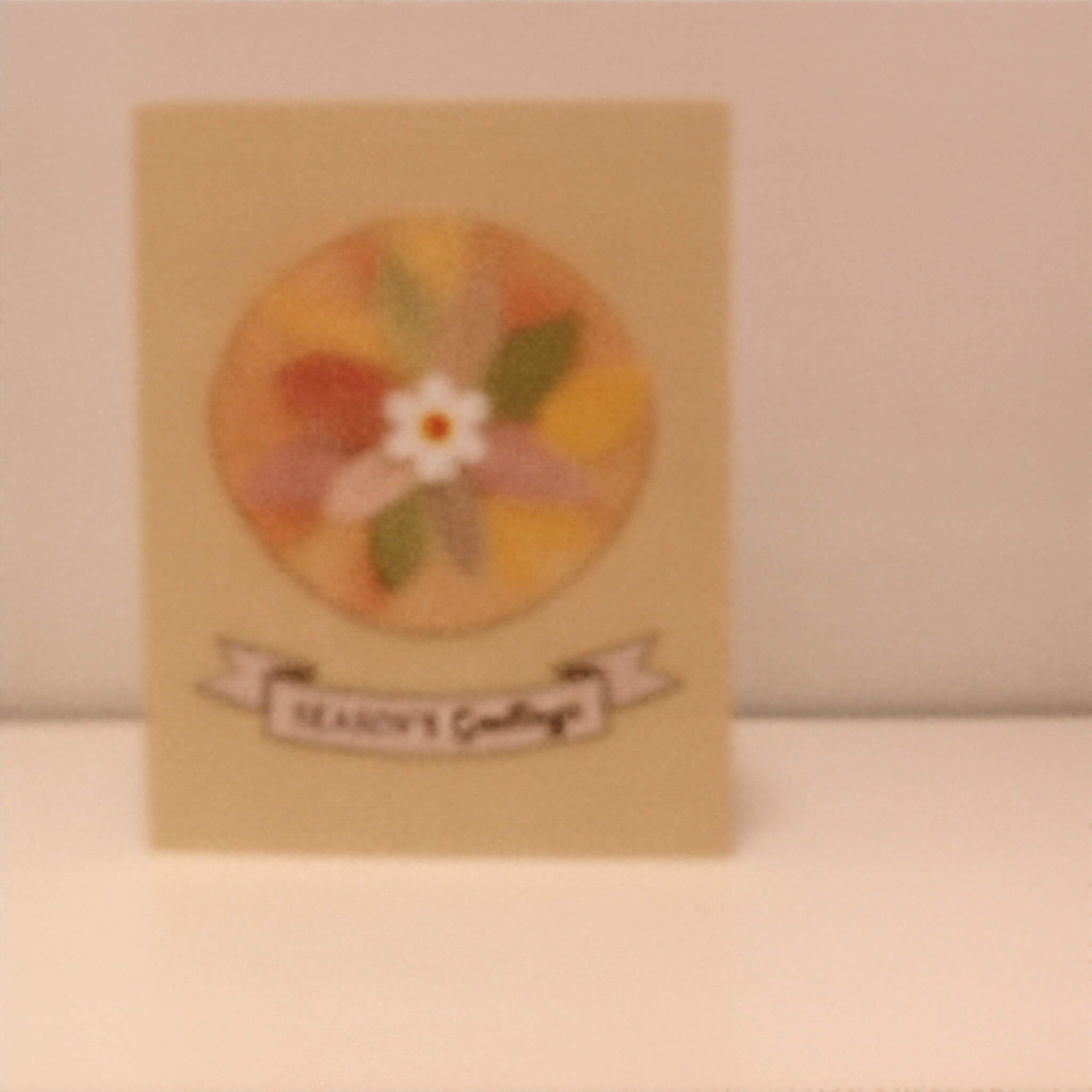} &
    \includegraphics[width=\linewidth,valign=m]{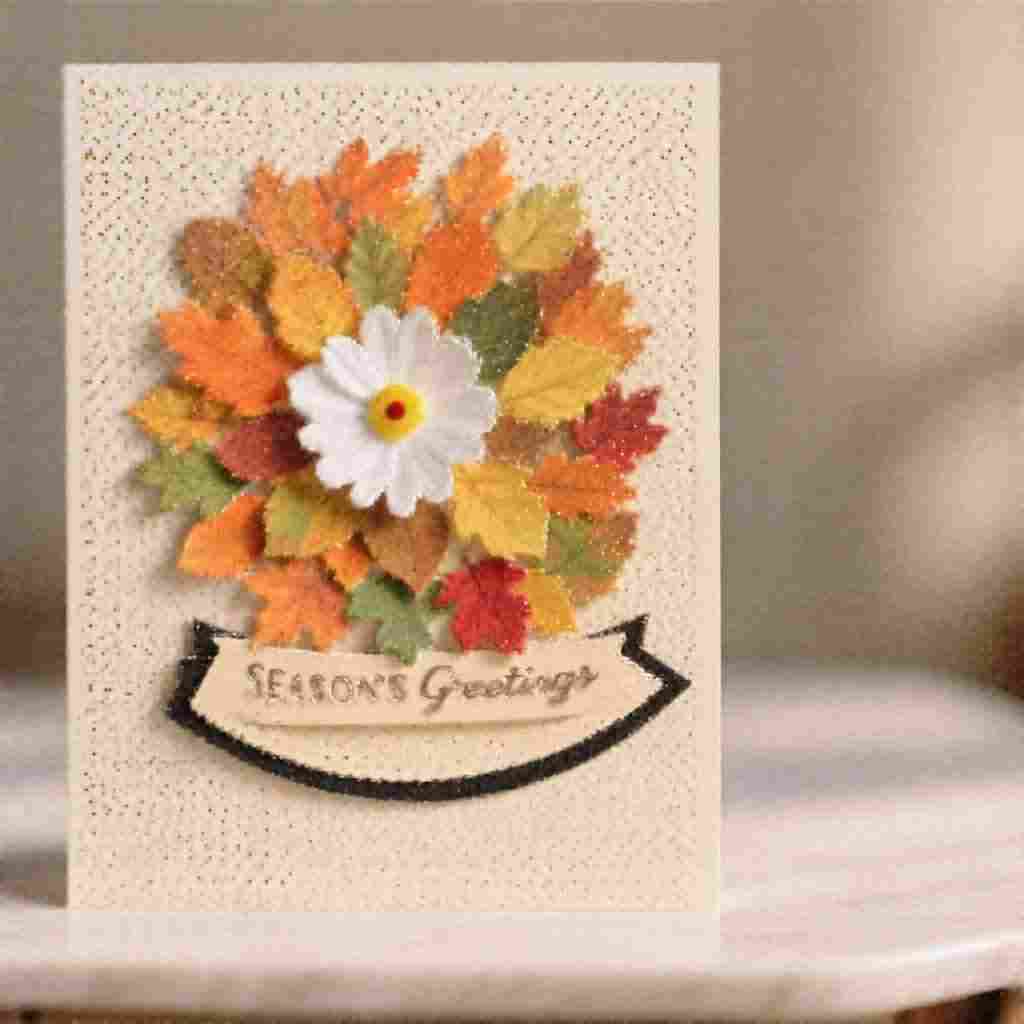} 
    \\
    \small{(a) } & \small{(b)} & \small{(c)}& \small{(d)}\\
    \bottomrule
    \end{tabular}
    \caption{\textbf{Cause of quality degradation.} Previous interpolation of a single high frequency (HF) in Fig.~\ref{fig:each frequency}b leads to similar quality loss during extrapolation. (a) Comparing attention maps (b) pre- and (c) post-interpolation, the distribution becomes significantly flattened.  Sharpening the attention in (d) restores the lost details, confirming that the degradation originates from reduced attention concentration.}
    
    \label{fig:cause}
\end{figure*}


We posit that these phenomena originate from a unified mechanism: the intrinsic periodicity of RoPE.
As introduced in Sec.~\ref{sec: background}, RoPE encodes spatial positions using periodic functions $\cos(\cdot)$ and $\sin(\cdot)$, where each frequency component~$i$ corresponds to an intrinsic period
\begin{equation}
\label{eq:period}
T_i^h = \frac{2\pi}{\theta_i^h}.
\end{equation}
When the relative positional distance exceeds this period, the encoding becomes ambiguous:
\begin{equation}
\label{eq:repeat}
\cos(\theta_i^h (p + T_i^h)) = \cos(\theta_i^h p), \quad
\sin(\theta_i^h (p + T_i^h)) = \sin(\theta_i^h p).
\end{equation}
Due to this periodicity, positions $p$ and $p + T_i^h$ share identical encodings, leading the model to treat them as equivalent and thus produce repeated content.
This interpretation aligns with our observations:
high-frequency components (with short periods, $T_i^h \ll h$) control local textures;
low-frequency components ($T_i^h \gg h$) remain unaffected;
and mid-band frequencies ($T_i^h \approx h$) govern global structure.
Consequently, the structure-level repetition reported in Sec.~\ref{sec: challenge} emerges once the extrapolated height~$H$ exceeds the intrinsic period of such a mid-band frequency.

Formally, we define the \textbf{dominant frequency} as the one whose period is closest to the training height~$h$:
\begin{equation}
\label{eq:dominant-frequency}
k_h = \arg\min_i |T_i^h - h|.
\end{equation}
To avoid repetition, the dominant frequency must remain within a single period during extrapolation, satisfying the following \textbf{non-repetition condition}:
\begin{equation}
\label{eq:non-repetition}
T_k'^h \geq H, \quad \theta_k'^h \leq \frac{2\pi}{H}.
\end{equation}
As illustrated in Fig.~\ref{fig:each frequency}e, settings where $H > T_k^h$ exhibit repetition, while $H \leq T_k^h$ avoid it (Fig.~\ref{fig:each frequency}f, g).
Under this perspective, NTK-based scaling fails to satisfy the non-repetition condition in Eq.~(\ref{eq:non-repetition}), thereby causing structure repetition; conversely, PI effectively amplifies high-frequency components, leading instead to blurred images.

\begin{algorithm}[b]
    \caption{Recursive Dominant Frequency Correction}
    \label{alg:msfi}
    \begin{algorithmic}[1]
        \Require Training resolution $(h, w)$, frequencies $\{\theta_i^h, \theta_i^w\}$ in RoPE, target resolution $(H, W)$
        \State Initialize observed repetition period $N_h \gets h$, $N_w \gets w$
        \While{$N_h < H$ or $N_w < W$}  \Comment{Repetition persists}
            \For{$i = 1$ to $\frac{d'}{2}$}
                \State $T_i^h = \frac{2\pi}{\theta_i^h}, \; T_i^w = \frac{2\pi}{\theta_i^w}$ \Comment{Compute period}
            \EndFor
            \State $k_h = \arg\min_i |T_i^h - N_h|,\;\; k_w = \arg\min_i |T_i^w - N_w|$ \Comment{Identify dominant frequency}
            \State $\theta_{k_h}^h \gets \frac{2\pi}{H},\;\; \theta_{k_w}^w \gets \frac{2\pi}{W}$  \Comment{Correct dominant frequency}
            \State Detect new repetition period $N_h, N_w$
        \EndWhile
    \end{algorithmic}
\end{algorithm}

\subsection{Recursive Dominant Frequency Correction}
\label{sec: repetition method}

Building on the above analysis, we directly modify the dominant frequencies for height and width as
\begin{equation}
\label{eq:new-theta}
\theta_{k}'^{h} = \frac{2\pi}{H}, \quad
\theta_{k}'^{w} = \frac{2\pi}{W}.
\end{equation}
Empirically, this modification largely mitigates structure-level repetition. However, in some models (e.g., Qwen-Image), we observe a residual, weaker repetition with a spatial period close to $T_k$. We hypothesize that this arises because dynamic-resolution training around $T_k$ introduces multiple frequency components that can align with the training length, yielding several competing dominant frequencies.

To address this, we propose a \emph{Recursive Dominant Frequency Correction} (RDFC), which iteratively adjusts the dominant frequency until the remaining repetition is eliminated and the dominant component is constrained within a single period after extrapolation. The complete procedure is summarized in Algorithm~\ref{alg:msfi}.

\section{Understanding and Solving Quality Degradation in Attention}
\label{sec:attention}

With the content repetition artifacts resolved, we now address the second key challenge identified in Sec.~\ref{sec: challenge}: quality degradation.

\subsection{Analyzing Quality Degradation in Attention}
\label{sec:attention_analysis}

Comparing the one-dimensional static attention scores before and after interpolation (Fig.~\ref{fig:cause}) reveals a noticeably flatter distribution. To quantify this effect, we introduce a global focus factor~$\lambda$ (an inverse temperature)~\cite{jin2023training} applied to the attention logits:
\begin{equation}
\label{eq:focus}
S'_{ij} = \lambda \cdot S_{ij},
\end{equation}
where $S_{ij}$ denotes the original attention score and $\lambda>1$. As shown in Fig.~\ref{fig:cause}, sharpening with a larger $\lambda$ restores fine details, confirming that the degradation stems from reduced attention concentration.

\begin{figure}[tbp]
    \centering
    \begin{tabular}{
        >{\centering\arraybackslash}m{0.13\textwidth} |
        >{\centering\arraybackslash}m{0.13\textwidth} |
        >{\centering\arraybackslash}m{0.13\textwidth}
    }
    \toprule
    \textbf{baseline} & $\lambda=1.1$ & $\lambda=1.2$\\ 
    \midrule
    \includegraphics[width=\linewidth,valign=m]{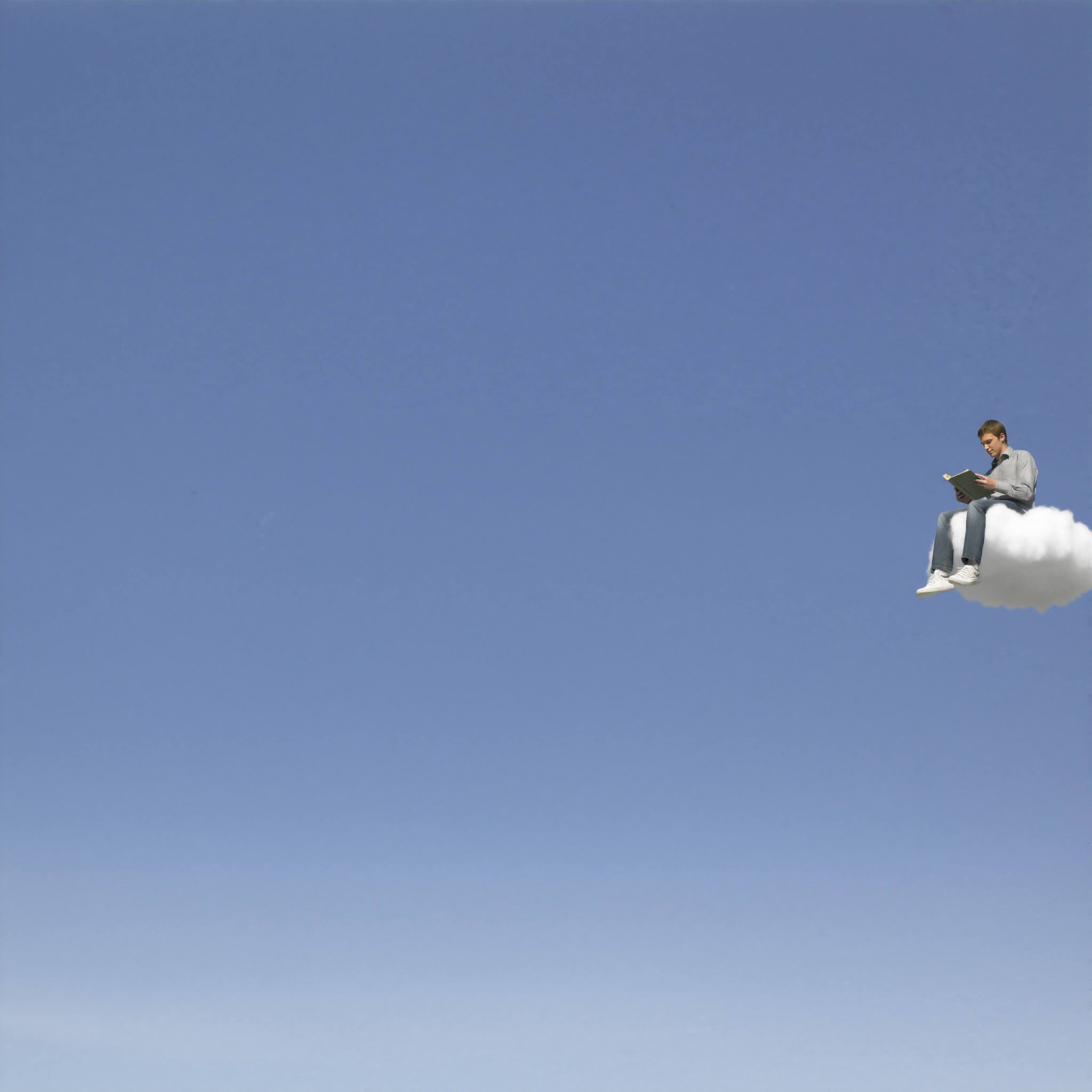} &
    \includegraphics[width=\linewidth,valign=m]{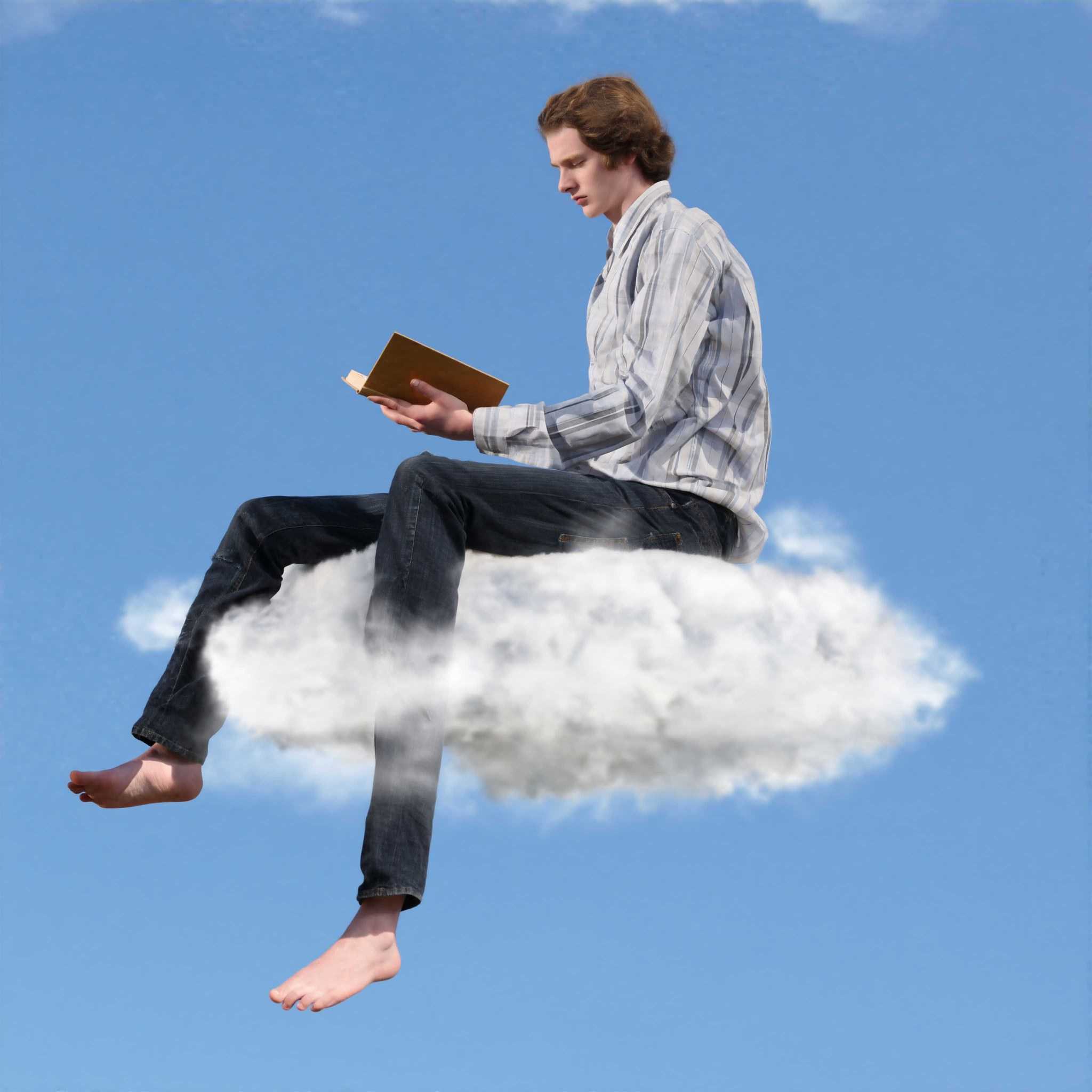} &
    \includegraphics[width=\linewidth,valign=m]{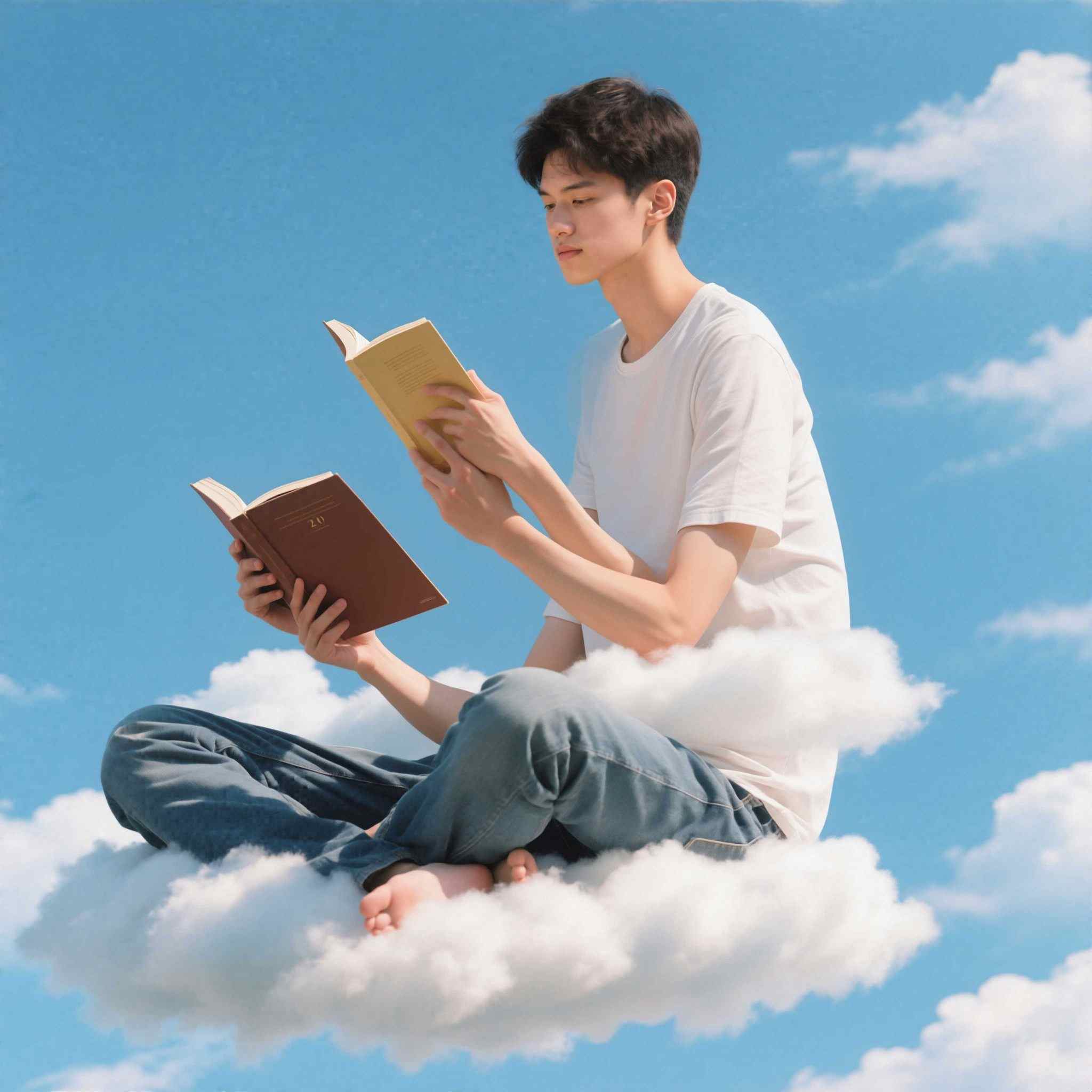} 
    \\
    \bottomrule
    \end{tabular}
    \caption{\textbf{Trade-off in attention concentration.} Increasing the focus factor $\lambda$ (e.g., 1.1) during extrapolation sharpens attention and enhances details, but further increase (e.g., 1.2) leads to spatial inconsistencies despite higher sharpness.}
    \label{fig:trade-off}
\end{figure}

We hypothesize that a similar flattening occurs during resolution extrapolation, where the expanded token field dilutes attention weights. Indeed, applying a focus factor $\lambda$ improves detail quality (Fig.~\ref{fig:trade-off}); however, excessive sharpening introduces new repetition artifacts and spatial inconsistency. This behavior can be explained by the attention mechanism in Eq.~(\ref{eq:attention}): a flat distribution averages diverse features, producing blur, while an overly sharp one restricts long-range dependencies, breaking spatial coherence.

\begin{figure}[tbp]
    \centering
    \resizebox{0.9\linewidth}{!}{%
    \begin{tabular}{
        >{\centering\arraybackslash}m{0.05\textwidth} |
        >{\centering\arraybackslash}m{0.17\textwidth} |
        >{\centering\arraybackslash}m{0.17\textwidth}
    }
    \toprule
    \textbf{type} & \textbf{generated image} & \textbf{attention pattern} \\ 
    \midrule
    \textbf{local} &
    \includegraphics[width=\linewidth]{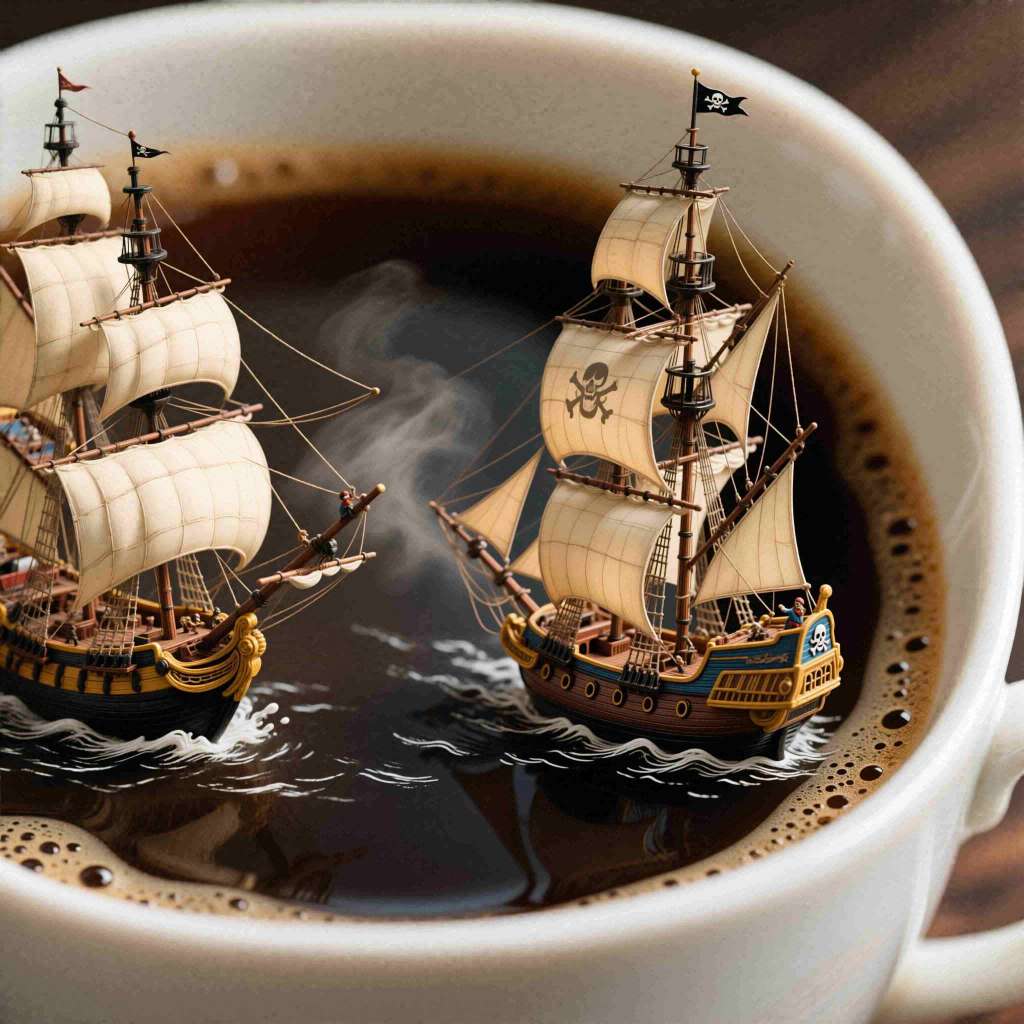} &
    \includegraphics[width=\linewidth]{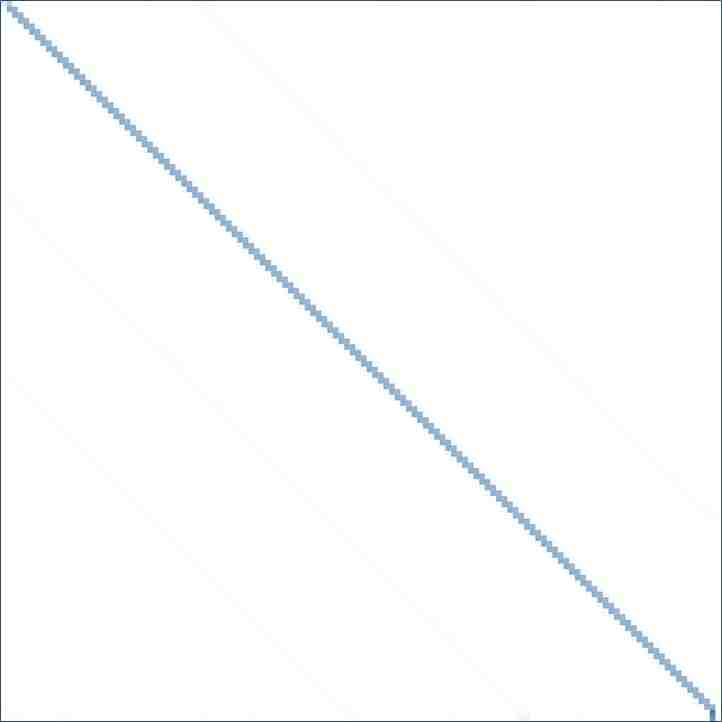} \\
    \midrule
    \textbf{global} &
    \includegraphics[width=\linewidth]{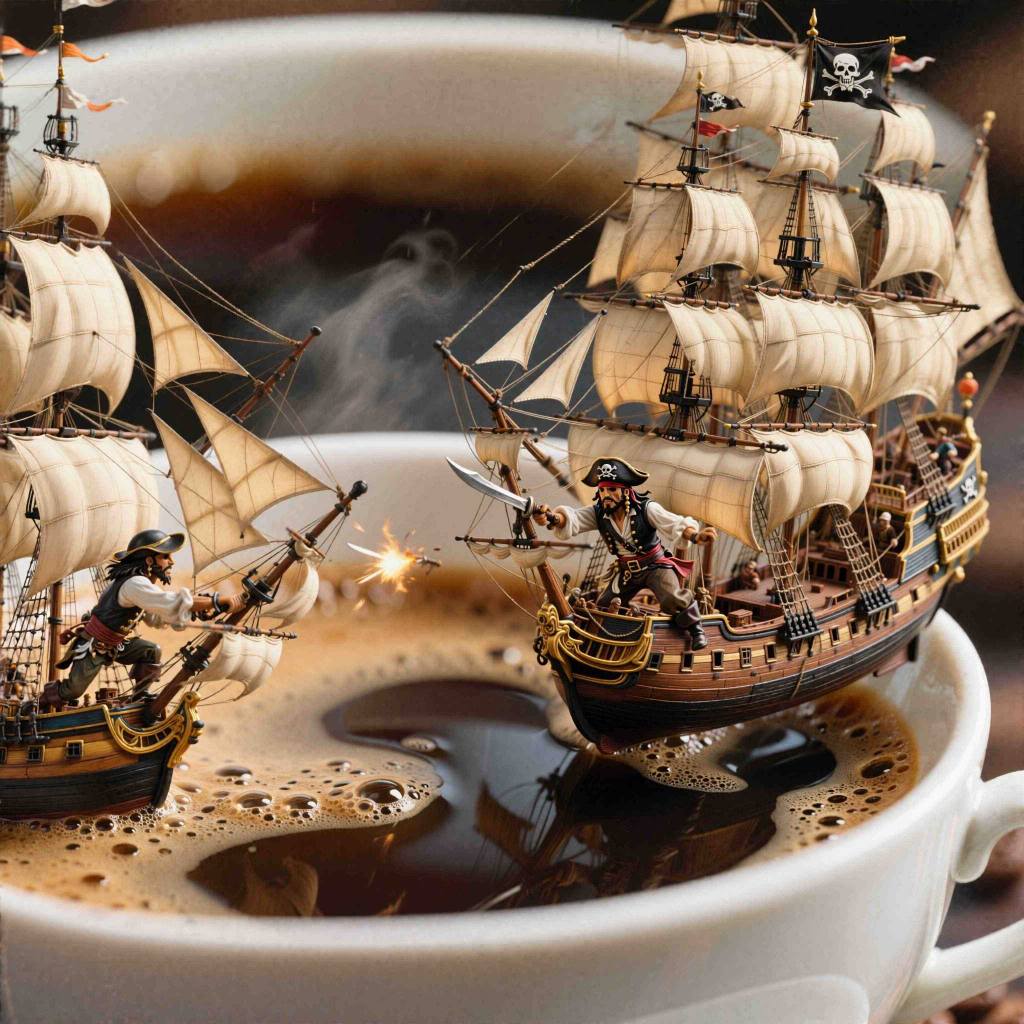} &
    \includegraphics[width=\linewidth]{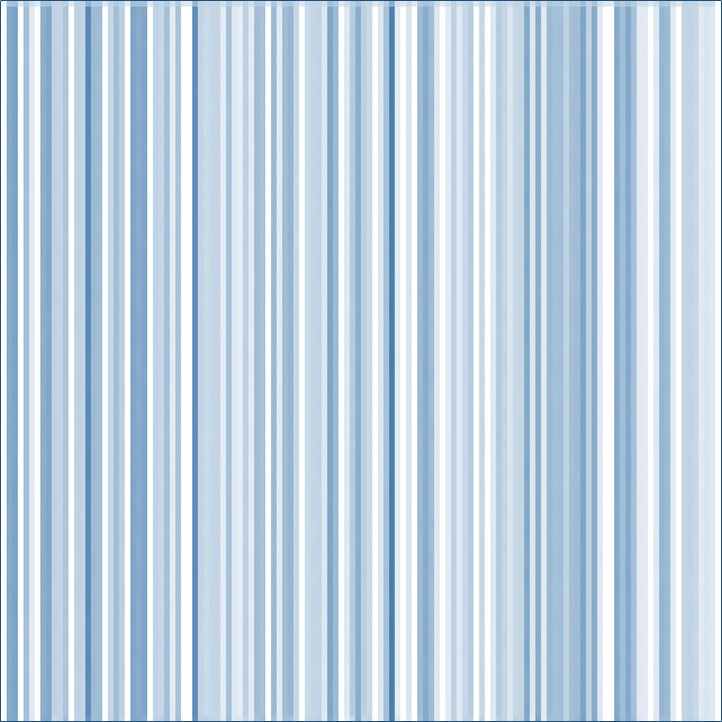} \\
    \bottomrule
    \end{tabular}
    }
\caption{\textbf{Effects of the focus factor on different attention patterns.} A single factor $\lambda$ may under-sharpen local attention (top) but over-suppress global attention (bottom), disrupting structural coherence and motivating adaptive focus for each pattern.}
    \label{fig:pattern}
\end{figure}

Visualizing multi-head attention further reveals distinct functional roles: some heads capture global structure with broad attention, while others focus on local textures. A single global $\lambda$ therefore affects them unevenly—values that benefit local attention pattern often over-concentrate global ones (Fig.~\ref{fig:pattern}). To address this, we require an adaptive focus factor $\lambda_\alpha$ for each attention head $\alpha$, \textbf{assigning smaller values to global attention patterns to preserve consistency and larger ones to local attention patterns to enhance details}.

\renewcommand\arraystretch{1.1}


\renewcommand\arraystretch{1.1}

\subsection{Entropy-guided Adaptive Attention Concentration}
\label{sec:attention_method}

\begin{algorithm}[t]
\small
    \caption{\small Entropy-Guided Adaptive Attention Concentration}
    \label{alg:adaptive_focus}
    \begin{algorithmic}[1]
        \Require Bounds $\lambda_{\min}, \lambda_{\max}$, exponent $p$. Matrices $Q, K, V \in \mathbb{R}^{N \times d}$, block size $b_q, b_{kv}$.
        
        \Statex \textbf{Stage 1: Entropy computation for the \underline{first denoising step}}
        \State Initialize $H_{\min}\gets+\infty$, $H_{\max}\gets-\infty$
            \For{each block-wise attention pattern $\alpha$}
                \State $H_\alpha$ = Eq.~\ref{eq:entropy} ($\text{softmax}( Q  K^{\top}/ \sqrt{d})$) ;
                
                \State $H_{\min}\gets\min(H_{\min},H_\alpha)$; 

                \State $H_{\max}\gets\max(H_{\max},H_\alpha)$; 

                \State Cache $H_\alpha$ ;
            \EndFor
            \For{each block-wise attention pattern $\alpha$}
                \State $\lambda_\alpha$ = Eq.~\ref{eq:adaptive_lambda} (cached $H_\alpha$, $\alpha$, $H_{\max}$, $H_{\min}$ )
                \State Cache $\lambda_\alpha$ for later use
            \EndFor

        \Statex
        \Statex \textbf{Stage 2: Head-wise adaptive concentration}
        \State Divide {$Q$} into $T_m = {N}/{b_q}$ blocks {$\{Q_i\}$}, and divide {$K$}, $V$ into $T_n = {N}/{b_{kv}}$ blocks {$\{K_i\}$} and $\{V_i\}$;
                
    \For {$\textbf{i}$ in [1, $T_m$]}
        \For {$\textbf{j}$ in [1, $T_n$]}
        
            \State $S_i^j = Q_i K_j^T$ ;

            \State $m_i^j = \mathrm{max}(m_i^{j-1}, \mathrm{rowmax}(S_i^j))$ ;
            
            \State $ \widetilde P_i^j = \mathrm{exp}(S_i^j - m_i^j)$ ;

            \For{each attention pattern $\alpha$ during generation}
            \State Retrieve cached $\lambda_\alpha$ ;
            \State Scale attention map:
            $\widetilde P_i^j = \mathrm{Softmax}(\lambda_\alpha \times \widetilde P_i^j)$ ;
            \EndFor
            
            \State $l_i^j = e^{m_i^{j-1}-m_i^j}\, l_i^{j-1} + \mathrm{rowsum}(\widetilde P_i^j)$ ;

            \State $O_i^j = \mathrm{diag}(e^{m_i^{j-1}-m_i^j})O_i^{j-1} + {\widetilde P_i^j} V_j$ ;
        \EndFor
        
        \State $O_i = \mathrm{diag}(l_i^{T_n})^{-1} O_i^{T_n}$ ;
         
    \EndFor
    
    \State \textbf{return} $O = \{O_i\}$;

    \end{algorithmic}
\end{algorithm}

The design principle introduces two challenges: (1) quantifying the dispersion of each attention pattern, and (2) mapping this metric to an appropriate focus factor~$\lambda_\alpha$.
We measure dispersion using the Shannon entropy~$H_\alpha$, which reflects how concentrated an attention distribution is—high entropy indicates a global pattern, while low entropy corresponds to a local one. Specifically, for the $\alpha$-th attention pattern:
\begin{equation}
\label{eq:entropy}
H_\alpha = -\frac{1}{HW} \sum_{i=1}^{HW} \sum_{j=1}^{HW} P^\alpha_{i,j} \log P^\alpha_{i,j},
\end{equation}
where $P^\alpha \in \mathbb{R}^{HW \times HW}$ is the attention map and $HW$ is the number of query tokens.

To assign focus factors, we construct a function that increases monotonically as entropy decreases, giving stronger sharpening to more concentrated patterns:
\begin{equation}
\label{eq:adaptive_lambda}
\lambda_\alpha = \lambda_{\min} + (\lambda_{\max} - \lambda_{\min})
\left(\frac{H_{\max} - H_\alpha}{H_{\max} - H_{\min}}\right)^p ,
\end{equation}
where $\lambda_{\min}$ and $\lambda_{\max}$ are the lower and upper bounds of scaling, and the exponent~$p$ controls the mapping curvature—smaller~$p$ yields more aggressive sharpening of low-entropy heads, while larger~$p$ provides smoother adjustment. See more details in Appendix.

Empirically, we find that the functional role of each attention head remains consistent across prompts and timesteps, allowing us to determine pattern types and corresponding focus factors from a single diffusion step (requiring only $\sim2\%$ of total inference cost for 50 steps).

\paragraph{Triton-based implementation.}
\label{sec:attention_impl}

The proposed method introduces two practical challenges:
(1) computing Eq.~(\ref{eq:entropy}) requires materializing the full attention map, which is memory-prohibitive—e.g., a $4096\times4096$ token input yields a $40\mathrm{K}\times40\mathrm{K}$ attention matrix, consuming over 80GB in \texttt{bf16};
(2) standard attention kernels do not support pattern-dependent modification of attention logits.

\renewcommand\arraystretch{1.1}
\begin{figure}[b]
    \centering
    \begin{tabular}{
        >{\centering\arraybackslash}m{0.13\textwidth} |
        >{\centering\arraybackslash}m{0.13\textwidth} |
        >{\centering\arraybackslash}m{0.13\textwidth}
    }
    \toprule
    \textbf{baseline} &\textbf{+RDFC} & \textbf{full version}\\ 
    \midrule
    \includegraphics[width=\linewidth,valign=m]{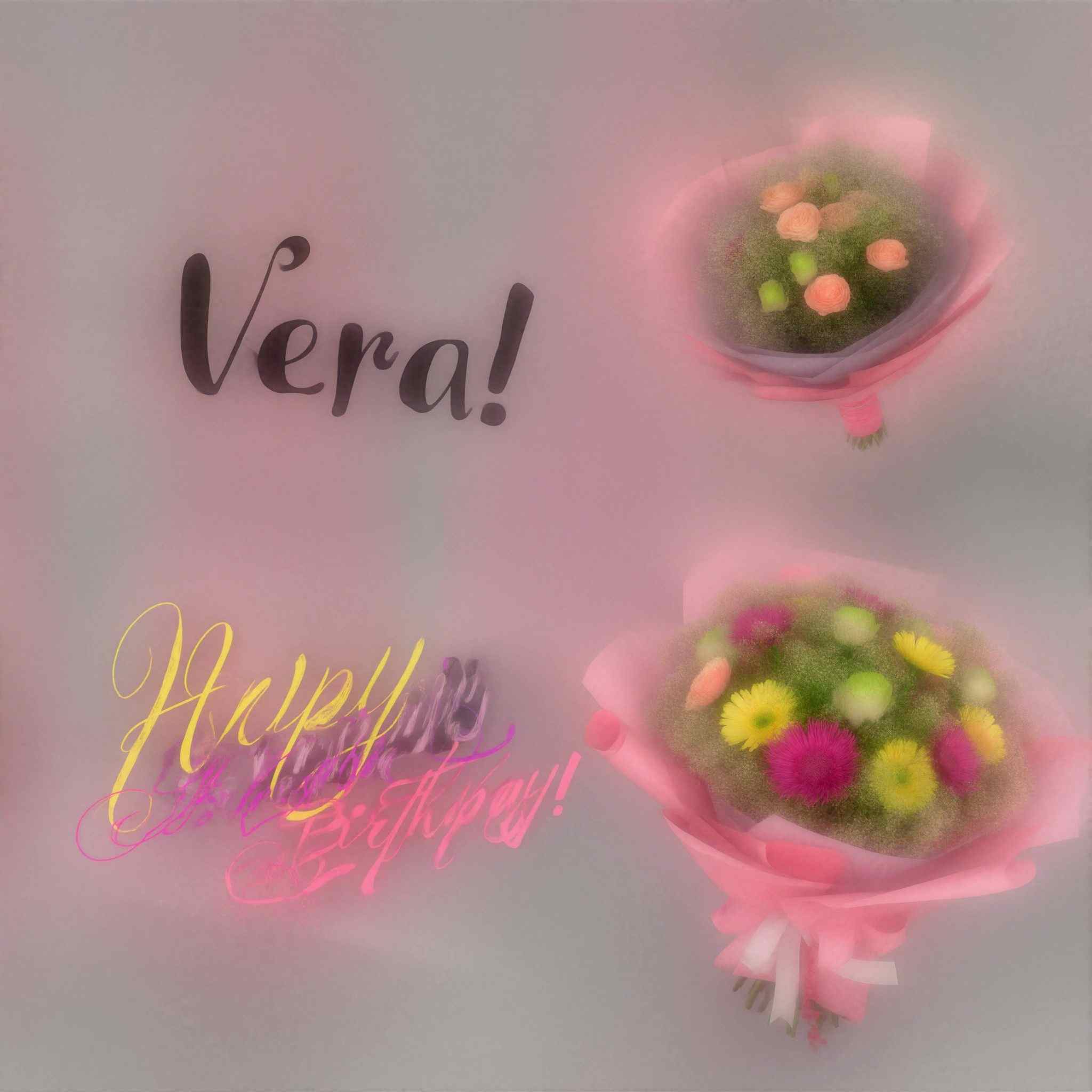} &
    \includegraphics[width=\linewidth,valign=m]{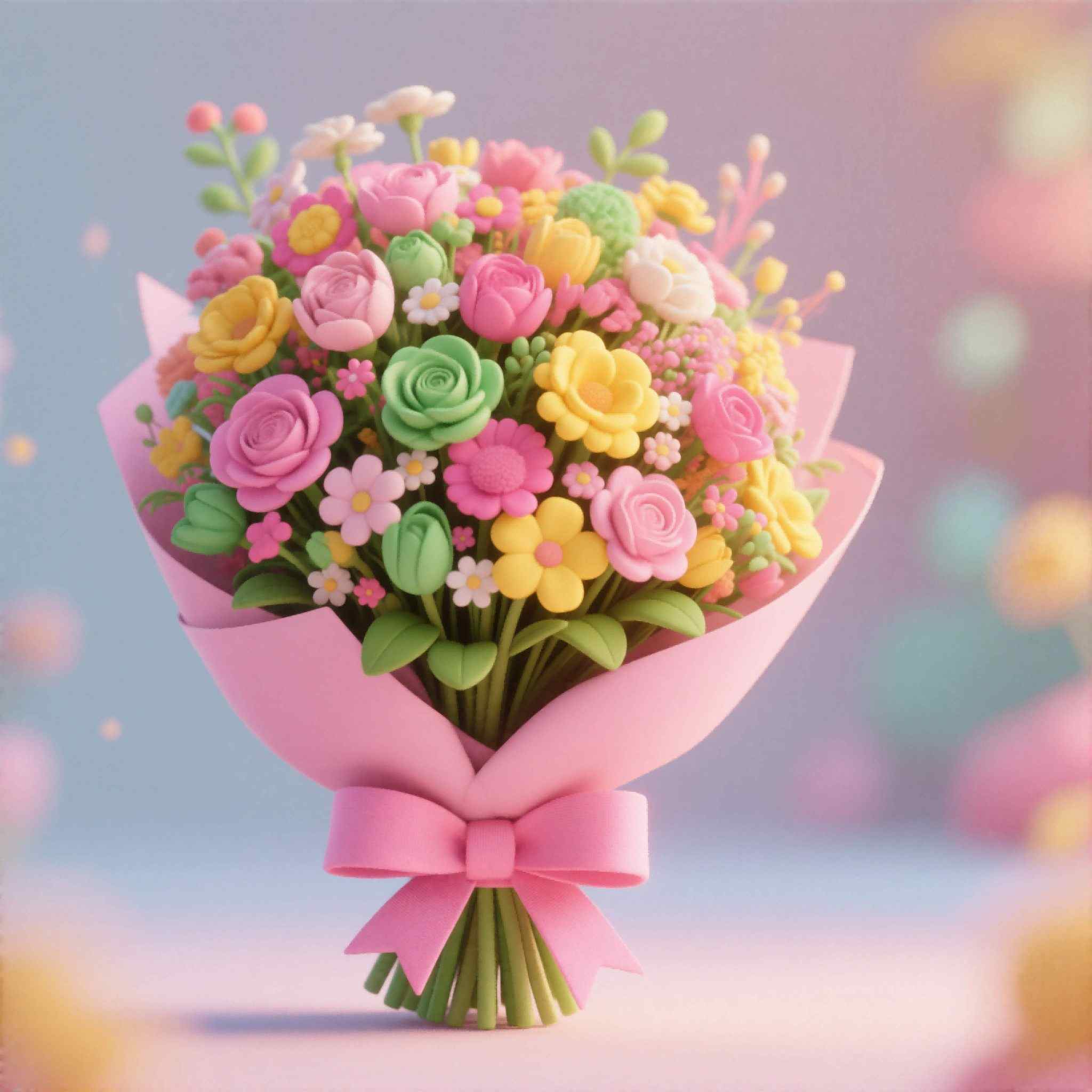} &
    \includegraphics[width=\linewidth,valign=m]{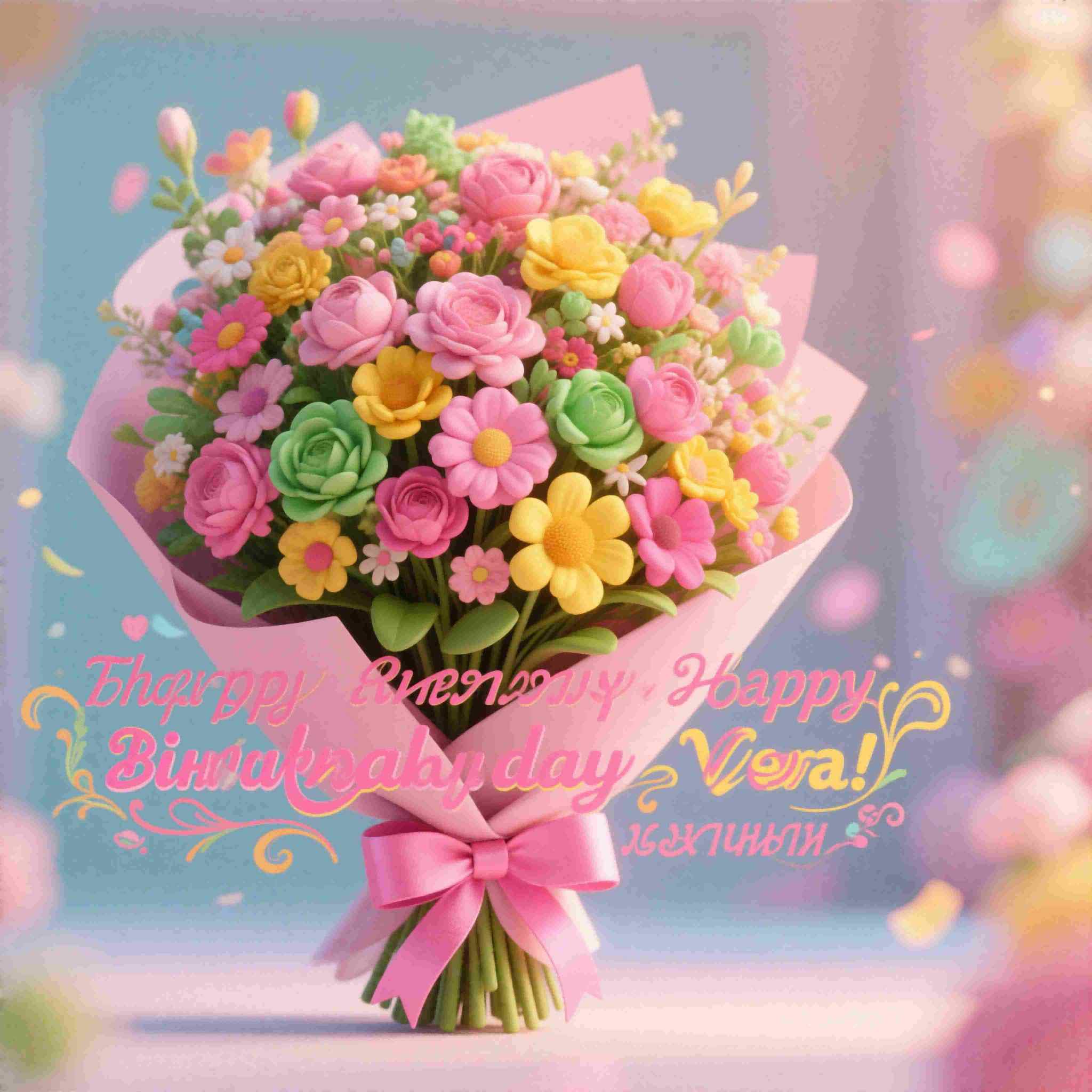} 
    \\
    \bottomrule
    \end{tabular}
    \caption{\textbf{Ablation studies.} Our RDFC effectively mitigates image repetition without compromising quality, while adaptive attention concentration further enhances visual fidelity. See the Appendix for quantitative ablation studies.
    }
    \label{fig:ablation}
\end{figure}

To address these two challenges, we develop a custom \emph{online softmax kernel in Triton}.
For the first challenge, we adopt a block-wise streaming strategy (e.g., block size $b_q=128, b_{kv}=128$) that computes attention scores incrementally without materializing the full attention matrix, ensuring memory efficiency. For the second challenge, our Triton kernel supports head-wise dynamic scaling, where each attention head retrieves its pre-computed focus factor~$\lambda_\alpha$ during computation. 
This implementation fully realizes adaptive concentration with negligible overhead. The full algorithm is shown in Algorithm~\ref{alg:adaptive_focus}.

\begin{table*}[t]
 \centering

\caption{\textbf{Quantitative comparison across three generation scenarios.} extra. denote extrapolation. All methods are evaluated at $4096\times4096$ on Qwen-Image (training resolution $1368\text{p}$) and $3600\times3600$ on Flux (training resolution ranging from $256^2$ to $2048^2$).}
 \label{tb:comparison}
 \renewcommand\arraystretch{1.0}
\resizebox{0.8\linewidth}{!}{
 \begin{tabular}{l@{\hskip 3em}ccc@{\hskip 2em}l@{\hskip 3em}ccc} 
 \toprule
 \multirow{2}{*}{\hspace{1cm}\vspace{-1mm}Method}
& \multicolumn{3}{c}{Direct extra. on Flux}
& \multirow{2}{*}{\hspace{1cm} \vspace{-1mm}Method} 
& \multicolumn{3}{c}{Direct extra. on Qwen-Image} \\
\cmidrule(lr){2-4} \cmidrule(lr){6-8} 
& \hspace{2mm}  \makecell{FID$\downarrow$} &  \hspace{0.5mm}\makecell{KID$\downarrow$} & \hspace{1mm}\makecell{CLIP Score$\uparrow$} 
& & \hspace{2mm}  \makecell{FID$\downarrow$} &  \hspace{0.5mm}\makecell{KID$\downarrow$} & \hspace{1mm}\makecell{CLIP Score$\uparrow$} \\ 
\midrule
    \hspace{1cm}PE& \hspace{2mm }206.2 & 0.1133 & 0.2280 & \hspace{1.1cm}PE &\hspace{2mm}\underline{86.93} & 0.0162 & 0.3257 \\
   \hspace{1cm}PI~\cite{chen2023extending}& \hspace{2mm }124.5 & 0.0391&  \underline{0.2789} & \hspace{1.1cm}PI~\cite{chen2023extending} &\hspace{2mm}94.03 & 0.0217 & 0.3310 \\
   \hspace{1cm}NTK~\cite{bloc97}& \hspace{2mm }196.6 &0.1055  & 0.2345 & \hspace{1.1cm}NTK~\cite{bloc97} &\hspace{2mm}86.94& \underline{0.0144} & 0.3246\\
   \hspace{1cm}YaRN~\cite{peng2023yarn}& \hspace{1.0mm } 186.2 & 0.0947& 0.2391  & \hspace{1.1cm}YaRN~\cite{peng2023yarn}&\hspace{2mm}96.47 & 0.0174 & 0.3304 \\
    \hspace{1cm}Entropy~\cite{jin2023training}& \hspace{2mm }\underline{117.3} &   \underline{0.0334} & 0.2708 & \hspace{1.1cm}Entropy~\cite{jin2023training} &\hspace{2mm}89.92& 0.0164 & \textbf{0.3399} \\
     \hspace{1cm}\textbf{Ours}& \hspace{2mm }\textbf{83.19} & \textbf{0.0114} & \textbf{0.3083} & \hspace{1.1cm}\textbf{Ours} &\hspace{2mm}\textbf{78.15} & \textbf{0.0086} & \underline{0.3337} \\
    \midrule
    \addlinespace[0.1em]
    \midrule
 \multirow{2}{*}{\hspace{1cm}\vspace{-1mm}Method}
& \multicolumn{3}{c}{Guided resolution extra. on Flux}
& \multirow{2}{*}{\hspace{1.1cm}\vspace{-1mm}Method}
& \multicolumn{3}{c}{Guided view extra. on Flux} \\
\cmidrule(lr){2-4} \cmidrule(lr){6-8}
& \hspace{2mm}  \makecell{FID$\downarrow$} &  \hspace{0.5mm}\makecell{KID$\downarrow$} & \hspace{1mm}\makecell{CLIP Score$\uparrow$} 
& & \hspace{2mm}  \makecell{FID$\downarrow$} &  \hspace{0.5mm}\makecell{KID$\downarrow$} & \hspace{1mm}\makecell{CLIP Score$\uparrow$} \\ 
\midrule
   \hspace{0.9cm} HiFlow~\cite{bu2025hiflow}& \hspace{2mm }73.13 &  0.0085 &  0.3375 & \hspace{1.1cm}NTK~\cite{bloc97} &\hspace{1mm} 118.6&0.0399 & 0.2291 \\
   \hspace{0.9cm} I-Max~\cite{du2024max}& \hspace{2mm }\underline{72.00} &\underline{0.0078} &  \underline{0.3392}& \hspace{1.1cm}YaRN~\cite{peng2023yarn} &\hspace{2mm}\underline{111.7} & \underline{0.0352}  & \underline{0.2339}\\   \hspace{0.9cm} \textbf{Ours}& \hspace{2mm }\textbf{68.98} & \textbf{0.0076} & \textbf{0.3417} & \hspace{1.1cm}\textbf{Ours} &\hspace{2mm}\textbf{104.7}& \textbf{0.0276}  & \textbf{0.2842}\\
    
\bottomrule
 \end{tabular}
}
\end{table*}

 \begin{figure*}
    \centering
    \includegraphics[width=1.7\columnwidth]{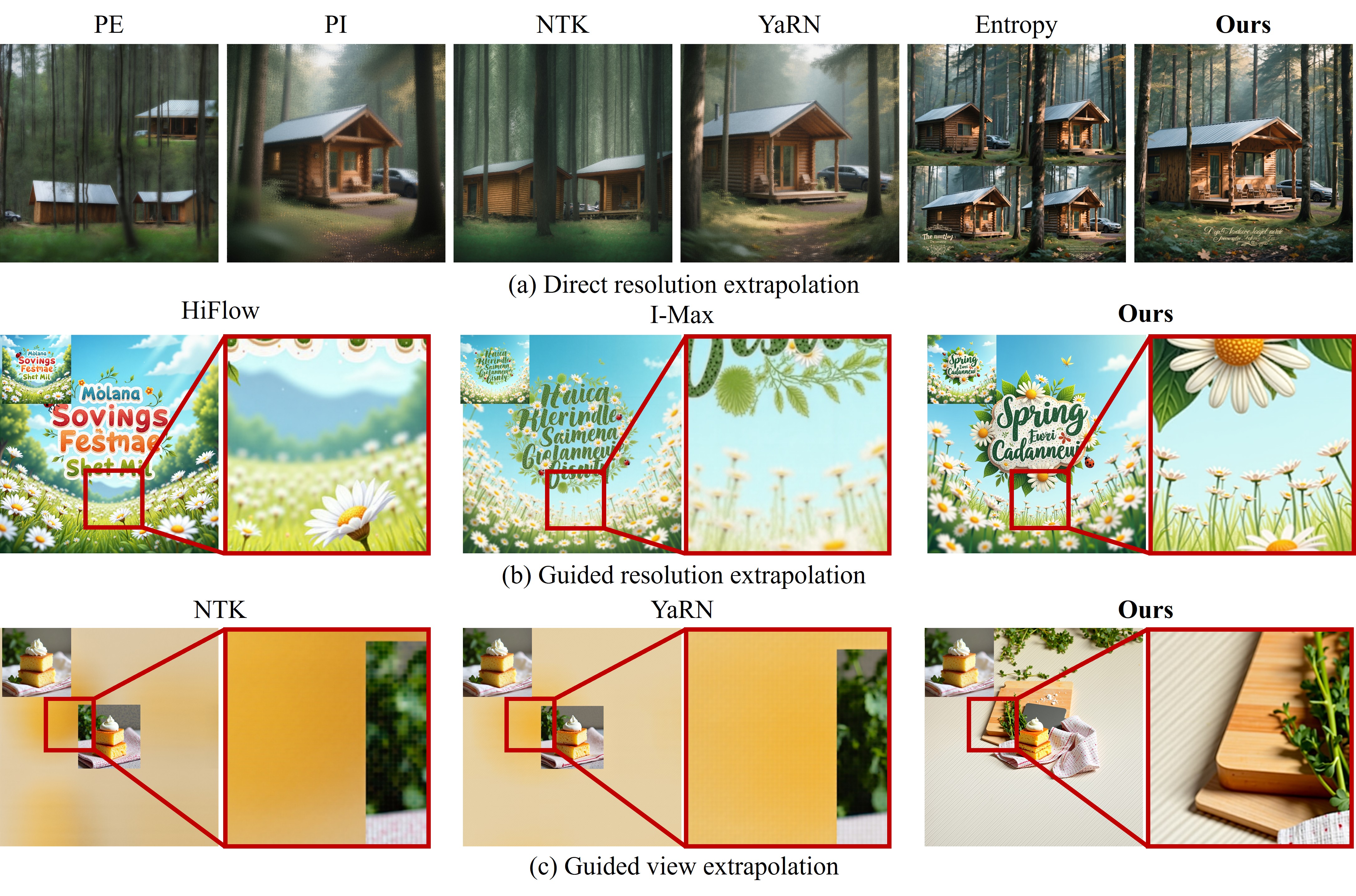}

\caption{\textbf{Qualitative comparison across three generation scenarios.} Our method outperforms baselines across all tasks by delivering high visual quality while mitigating content repetition.}    
    
    \label{fig: qualitive}
\end{figure*}

\section{Experiments}
\subsection{Setup}
\label{sec: setup}

\paragraph{Experiment Setup.}

We evaluate UltraImage under three generation scenarios. 
(1) \textbf{Direct resolution extrapolation} directly generates ultra-resolution images without any guidance, testing the model’s ability to extrapolate beyond the training resolution ($4096\times4096$ for Qwen, $3600\times3600$ for Flux). 
(2) \textbf{Guided resolution extrapolation} follows prior work~\cite{du2024max,bu2025hiflow}: we first generate images at the training resolution (1024$\times$1024), upsample them to the target resolution with 3600$\times$3600 on Flux), and use the upsampled image as low-resolution guidance to generate the final high-resolution output. 
(3) \textbf{Guided view extrapolation} generates a 1024$\times$1024 image at the training resolution and places it at the center of a target resolution of $3600\times3600$ on Flux, following the SDEdit~\cite{meng2021sdedit}, allowing the model to expand content around the central low-resolution image. See more experimental details in the Appendix. The dominant frequency $k=9$ for Flux and $k=8,9$ for Qwen-Image. $\lambda_{\min}=1.0, \lambda_{\max}=1.3, p=2$ for all models.

\paragraph{Evaluation.} 
To evaluate image generation, we randomly select 1K high-quality captions from LAION-5B covering diverse scenarios, following the general setup of prior work~\cite{he2023scalecrafter}. 
We conduct evaluations on Qwen-Image~\cite{wu2025qwen} and Flux models. 
Prompt-following ability is measured using the CLIP score~\cite{radford2021learning}, while overall image quality is assessed via Frechet Inception Distance (FID)~\cite{heusel2017gans} and Kernel Inception Distance (KID)~\cite{binkowski2018demystifying}. 
The generated images are compared against 10K high-quality real images from LAION-5B~\cite{schuhmann2022laion}.

\subsection{Results}

\paragraph{Main Results.}

Quantitative and qualitative comparisons are presented in Tab.~\ref{tb:comparison} and Fig.~\ref{fig: qualitive}. Our method consistently outperforms baselines across almost all metrics and tasks. In \emph{direct resolution extrapolation}, UltraImage achieves substantial FID improvements of $113.41$ over NTK and $41.31$ over PI on Flux, demonstrating notable gains in visual fidelity and a significant reduction of repetition artifacts. In \emph{guided resolution extrapolation}, where low-resolution images are upsampled and used as guidance, our approach effectively improves image quality, surpassing prior methods in FID, KID, and CLIP score. In \emph{guided view extrapolation}, which expands content around a central low-resolution image, UltraImage preserves structural consistency while recovering fine details, outperforming baselines across all metrics. Qualitative results in Fig.~\ref{fig: qualitive} further support these findings. Overall, these results confirm that UltraImage can generate ultra-resolution images that are both visually coherent and rich in detail, demonstrating its versatility across diverse resolution extrapolation scenarios.

\paragraph{Ablation Studies.} 
As shown in Fig.~\ref{fig:ablation}, compared with the direct extrapolation baseline, our Recursive Dominant Frequency Correction strategy effectively mitigates image repetition without sacrificing quality. 
Building upon this, the Entropy-Guided Adaptive Attention Concentration further enhances visual quality. 
Additional qualitative examples are presented in Fig.~\ref{fig: demo}. More ablation of the hyperparameters $\lambda_{\min}, \lambda_{\max}, p$ is provided in the Appendix.

\section{Conclusion}

We propose UltraImage, a principled framework for generating ultra-resolution images beyond the training scale. It tackles two key challenges in extrapolation: content repetition and quality degradation. UltraImage introduces a recursive dominant-frequency correction to eliminate repetition and an entropy-guided adaptive attention concentration to restore sharpness lost during extrapolation. Together, these components preserve global coherence and fine details, enabling high-fidelity generation at extreme resolutions.

\clearpage
{
    \small
    \bibliographystyle{ieeenat_fullname}
    \bibliography{main}
}

\setcounter{page}{1}
\maketitlesupplementary

\section{Related Work}

\subsection{Text-to-Image Generation}

Diffusion-based generative models have become the dominant paradigm for text-to-image synthesis due to their strong visual fidelity and semantic controllability. Early diffusion formulations such as DDPM~\cite{ho2020denoising} and Guided Diffusion~\cite{dhariwal2021diffusion} demonstrated that iterative denoising can produce high-quality images. The introduction of latent-space diffusion in LDM~\cite{rombach2022high} greatly improved efficiency by operating in a compressed representation space, leading to widely adopted systems including Stable Diffusion and SDXL~\cite{podell2023sdxl}. 

More recent approaches explore transformer-based diffusion architectures, which enhance global reasoning and scale more effectively than U-Net backbones. DiT~\cite{peebles2023scalable}, PixArt-$\alpha$/$\Sigma$~\cite{chen2023pixart,chen2024pixart}, and the Lumina series~\cite{gao2024lumina,zhuo2024lumina} exemplify this trend. Rectified-flow models such as Flux~\cite{flux2024} further refine the generative process through stable ODE-based formulations.
In addition to architectural advancements, a parallel line of work focuses on improving controllability through energy-based guidance. Energy-Guided SDEs (EGSDE)~\cite{zhao2022egsde} introduce an elegant framework for unpaired image-to-image translation by shaping the diffusion trajectory with learned energy functions.

In this work, we adopt Flux.1.0-dev~\cite{flux2024} as our text-to-image backbone due to its strong image quality and stable large-scale behavior.

\subsection{High-Resolution Image Generation}

Synthesizing images at resolutions far beyond the training scale remains a longstanding challenge for diffusion models, primarily due to the limited availability of high-resolution datasets and the significant computational demands of modeling large spatial grids. Existing research can be broadly divided into two categories.

\paragraph{Training or fine-tuning with high-resolution data.} 
A number of approaches~\cite{guo2024make,hoogeboom2023simple,liu2024linfusion,ren2024ultrapixel,teng2023relay,zheng2024any,zhang2025diffusion,yu2025urae} directly train or adapt diffusion models on higher-resolution datasets. While these methods can capture fine details, they require substantial computational resources and are limited by the scarcity of high-quality high-resolution data.

\paragraph{Training-free resolution expansion.} 
To bypass costly retraining, many works modify the inference process to scale resolution while keeping model weights unchanged. Patch-based fusion strategies such as MultiDiffusion~\cite{bar2023multidiffusion} and SyncDiffusion~\cite{lee2023syncdiffusion} combine overlapping denoising paths to enlarge the output canvas. DemoFusion~\cite{du2024demofusion} extends this idea with global layout perception.  
Other methods reshape or reinterpret intermediate features to align with larger spatial grids, as demonstrated in HiDiffusion~\cite{zhang2023hidiffusion} and I-Max~\cite{du2024max}. Receptive-field manipulation with dilated convolutions (ScaleCrafter~\cite{he2023scalecrafter}), frequency-domain alignment (FouriScale~\cite{huang2024fouriscale}), wavelet-based structural guidance~\cite{kim2024diffusehigh}, and training-free detail enhancement~\cite{qiu2024freescale,jin2024training,shi2024resmaster} offer additional routes.

\begin{table}[h]
\centering
\caption{\textbf{Quantitative ablation studies.} }
\label{tb:ablation}

\renewcommand\arraystretch{1.15}

\begin{tabular}{lccc}
\toprule
Method & FID$\downarrow$ & KID$\downarrow$ & CLIP$\uparrow$ \\
\midrule
baseline & {206.2} & {0.1133} & {0.2280} \\
+RDFC & \underline{107.81} & \underline{0.0257} & \underline{0.2829} \\
full version & \textbf{83.19} & \textbf{0.0114} & \textbf{0.3083} \\
\bottomrule
\end{tabular}

\end{table}

\section{Experiment Setup}
\label{appendix:experiments}

For the guided resolution extrapolation setting, building upon I-max, we first generate a $1024\times1024$ image and then upsample it to the target high resolution, using the upsampled result as guidance during high-resolution generation. For the guided view extrapolation setting, we first generate a $1024\times1024$ low-resolution image. During high-resolution generation, we replace the central region of the high-resolution $x_0$-prediction with noisy versions of the low-resolution image obtained via forward diffusion, ensuring consistent view alignment across scales.

\section{More Results}
\label{appendix:results}
\begin{figure*}
    \centering
    \includegraphics[width=1.8\columnwidth]{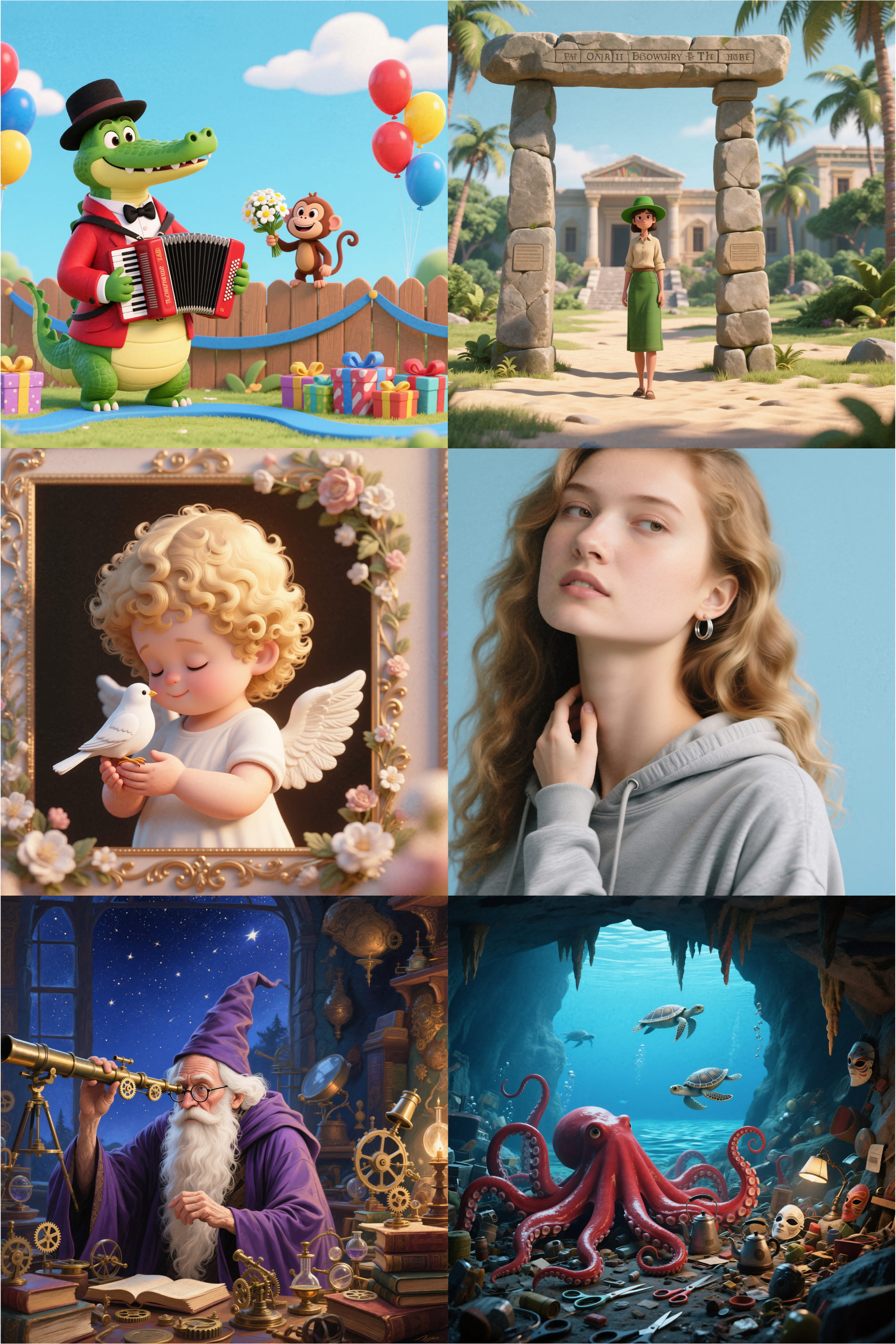}
\caption{\textbf{Results of UltraImage at $4096\times4096$}, where training resolution is 1328p.}
    \label{fig: app_demo}
\end{figure*}

As shown in Fig.~\ref{fig: app_demo}, we provide additional UltraImage samples generated without any low-resolution guidance. Despite being trained only at $1328p$, UltraImage can produce much higher-resolution images on Qwen-Image, demonstrating its strong capability to extrapolate far beyond the training resolution.

\section{Ablation Studies}
\label{appendix:ablation}

 \begin{figure}[h]
    \centering
    \includegraphics[width=\columnwidth]{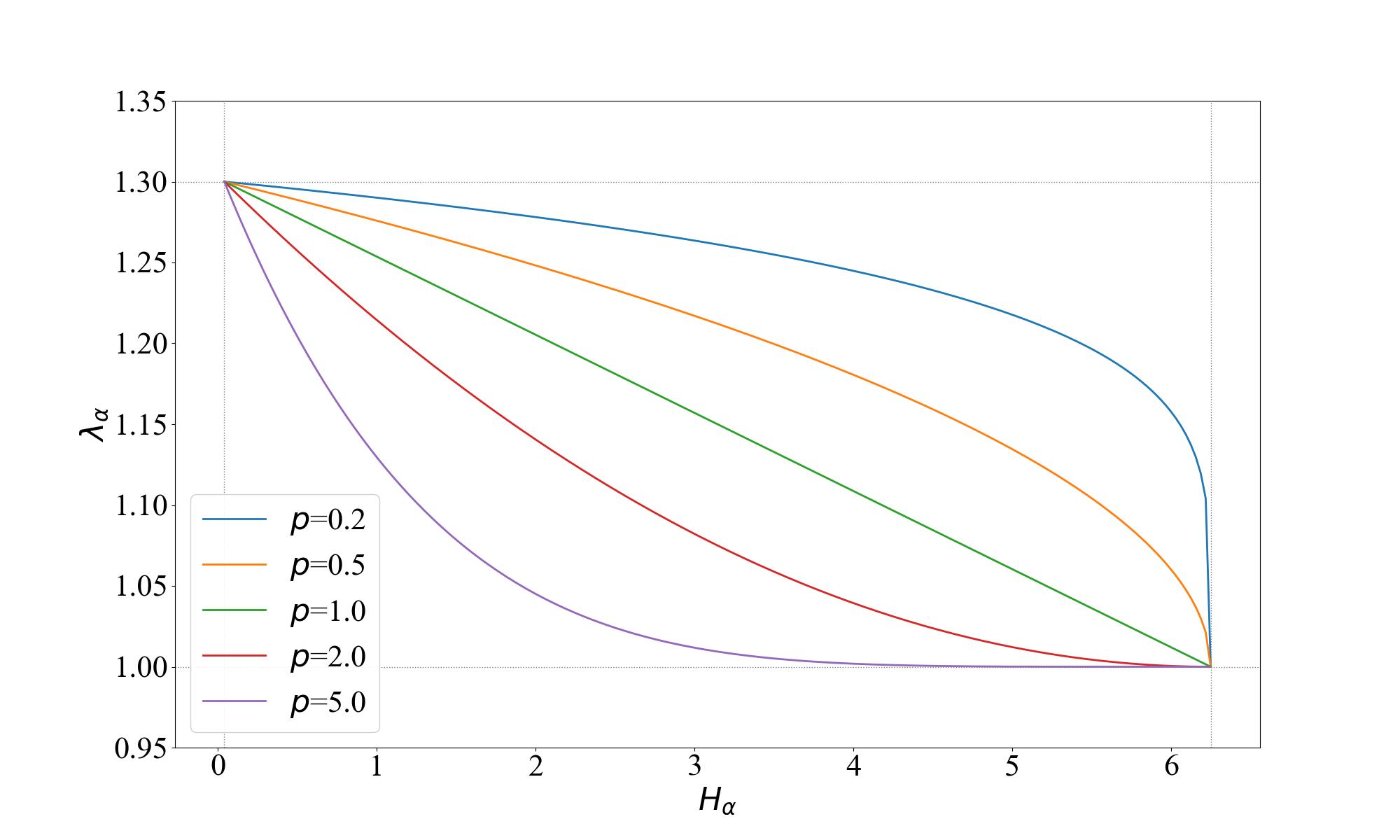}
\caption{\textbf{Visualization of $\lambda$ with different $p$ in Eq.~(11).}, with $\lambda_{\min}=1.0$ and $\lambda_{\max}=1.3$. Smaller $p$ produces stronger and broader attention focusing, while larger $p$ results in weaker concentration.}
    \label{fig:app_visual_p}
\end{figure}

\begin{figure*}[tbp]
    \centering
    \begin{tabular}{
        >{\centering\arraybackslash}m{0.25\textwidth} |
        >{\centering\arraybackslash}m{0.25\textwidth} |
        >{\centering\arraybackslash}m{0.25\textwidth}
    }
    \toprule
    $p=5.0$ & $p=2.0$ & $p=0.2$\\ 
    \midrule
    \includegraphics[width=\linewidth,valign=m]{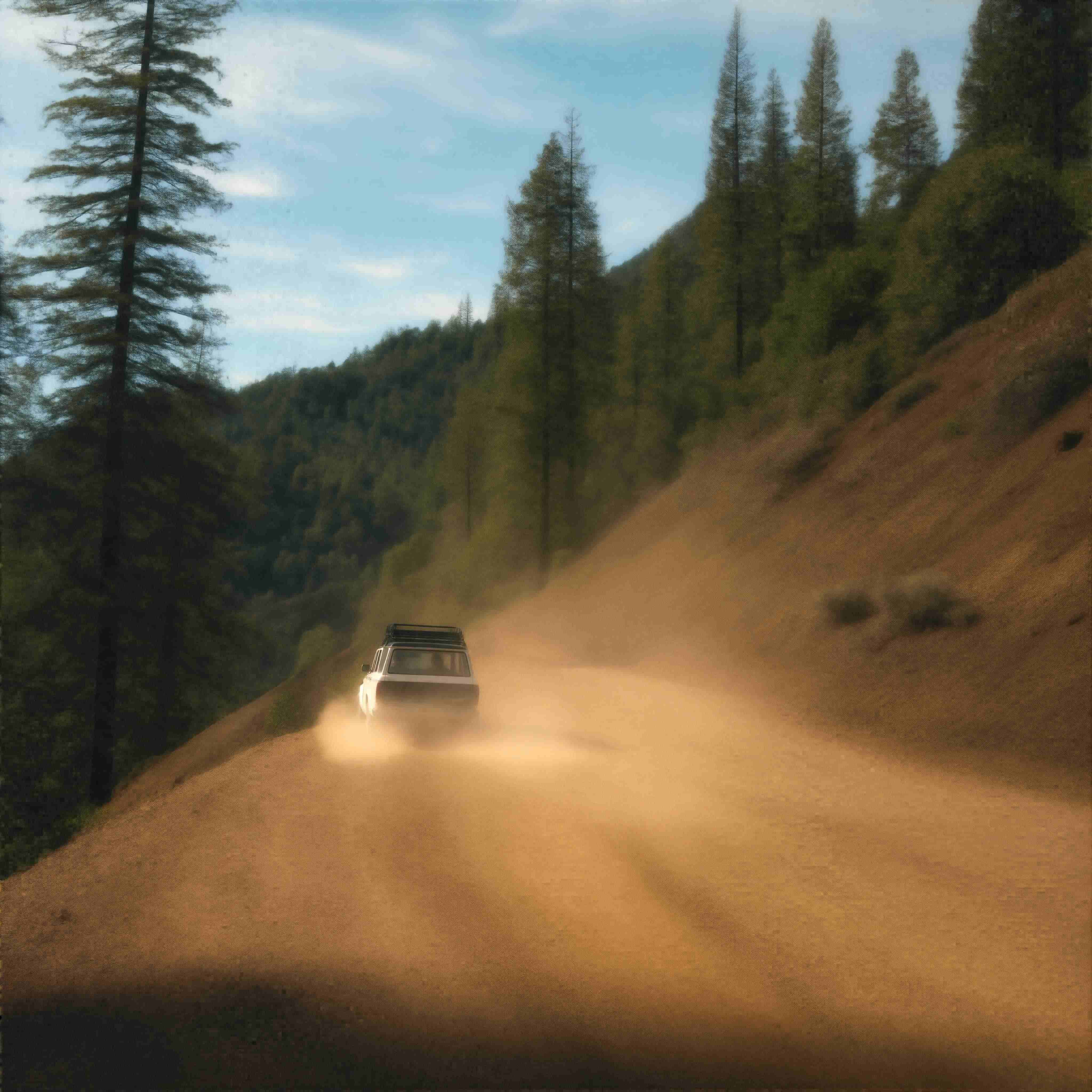} &
    \includegraphics[width=\linewidth,valign=m]{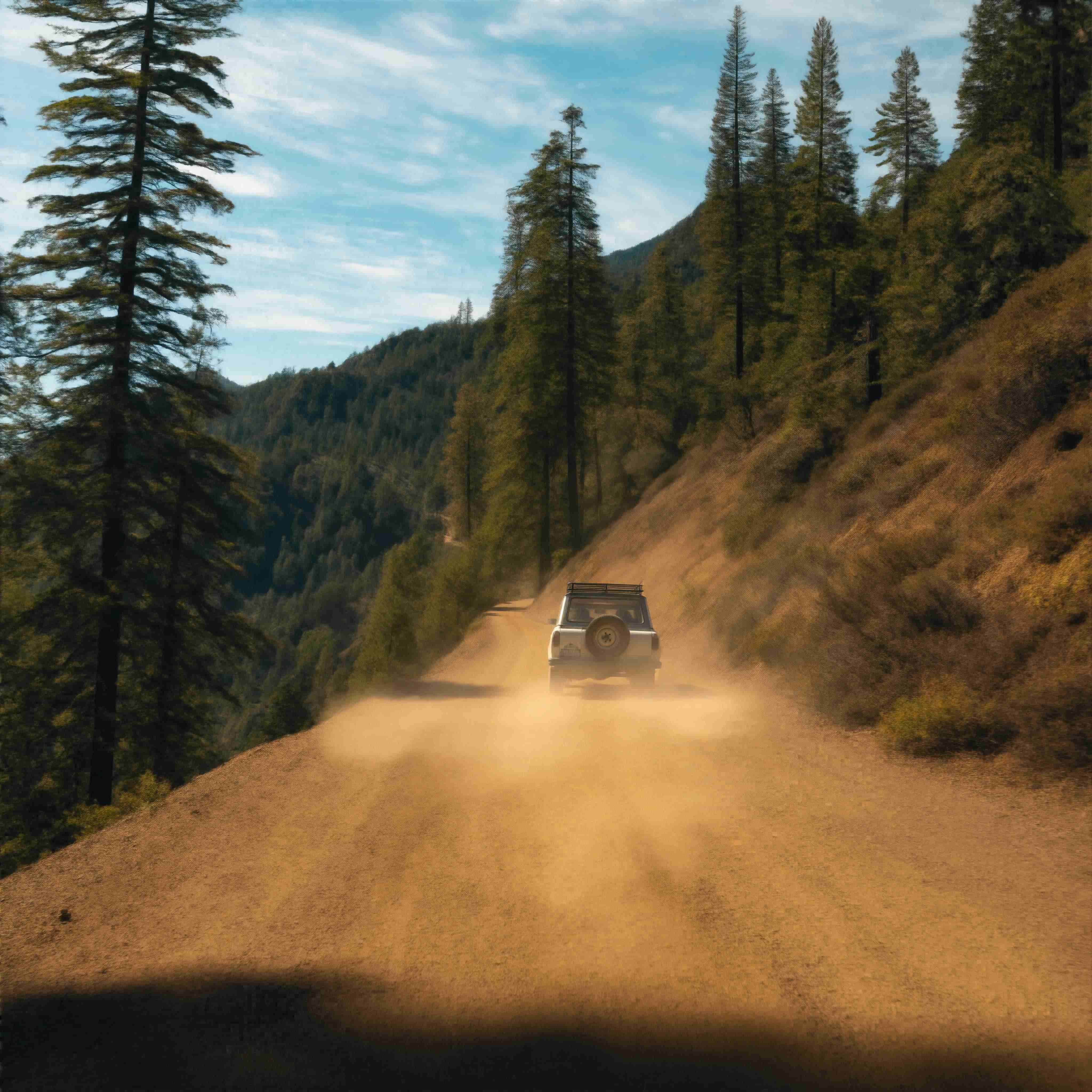} &
    \includegraphics[width=\linewidth,valign=m]{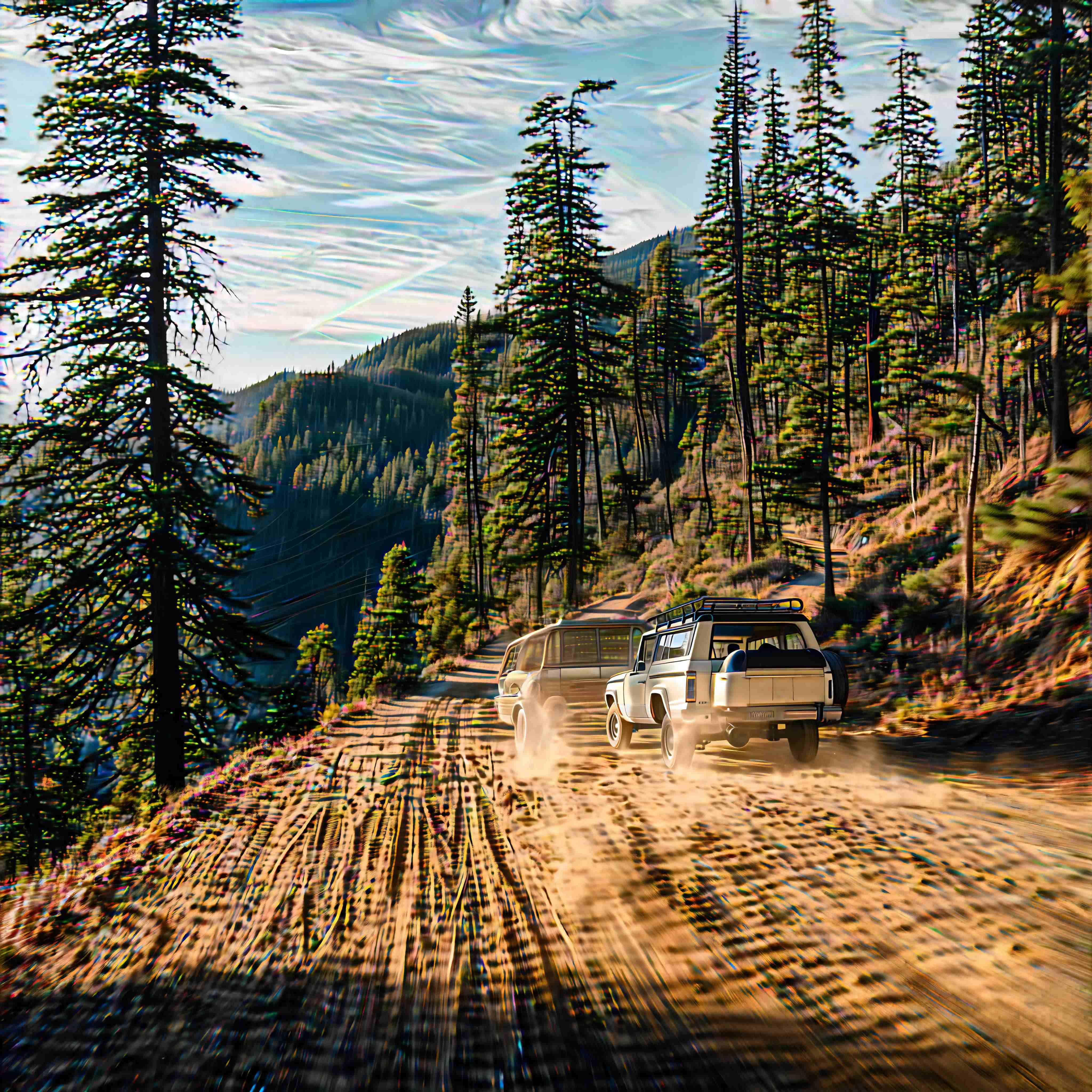} 
    \\
    \bottomrule
    \end{tabular}
   \caption{\textbf{Ablation on $p$.} Small $p$ (e.g., $0.2$) causes overly concentrated attention and over-sharpening, while large $p$ (e.g., $5.0$) leads to insufficient focus and degraded visual quality. An intermediate value yields the best trade-off.}
    \label{fig:app_p}
\end{figure*}

\begin{figure*}[tbp]
    \centering
    \begin{tabular}{
        >{\centering\arraybackslash}m{0.25\textwidth} |
        >{\centering\arraybackslash}m{0.25\textwidth} |
        >{\centering\arraybackslash}m{0.25\textwidth}
    }
    \toprule
    $\lambda_{\max}=1.1$ & $\lambda_{\max}=1.3$ & $\lambda_{\max}=1.6$\\ 
    \midrule
    \includegraphics[width=\linewidth,valign=m]{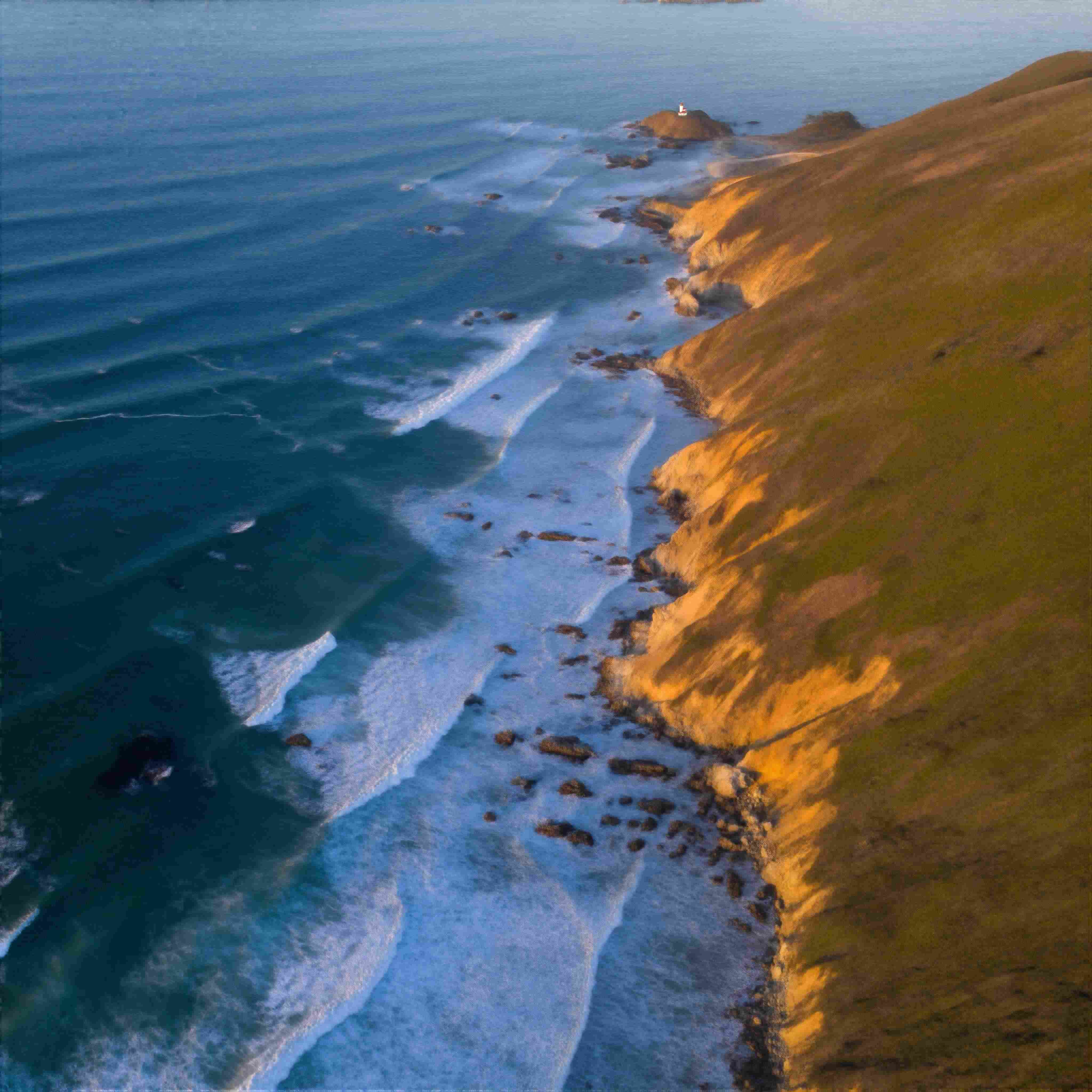} &
    \includegraphics[width=\linewidth,valign=m]{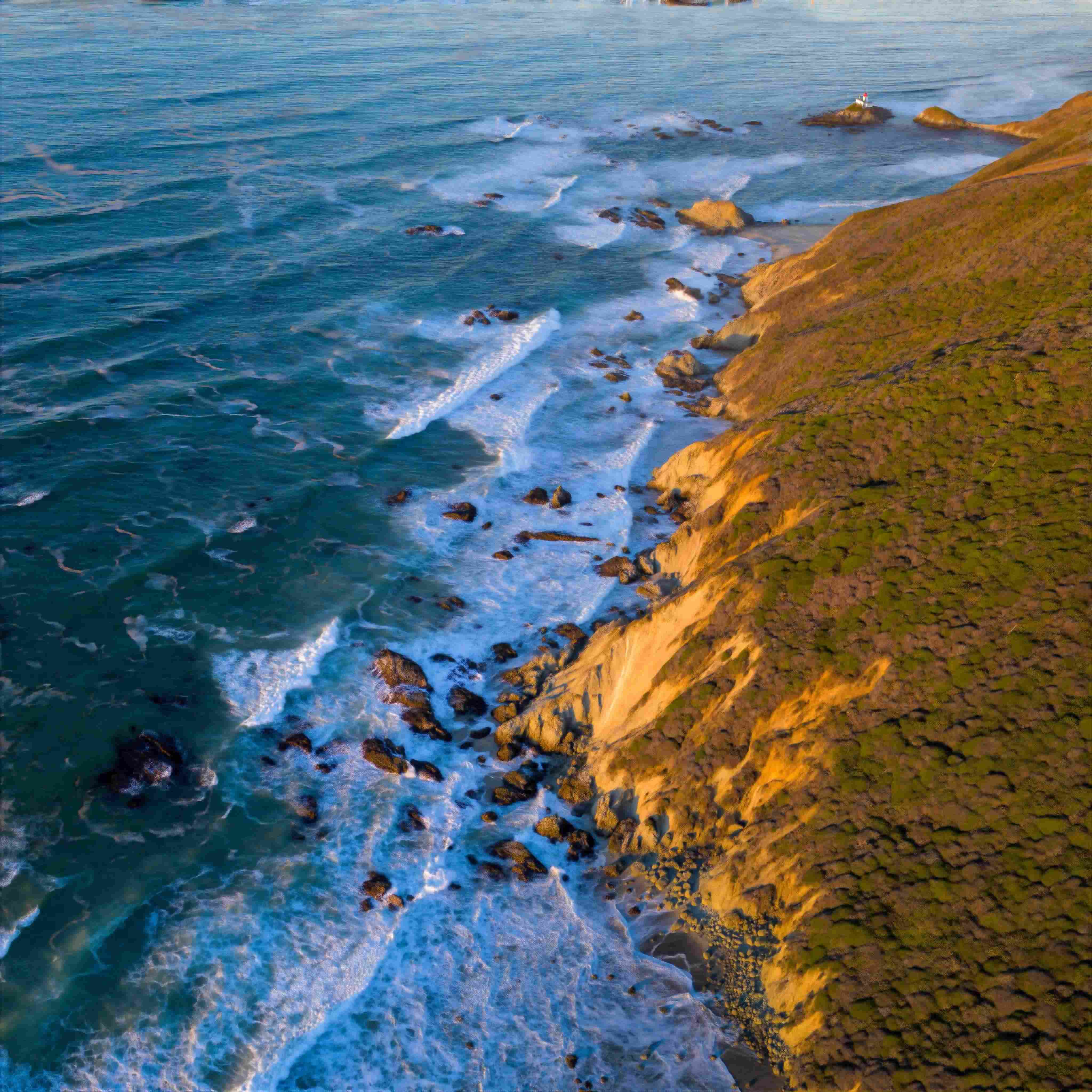} &
    \includegraphics[width=\linewidth,valign=m]{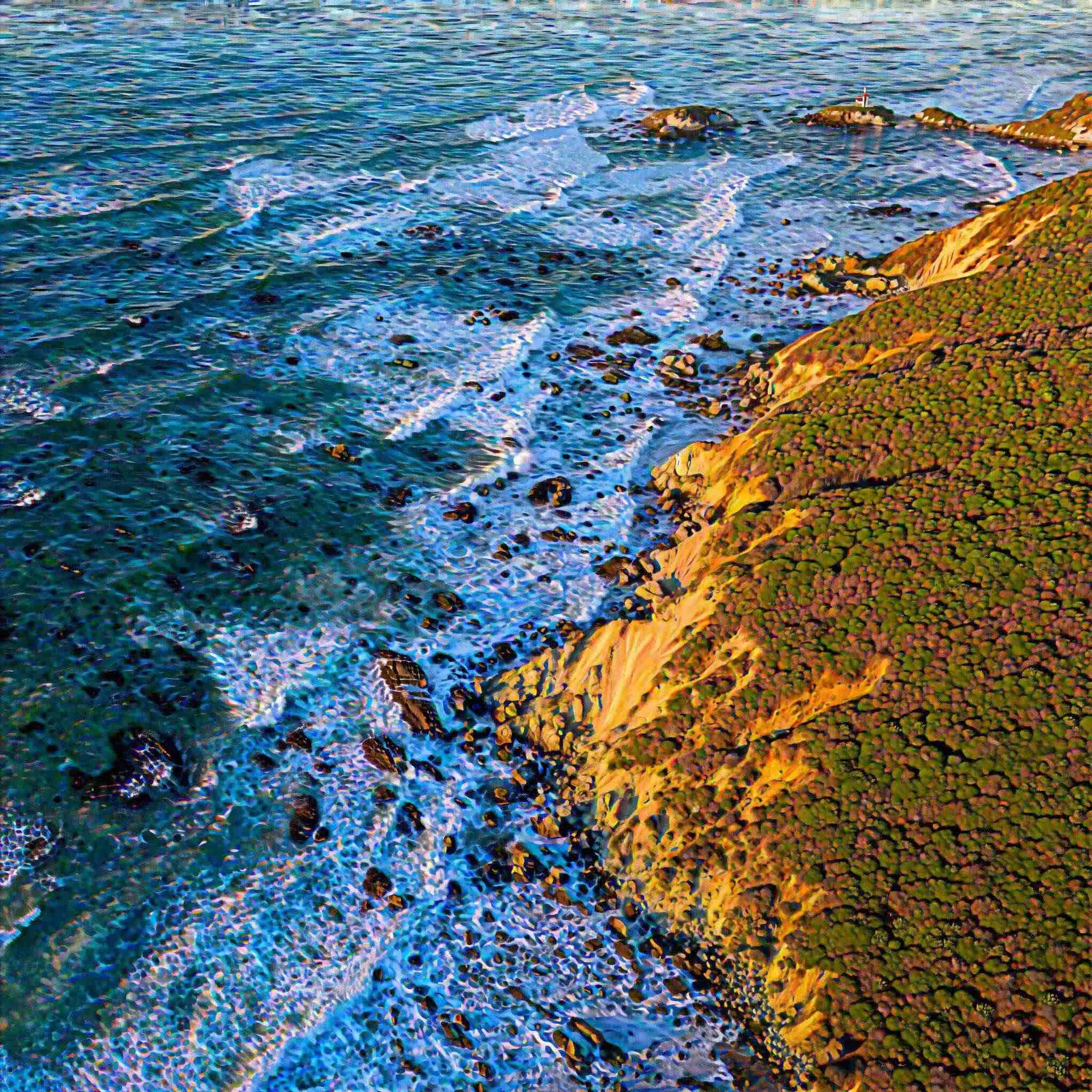} 
    \\
    \bottomrule
    \end{tabular}
    \caption{\textbf{Ablation on $\lambda_{\max}$.} A small $\lambda_{\max}$ (e.g., $1.1$) yields insufficient concentration and suboptimal quality, while a large value (e.g., $1.6$) causes over-sharpening. An intermediate setting achieves the best balance.}
    \label{fig:app_max_lambda}
\end{figure*}

\begin{figure*}[tbp]
    \centering
    \begin{tabular}{
        >{\centering\arraybackslash}m{0.25\textwidth} |
        >{\centering\arraybackslash}m{0.25\textwidth} |
        >{\centering\arraybackslash}m{0.25\textwidth}
    }
    \toprule
    $\lambda_{\min}=1.0$ & $\lambda_{\min}=1.1$ & $\lambda_{\min}=1.2$\\ 
    \midrule
    \includegraphics[width=\linewidth,valign=m]{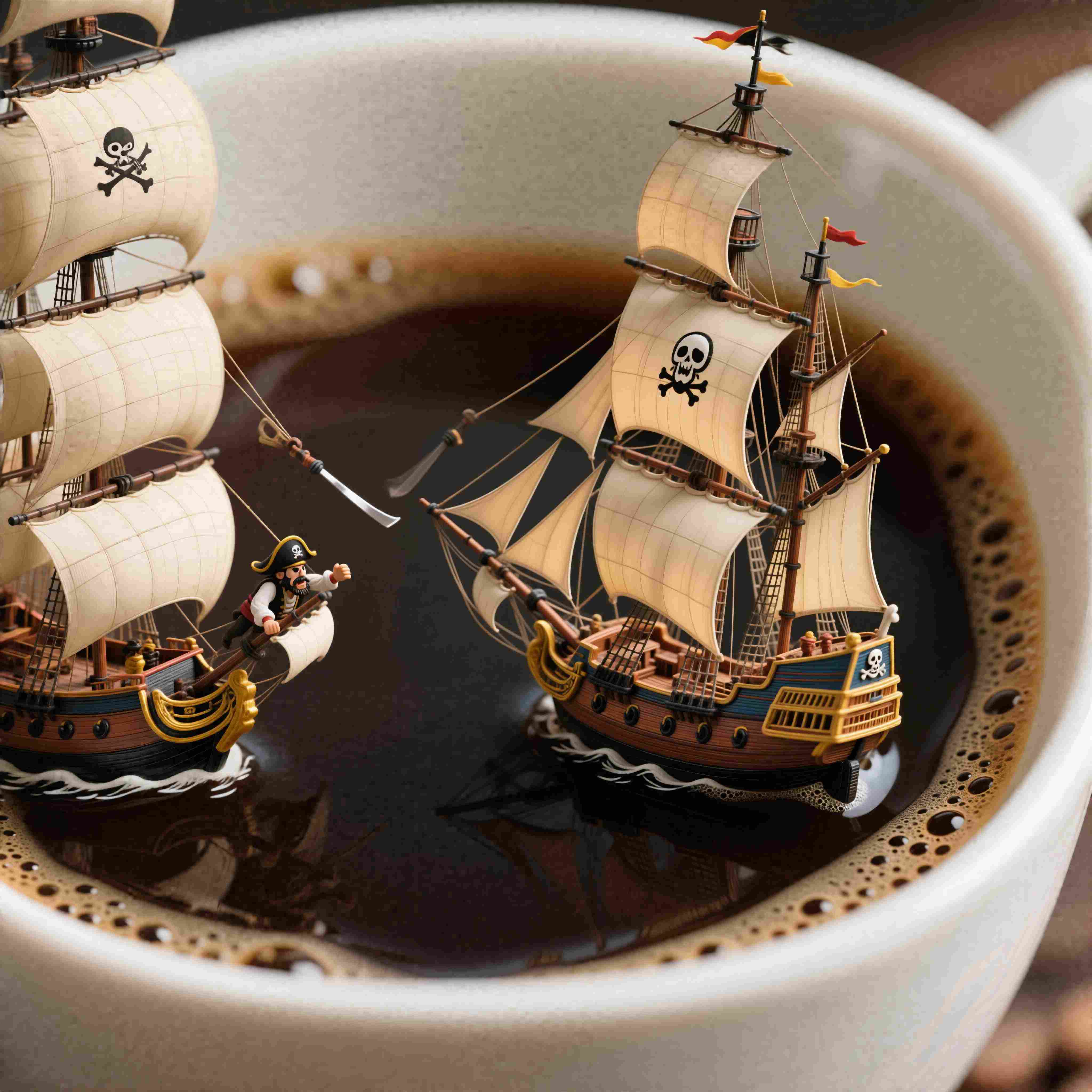} &
    \includegraphics[width=\linewidth,valign=m]{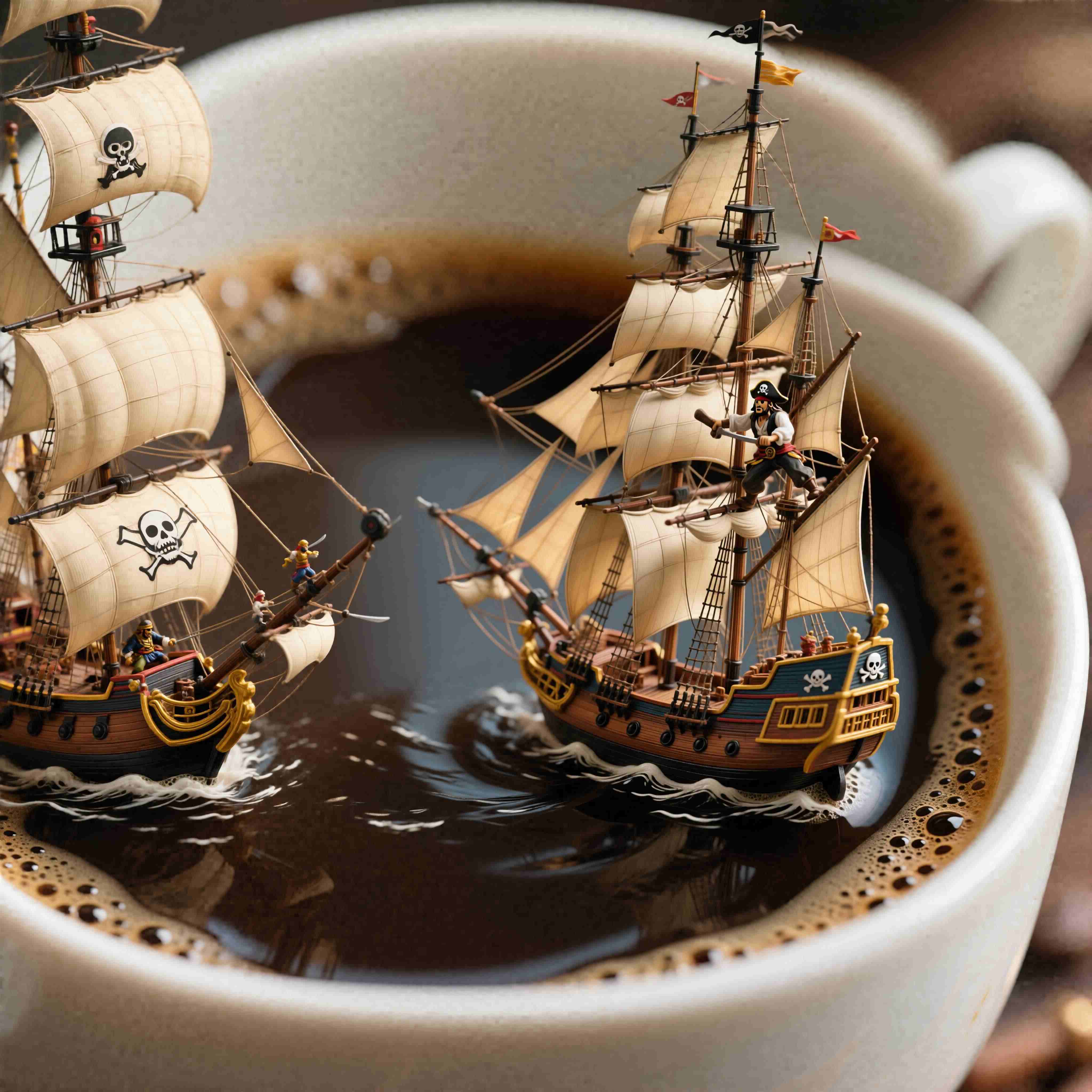} &
    \includegraphics[width=\linewidth,valign=m]{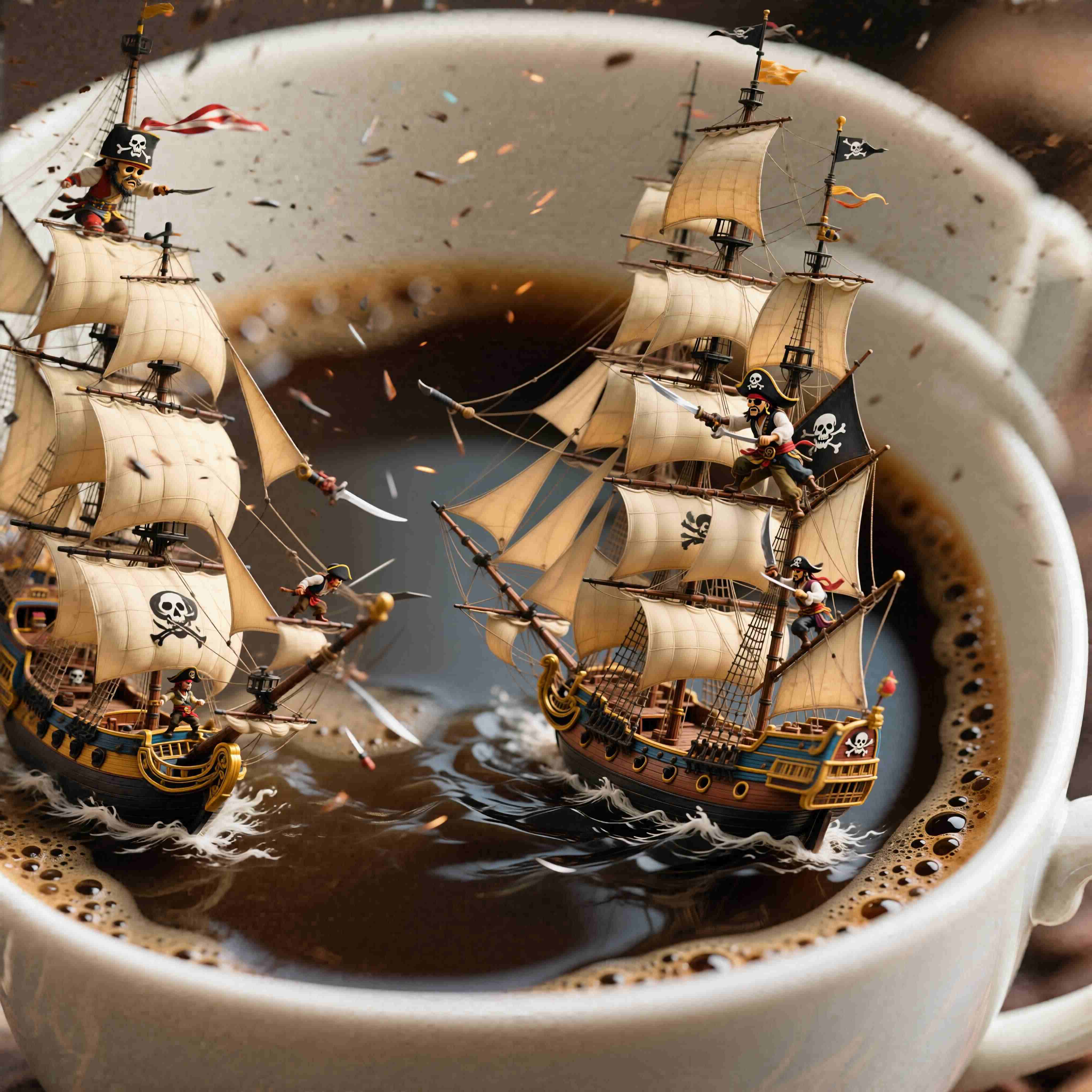} 
    \\
    \bottomrule
    \end{tabular}
    \caption{\textbf{Ablation on $\lambda_{\min}$.} Even a small $\lambda_{\min}$ (e.g., $1.1$) introduces structural inconsistencies, showing that not all attention patterns should be concentrated and supporting our choice to apply stronger focusing only to local patterns.}
    \label{fig:app_min_lamba}
\end{figure*}

\paragraph{Ablation studies for the hyperparameters $\lambda_{\min}, \lambda_{\max}$ and $p$.}
As shown in Fig.~\ref{fig:app_visual_p}, different exponent values $p$ correspond to different degrees of concentration. A smaller $p$ applies stronger focusing to a larger portion of the attention patterns, leading to more concentrated attention. As shown in Fig.~\ref{fig:app_p}, when $p$ is too small (e.g., $p=0.2$), the attention becomes overly concentrated, leading to over-sharpening and structural artifacts; when $p$ is too large (e.g., $p=5.0$), the attention is insufficiently focused and the visual quality degrades. An intermediate value yields the best trade-off. Similarly, Fig.~\ref{fig:app_max_lambda} shows that a small $\lambda_{\max}=1.1$ does not provide enough concentration, leading to suboptimal visual quality, while a large $\lambda_{\max}=1.6$ results in over-sharpening. An intermediate value yields the best trade-off. For $\lambda_{\min}$ (see Fig.~\ref{fig:app_min_lamba}), even a relatively small value such as $\lambda_{\min}=1.1$ introduces structural inconsistency, indicating that not all attention patterns should be concentrated. This further supports our design of selectively applying stronger focusing only to local patterns.

\begin{figure*}[b]
    \centering
    \begin{tabular}{
        >{\centering\arraybackslash}m{0.22\textwidth} |
        >{\centering\arraybackslash}m{0.22\textwidth} |
        >{\centering\arraybackslash}m{0.22\textwidth}|
        >{\centering\arraybackslash}m{0.22\textwidth}
    }
    \toprule
    global factor $\lambda=1.0$ & global factor $\lambda=1.1$ & global factor $\lambda=1.2$ & our entropy-guided factor    \\ 
    \midrule
    \includegraphics[width=\linewidth,valign=m]{images/Fig5/2_1_0_baseline.jpg} &
    \includegraphics[width=\linewidth,valign=m]{images/Fig5/2_1_1.jpg} &
    \includegraphics[width=\linewidth,valign=m]{images/Fig5/2_1_2.jpg} &
    \includegraphics[width=\linewidth,valign=m]{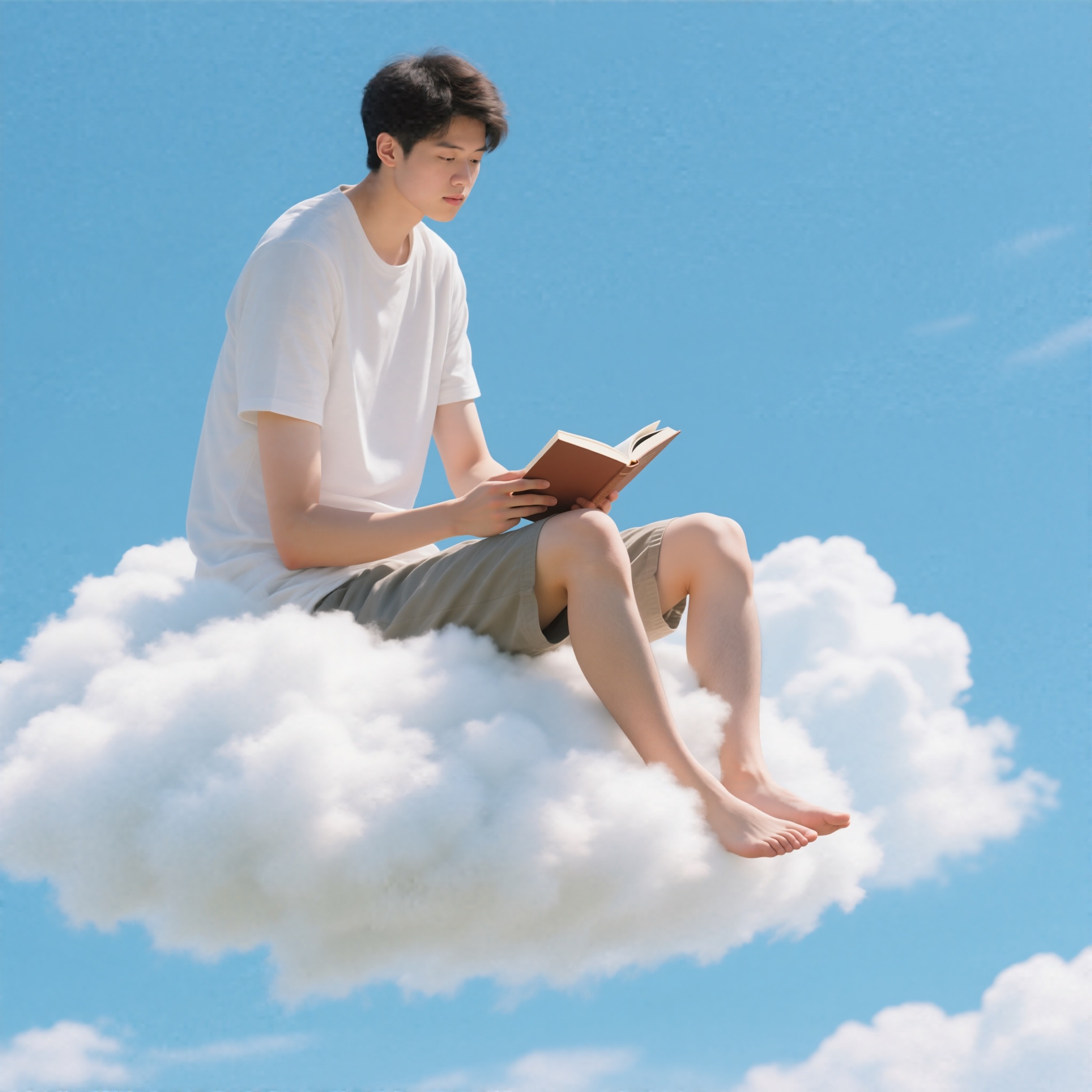}
    \\
    \midrule
    \includegraphics[width=\linewidth,valign=m]{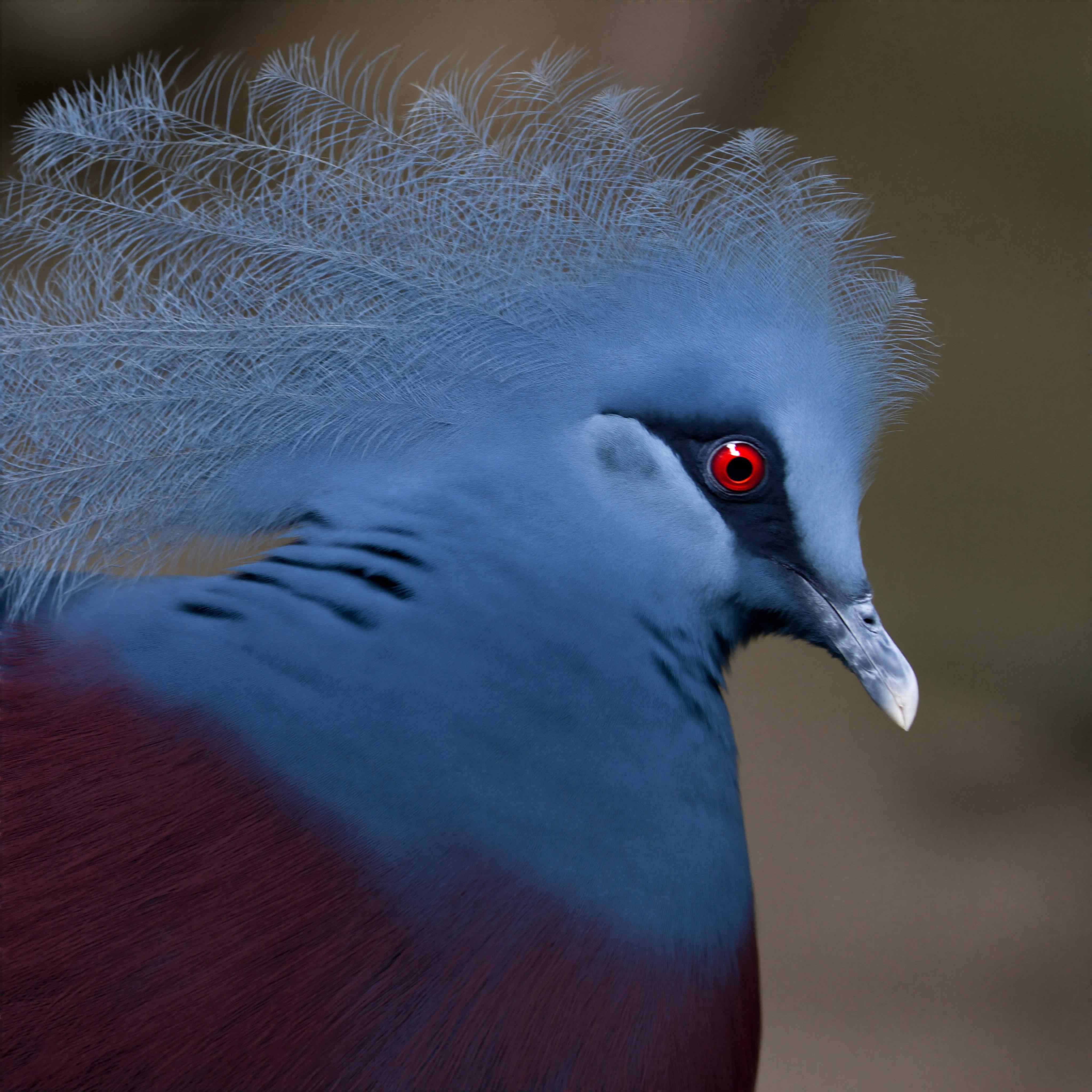} &
    \includegraphics[width=\linewidth,valign=m]{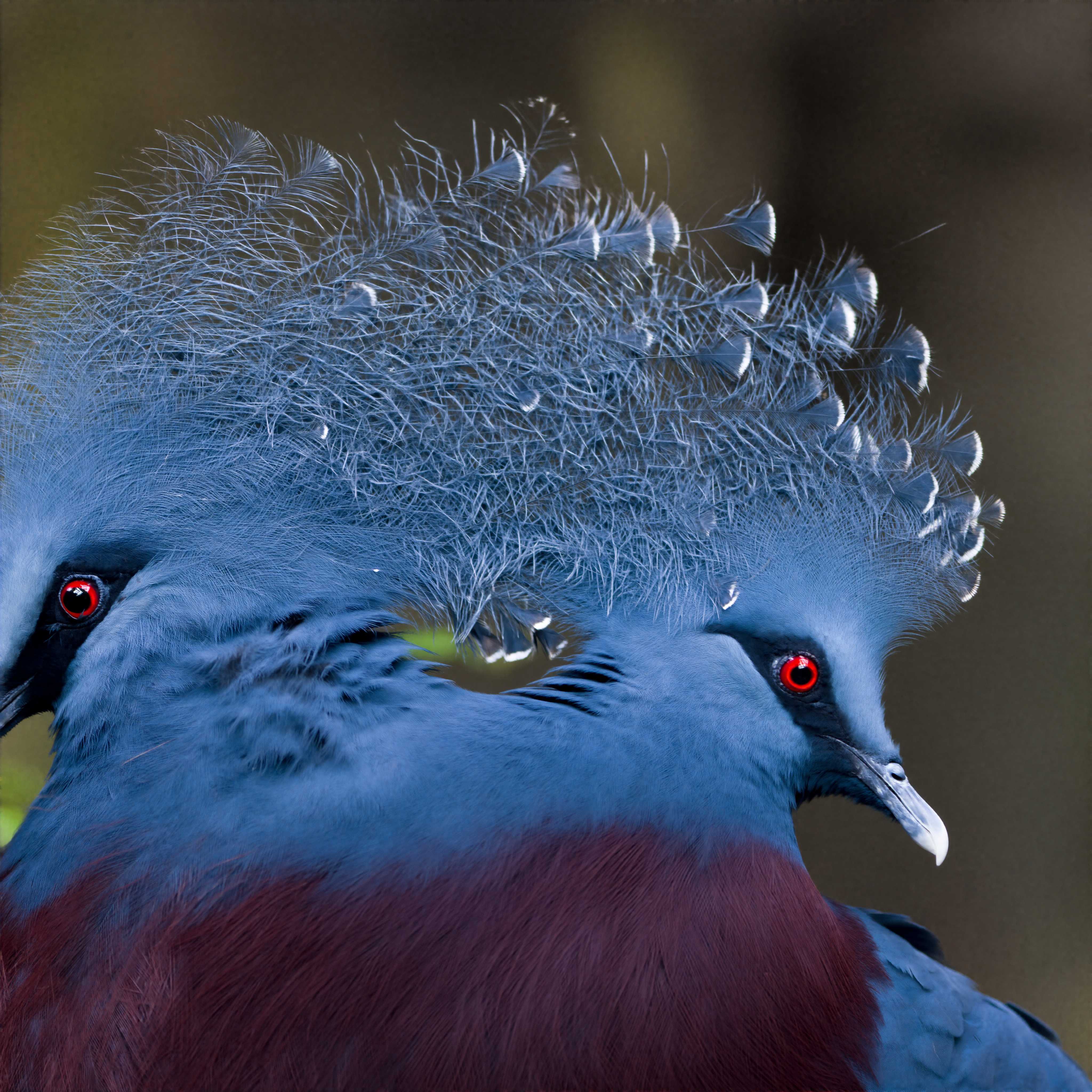} &
    \includegraphics[width=\linewidth,valign=m]{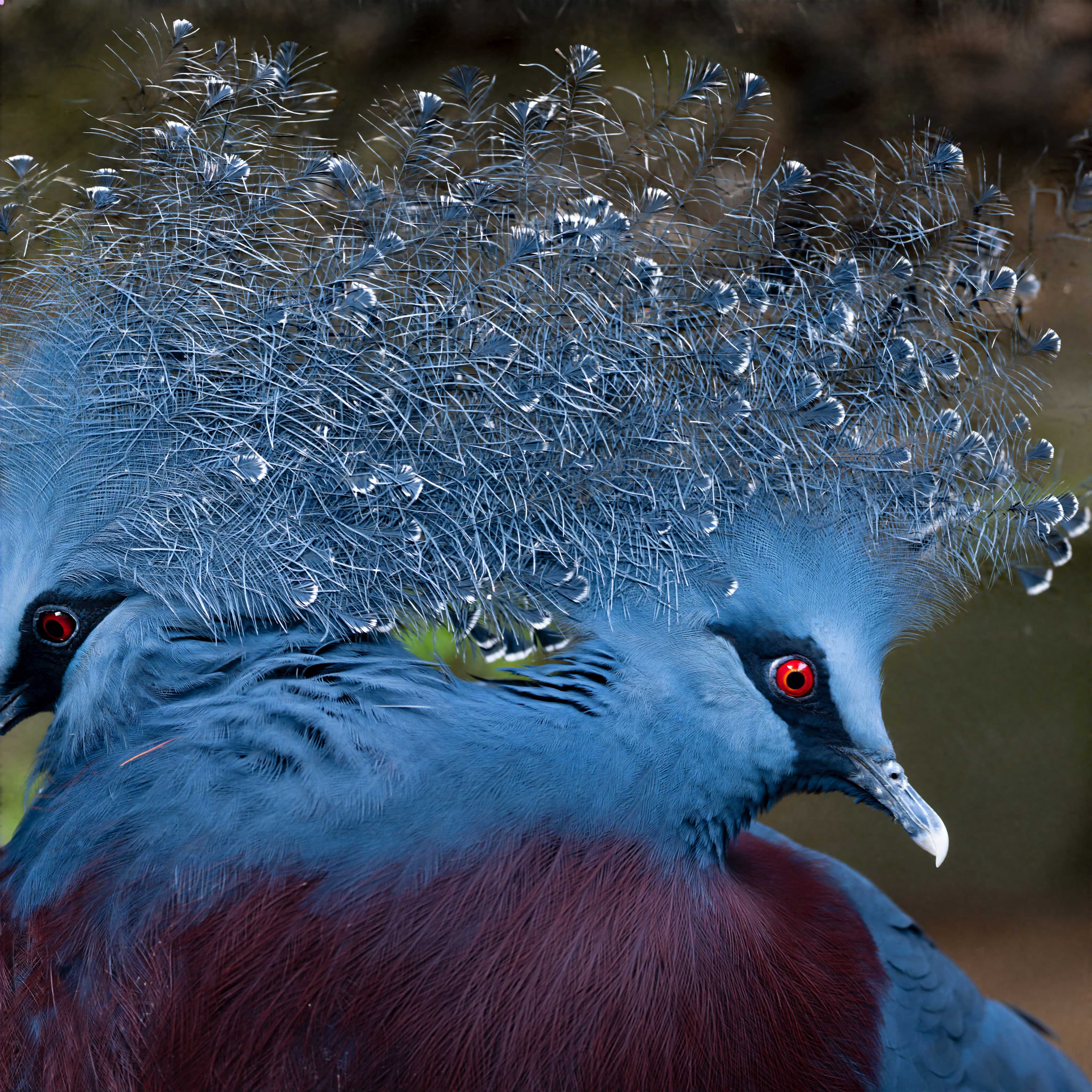} &
    \includegraphics[width=\linewidth,valign=m]{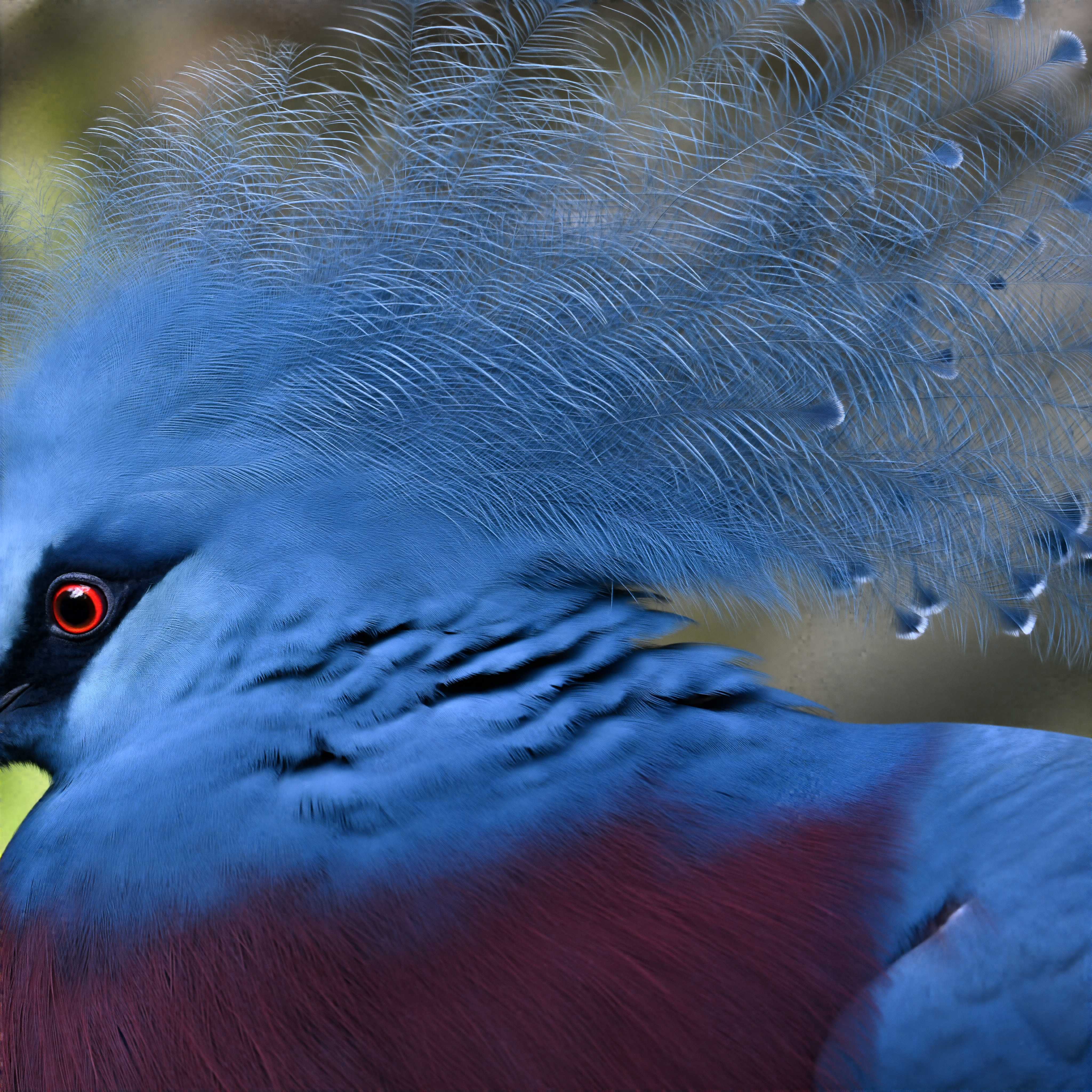}
    \\
    \midrule\includegraphics[width=\linewidth,valign=m]{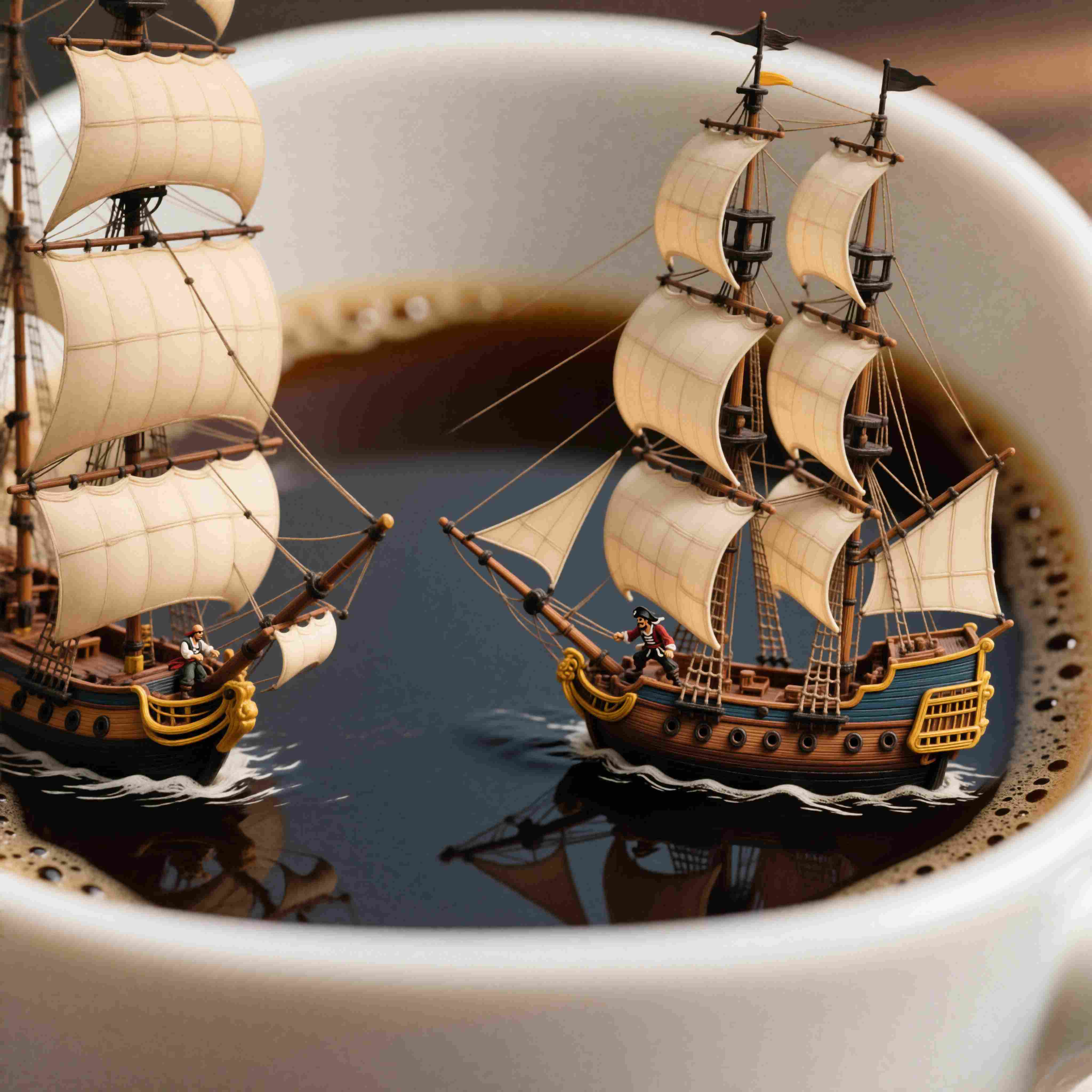} &
    \includegraphics[width=\linewidth,valign=m]{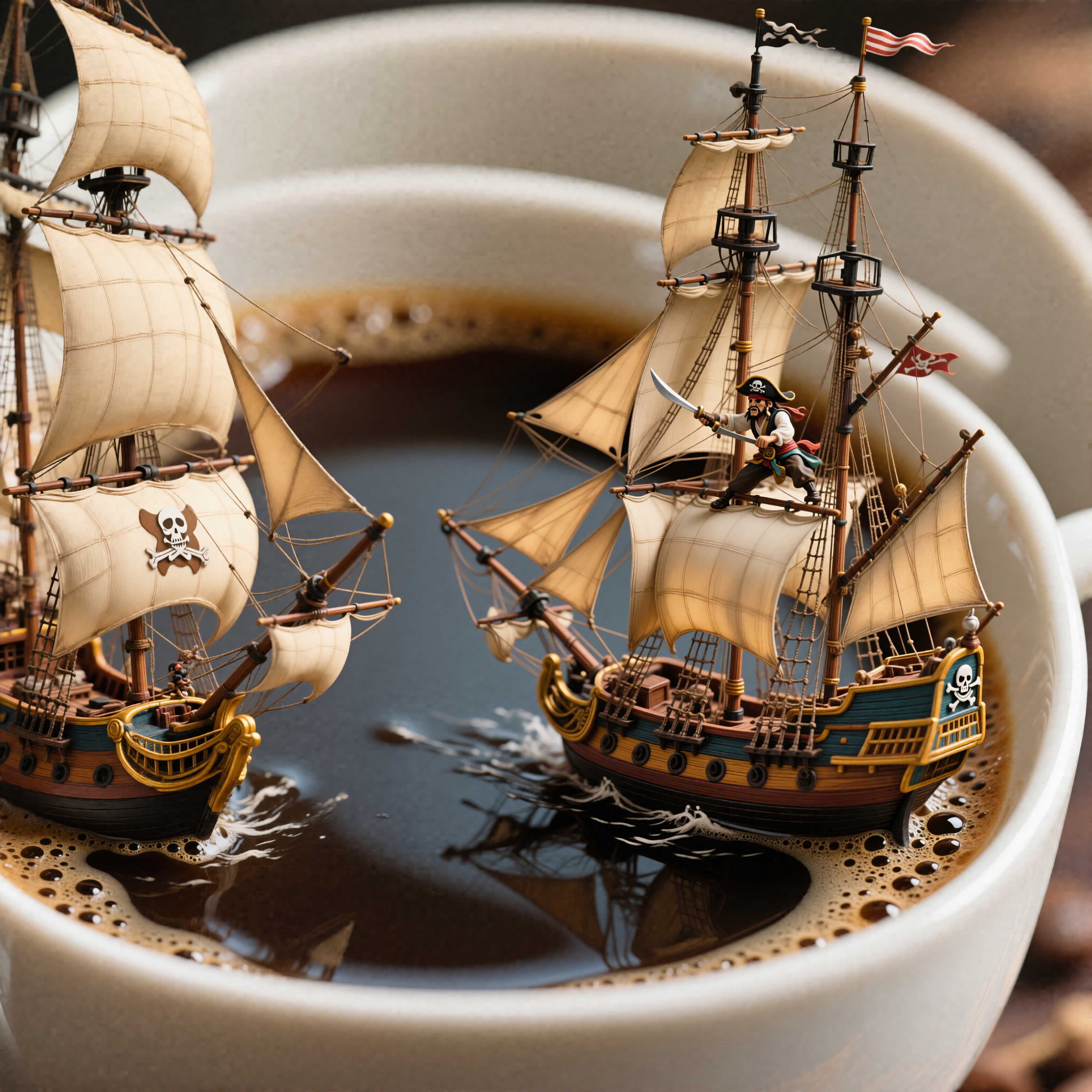} &
    \includegraphics[width=\linewidth,valign=m]{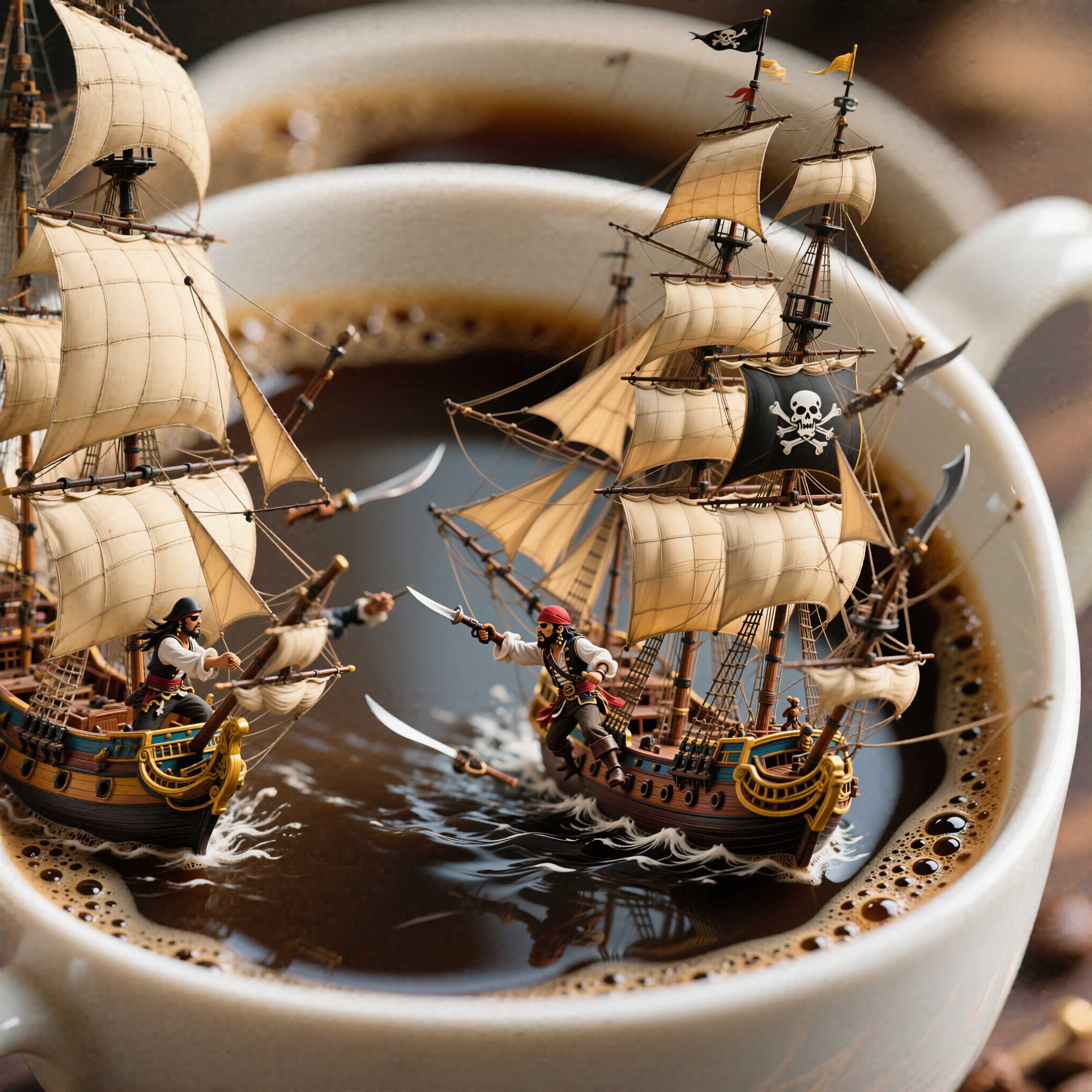} &
    \includegraphics[width=\linewidth,valign=m]{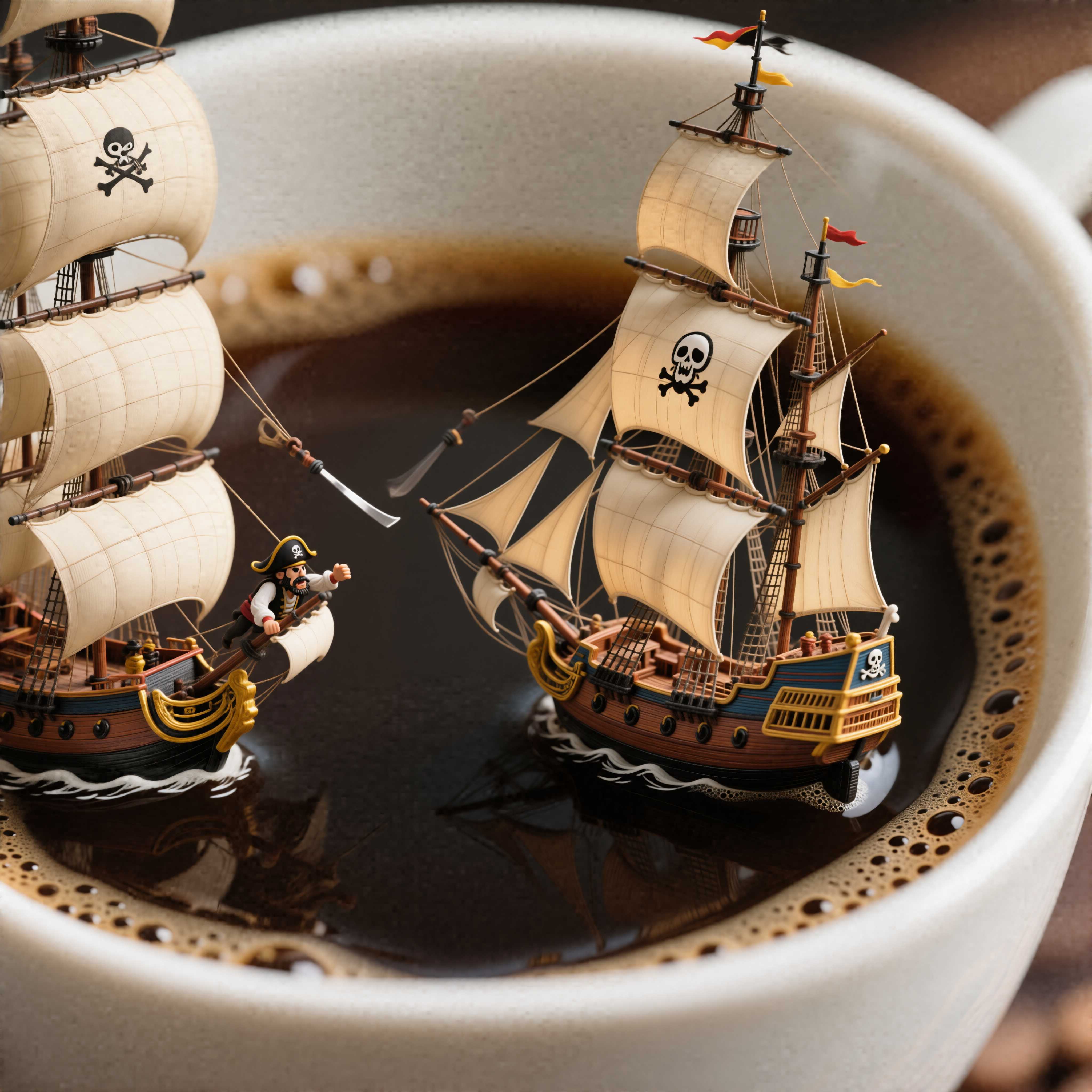}
    \\
    \bottomrule
    \end{tabular}
    \caption{\textbf{Ablation studies.} Using a single global concentration factor produces a trade-off: small factors (e.g., $\lambda=1.0, 1.1$) hurt visual quality, while large factors introduce structural inconsistency (e.g., $\lambda=1.2$), and even mild values (e.g., $\lambda=1.1$) cause noticeable mismatches. In contrast, our entropy-guided strategy applies weak focusing to global patterns and stronger focusing to local patterns, achieving both high visual quality and stable structure.}
    \label{fig:app_global}
\end{figure*}

\paragraph{Ablation studies for entropy-guided adaptive attention concentration.}
As shown in Fig.~\ref{fig:app_global}, using a single global concentration factor leads to a clear trade-off: a smaller focus factor results in poor visual quality, whereas a larger one introduces structural inconsistencies. Moreover, even a small focus factor (e.g., $\lambda = 1.1$) still causes noticeable structural mismatches. In contrast, our entropy-guided adaptive attention concentration strategy applies a mild focus factor to the global attention pattern to preserve structural consistency, while assigning a stronger concentration factor to the local attention pattern to enhance visual quality, thereby simultaneously achieving high-quality generation and strong structural consistency.

\renewcommand\arraystretch{1.1}

\renewcommand\arraystretch{1.1}

 \begin{figure*}
    \centering
    \includegraphics[width=2.0\columnwidth]{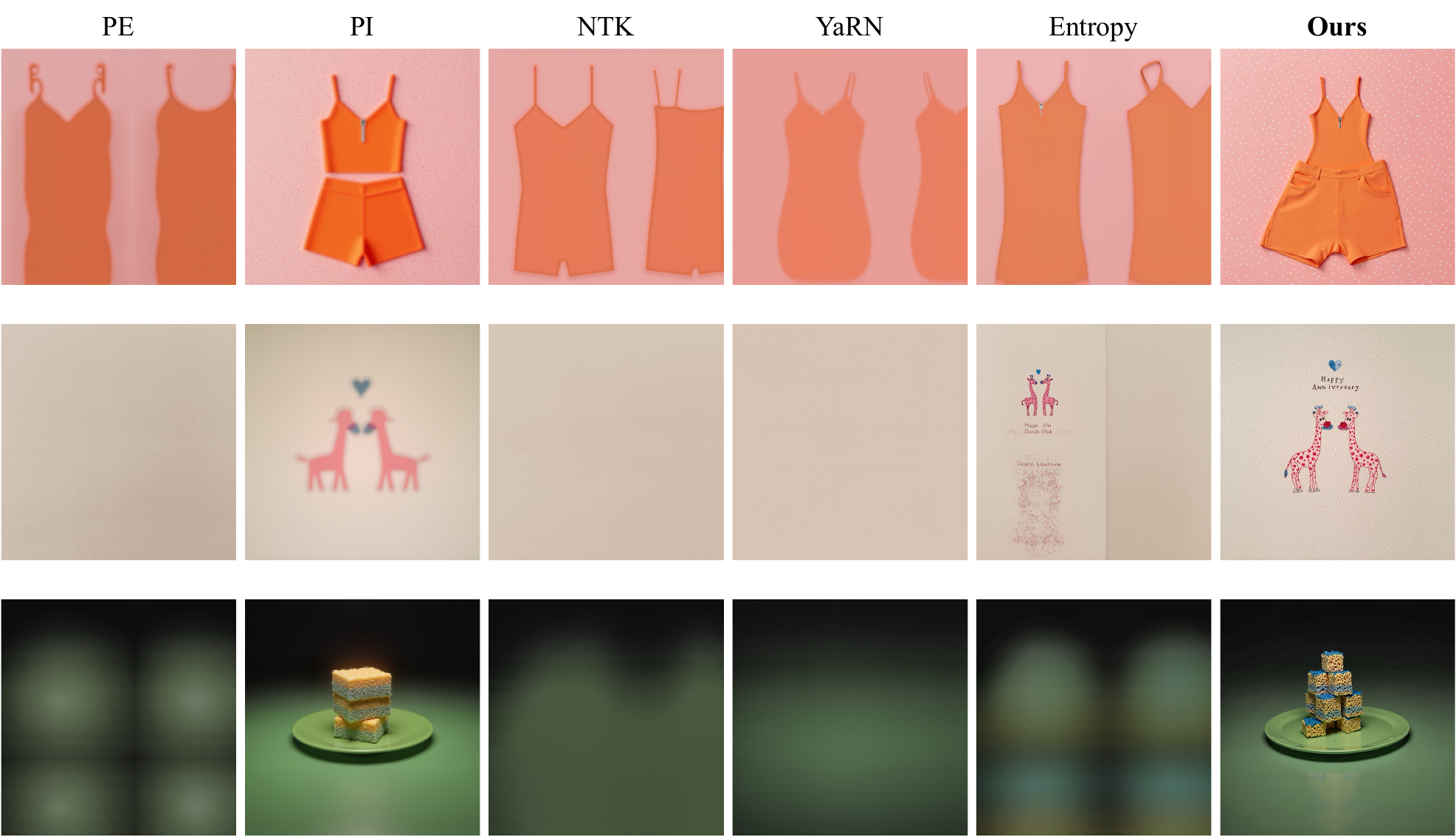}

\caption{\textbf{Qualitative comparison in direct resolution extrapolation on Flux.} Our method outperforms baselines by delivering high visual quality while mitigating content repetition.}    
    \label{fig: app_qualitive}
\end{figure*}

 \begin{figure*}
    \centering
    \includegraphics[width=2.0\columnwidth]{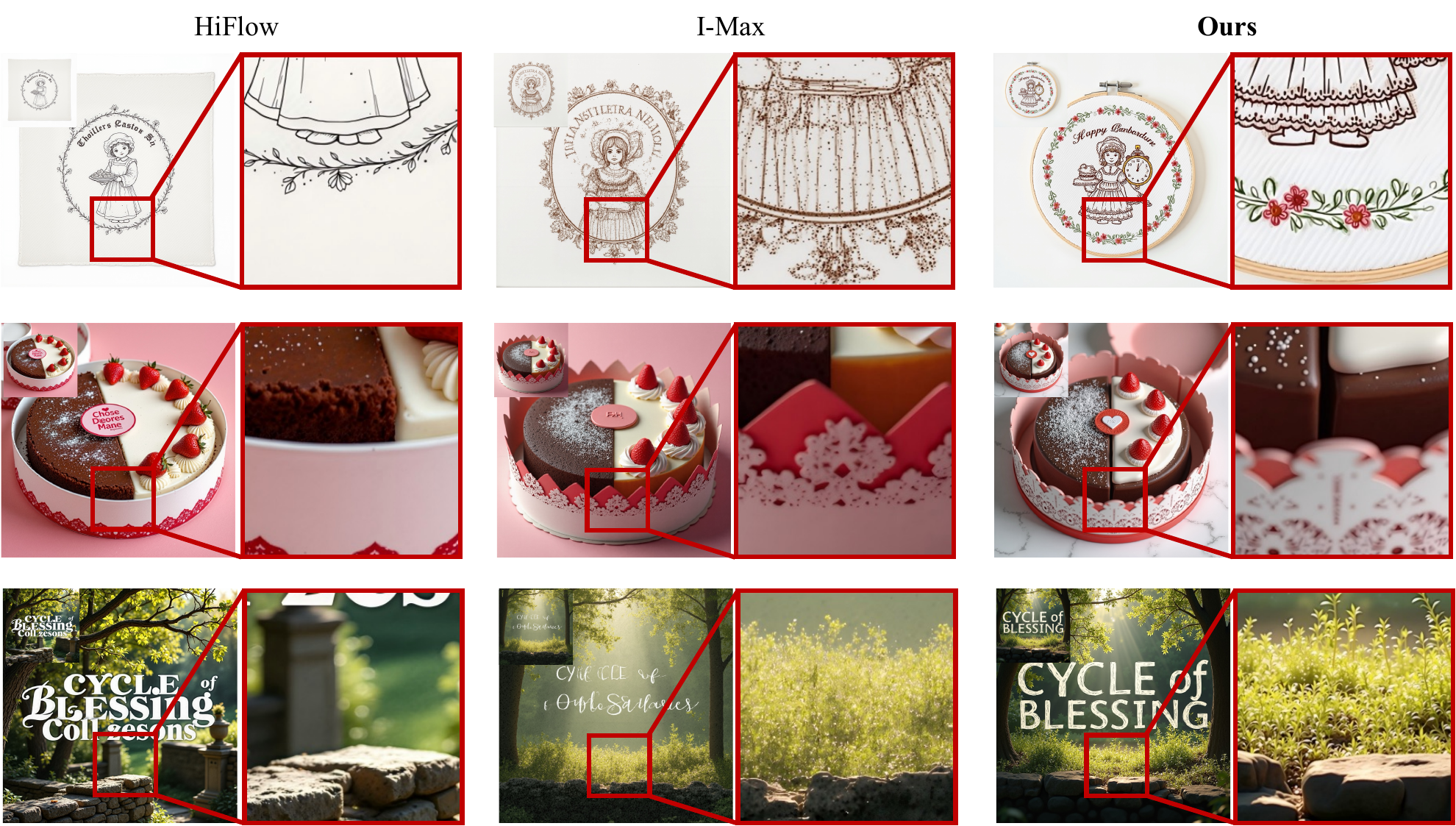}
\caption{\textbf{Qualitative comparison on guided resolution extrapolation.} Our method outperforms baselines across by delivering high visual quality.}    
    \label{fig: app_super}
\end{figure*}

 \begin{figure*}
    \centering
    \includegraphics[width=2.0\columnwidth]{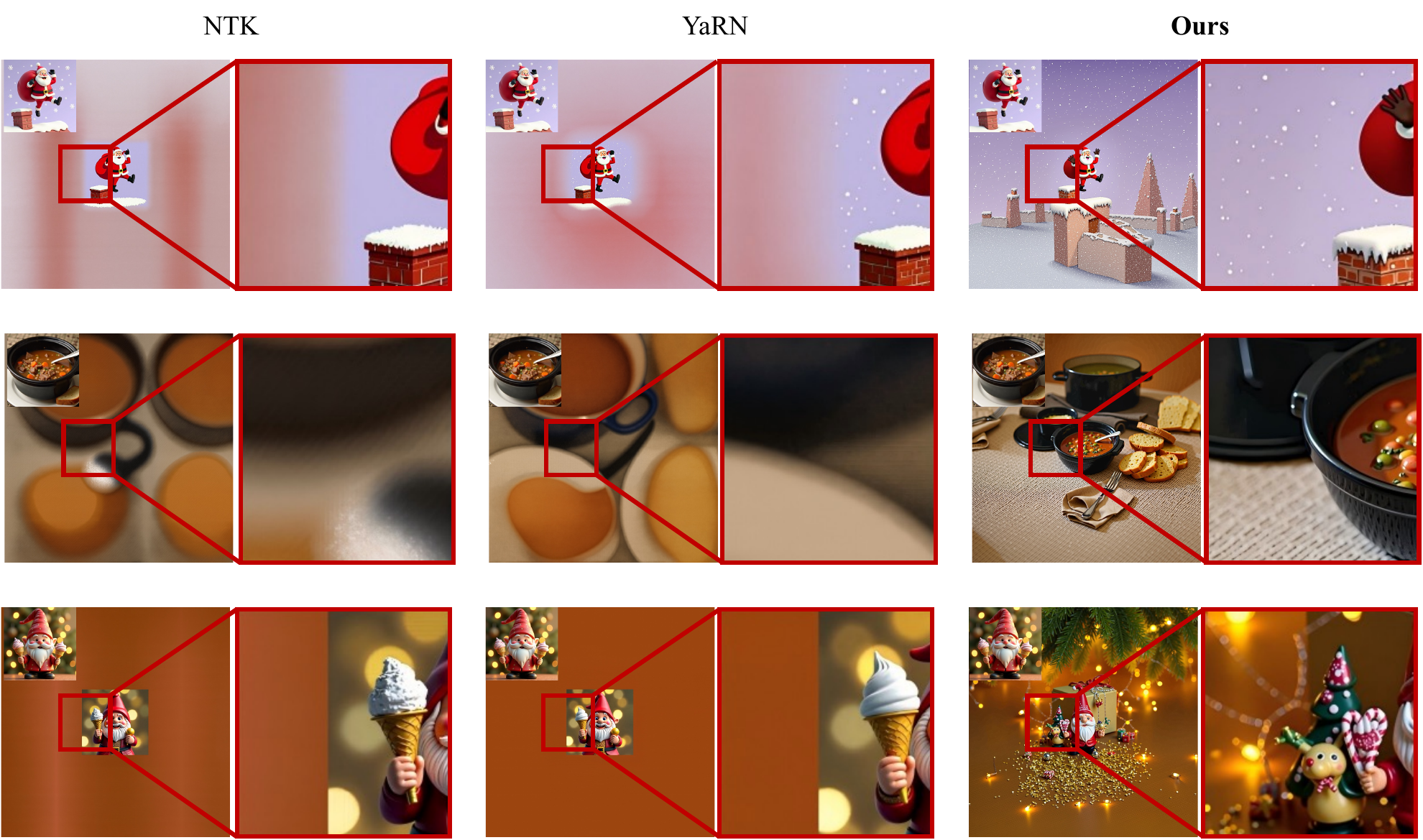}

\caption{\textbf{Qualitative comparison on guided view extrapolation.} Our method outperforms baselines by delivering high visual quality while mitigating content repetition.}    
    \label{fig: app_view}
\end{figure*}


\end{document}